\documentclass{article}

\usepackage{arxiv}

\usepackage[utf8]{inputenc} 
\usepackage[T1]{fontenc}    
\usepackage[pagebackref]{hyperref} 
\usepackage{url}            
\usepackage{booktabs}       
\usepackage{nicefrac}       
\usepackage{microtype}      
\usepackage{lipsum}         
\usepackage{wrapfig}        
\usepackage{graphicx}
\usepackage[square,numbers,sort&compress]{natbib}
\usepackage{doi}
\usepackage{comment}
\usepackage{xcolor}
\usepackage[table,xcdraw]{xcolor}
\usepackage{placeins}

\usepackage{tabularx}

\usepackage{amsfonts}       
\usepackage{isomath}
\usepackage{amssymb}
\usepackage{amsmath, amsthm}
\usepackage{amsthm}
\usepackage{stmaryrd}
\usepackage{makeidx}
\usepackage{lmodern}
\usepackage{blindtext}
\usepackage{MnSymbol}
\usepackage{mathtools}

\usepackage{wrapfig}

\usepackage{cleveref}       

\title{
Event-Driven Reinforcement Learning Enables Long-Horizon Control in Semiconductor Fabrication
}

\date{}

\author{
Yavar Yeganeh~\textsuperscript{1}\thanks{Correspondence to \href{mailto:yavar.taheri@polimi.it}{yavar.taheri@polimi.it}}~
\And
Mahsa Shekari~\textsuperscript{1}
\And
Nicla Frigerio~\textsuperscript{1}
\And
Daniele Pagano~\textsuperscript{2}
\And
Andrea Matta~\textsuperscript{1}
\\~\\~\\
\textsuperscript{1}~Politecnico di Milano \qquad
\textsuperscript{2}~STMicroelectronics 
}

\renewcommand{\shorttitle}{Event-Driven RL for Long-Horizon Control}

\hypersetup{
  pdftitle={Event-Driven Reinforcement Learning Enables Long-Horizon Control in Semiconductor Fabrication},
  pdfsubject={Reinforcement learning for long-horizon event-driven control in semiconductor fabrication},
  pdfauthor={Yavar Yeganeh, Mahsa Shekari, Nicla Frigerio, Daniele Pagano, Andrea Matta},
  pdfkeywords={reinforcement learning, event-driven control, temporal-difference learning, long-horizon control, semiconductor manufacturing, wafer fabrication, dispatching, offline reinforcement learning, online reinforcement learning}
}

\begin{document}

\maketitle

\begin{abstract}

Reinforcement learning promises to optimize sequential decisions in large-scale systems. Semiconductor manufacturing systems are stochastic and highly constrained environments where heterogeneous wafers traverse hundreds of processing steps across extensive equipment networks. These characteristics yield complex, high-dimensional decision problems with delayed feedback and long-horizon requirements, complicating production planning and control. We propose a deep reinforcement learning framework for multi-objective policy optimization at this scale. Specifically, we formulate control as a centralized-agent problem, where a core policy coordinates system-wide decisions, while system evolution is represented as an interconnected temporal process driven by discrete events. Accordingly, we develop a tailored event-driven temporal-difference formulation that remains general and can be integrated with various policy optimization methods under relevant training settings. We investigate several core model-free algorithms incorporated into this framework and evaluate their effectiveness using high-fidelity simulations of diverse, industry-real operating scenarios. Across extensive validation experiments, agents trained in both offline and online settings show significant and consistent gains in throughput and utilization. We further evaluate performance and generalization across training phases, clarifying the relative strengths of alternative reinforcement learning formulations and algorithms. Overall, the results support the scalability, generality, and transferability of the proposed framework for controlling event-driven complex adaptive systems.

\end{abstract}


\section{Introduction}
Semiconductor manufacturing is critical for advancements across diverse sectors, including computing, electronics, telecommunications, transportation, and healthcare. However, front-end wafer fabrication (fab) remains among the most demanding production environments \cite{monch2012production}. In fabs, wafers traverse long, re-entrant process flows across heterogeneous equipment, subject to stochastic processing times, transport delays, and equipment failures. Numerous wafers are processed simultaneously across equipment with batching requirements and diverse product mixes, all while adhering to technological and quality constraints. This dynamic environment results in continuously evolving state and bottleneck conditions. These characteristics make semiconductor fabs a representative example of complex adaptive systems: large-scale systems composed of many interacting components whose collective behavior evolves through local decisions, stochastic disturbances, resource constraints, and feedback effects. Similar event-driven, interconnected, multi-stage structures also arise in healthcare operations, logistics and supply chains, advanced manufacturing, and telecommunication networks. Effective decision-making and planning in such fabs is essential and occurs across multiple hierarchical levels to optimize competing Key Performance Indicators (KPIs)—such as throughput and utilization \cite{monch2012production}. Because static, rule-based policies struggle to remain effective across these fluctuating operating regimes, there is increasing interest in adaptive, data-driven methodologies that navigate this scale of complexity~\cite{stricker2018reinforcement,waschneck2018deep,zhou2022reinforcement,tassel2023semiconductor}. 

Reinforcement learning (RL) provides a natural approach for such sequential decision problems, as agents optimize policies through interaction with an environment and feedback \cite{sutton2018reinforcement}. Combined with deep neural networks, which enable high-dimensional function approximation and representation learning \cite{lecun2015deep,bishop2023deep,lu2017expressive}, RL has supported advances in domains ranging from game playing and fusion control to chip design and large language model (LLM) reasoning~\cite{mnih2015human,degrave2022magnetic,mirhoseini2021graph,guo2025deepseek}. RL agents have also been developed for manufacturing decision levels ranging from scheduling and adaptive dispatching in flexible job shops \cite{khadivi2025deep} to semiconductor fab control \cite{waschneck2018deep,kuhnle2019autonomous}. These problems are commonly formulated as event-driven decision processes, where actions are triggered by machine-idle or lot-selection events rather than fixed time intervals \cite{tassel2023semiconductor,taheri2024active}. In fabs, however, actions initiate temporally extended and partially overlapping events; treating successive decisions as ordinary one-step transitions can therefore misalign temporal credit assignment over long horizons~\cite{taheri2024active,klissarov2025discovering}. 

Despite this promise and success in small-scale testbeds \cite{stockermann2025scalability,sood2024supporting}, semiconductor manufacturing exposes several RL failure modes simultaneously. Along with high dimensional state and adaptive action spaces, existing formulations often do not adequately capture the long-horizon, delayed-reward nature of these systems, making credit assignment and robust policy learning difficult. At the same time, advanced training and deployment paradigms such as offline training, regularization, and tailored exploration remain underexplored, while systematic comparisons across methods are scarce, limiting evidence on generalization and transferability. Scalability also remains unresolved, as many studies simplify the problem to localized control instead of learning system-wide policies under real industrial complexity.

Broader agentic pipelines can include learned encoders, model-based components, LLM/foundational model tools~\cite{plaat2025agentic}, policy fine-tuning procedures~\cite{guo2025deepseek}, or scalable multi-agent decompositions, while here we isolate the policy-optimization layer and use feature-based encodings to maintain focus. We therefore address the central problem of sequential decision-making and data-driven policy optimization in these systems using deep RL. Specifically, we develop a framework based on a single centralized, \emph{model-free} agent that controls all components of the system, tailored to their temporally interconnected structure. This design enables parameter sharing across decision units, improving scalability in high-dimensional settings while simplifying training relative to multi-agent approaches that require explicit coordination and equilibrium design. The framework can be applied at different decision stages; here, we focus on dispatching, the most fine-grained level of operational control, where system objectives are optimized by directly selecting from adaptive sets of candidates. We further consider \emph{event-driven} decision-making and examine tailored formulations across several algorithms with different inductive biases, as well as multi-objective optimization and constraint adherence. In addition to efficient training approaches, we study distinct regimes. These range from offline methods based on historical data, despite their well-known challenges, including distributional shift and divergence from the behavior policy \cite{zitovsky2023revisiting,haarnoja2018soft,kostrikov2021offline}, to online interaction using a high-fidelity simulator of real large-scale industrial scenarios.

Our main contributions are as follows:
\begin{itemize}
\item \emph{Modular event-driven RL framework.} We introduce a framework for efficient policy interaction and optimization in complex adaptive systems, with semiconductor fabrication serving as a challenging, real-world, large-scale representative. 
\item \emph{Centralized formulation.} We formulate fab control as a centralized policy-optimization problem over event-driven system dynamics, enabling system-wide decision-making under diverse operating conditions. 
\item \emph{Multi-phase policy optimization.} We develop tailored RL formulations and training strategies that support learning across offline and online phases while remaining compatible with different policy-optimization methods.
\item \emph{Generalization-focused evaluation.} We provide a comprehensive evaluation of performance, generalization, and transferability across diverse operating scenarios, demonstrating consistent performance gains and clarifying the relative behavior of alternative RL formulations and algorithms.
\end{itemize}

The remainder of this paper is organized as follows. Section~\ref{sec:related} reviews related work. Section~\ref{sec:policy} introduces the methodological framework that investigates policy optimization. Section~\ref{sec:experiments} presents experimental results and ablation studies. Section~\ref{sec:discussion} discusses the implications of the findings and outlines directions for future research, followed by the  conclusion in Section~\ref{sec:conclusion}.

\section{Related Work}
\label{sec:related}

Existing fab studies differ in action abstraction. Some methods select among predefined dispatching rules, keeping the action space compact and compatible with industrial control logic~\cite{sood2024supporting,zhang2023reinforcement}. Finer-grained formulations instead directly select candidate lots or learn priority scores, providing greater control resolution but creating variable, state-dependent action spaces that are more prone to scalability and credit-assignment challenges~\cite{lee2022deep,sakr2023simulation,tassel2023semiconductor,zhang2022dynamic}. In practice, however, this can be combined in a hybrid manner, with rule-based logic or masking to still enforce hard technological constraints and feasibility~\cite{wang2025novel}. Although most studies adopt single-agent formulations, multi-agent decompositions across equipment, work centers, or system cells have also been explored to improve scalability \cite{xing2023distributed,sood2024supporting}. Recent studies further move beyond single-KPI optimization toward multi-objective formulations and expand the control scope from localized or bottleneck problems to system-wide decision making \cite{zhou2022reinforcement,liu2022dynamic,stockermann2025scalability}.

While problem formulation and policy-optimization design are central to RL solutions, state encoding and reward signals are also important for policy effectiveness. Commonly, states and actions are represented using structured, hand-engineered features derived from domain knowledge, yielding low-dimensional representations of the system \cite{waschneck2018deep,stockermann2025scalability}. Recent works apply dimensionality reduction \cite{tassel2023semiconductor} or representation learning \cite{ghasemi2023deep}, including effective use of graph neural networks \cite{wu2020comprehensive,schulz2022graph}, especially in scheduling problems \cite{liu2025flexible}, where the spatio-temporal structure of the schedule adheres to graphical structures \cite{khadivi2025deep}. These architectures, when integrating representation learning and policy, enable \emph{end-to-end} optimization, which further boosts performance \cite{zhang2020learning}. For rewards, the slow response of system-level KPIs, such as throughput or cycle time, leads many studies to use shaped rewards based on intermediate events, local penalties, or weighted KPI combinations to improve credit assignment \cite{chang2022deep,wang2021fuzzy,yedidsion2022deep,tassel2023semiconductor}. Algorithmically, value-based methods—including variants of Q-learning—dominate the field \cite{kuhnle2019autonomous,luo2020dynamic}. Multi-agent variants such as QMIX appear when control is spatially decomposed \cite{sood2024supporting,xing2023distributed}. Policy-gradient methods, especially PPO, are typically adopted when smoother policy optimization or larger action spaces are required \cite{schulman2017proximal,zhang2022dynamic}. Recent work also incorporates hierarchical control \cite{zhang2024adaptive}, explainability modules \cite{immordino2025explainable}, and hybrid RL with evolutionary methods \cite{stockermann2023dispatching, stockermann2025scalability}, improving how the RL problem is structured rather than adopting new backbones or conceptualizations.

Despite this progress, the practical advancement of RL in these systems has been limited. A key challenge is the formulation of system control as a temporally extended sequential decision problem. Such problems are characterized by \emph{delayed} (sparse) rewards and \emph{long-horizon} dependencies, which complicate credit assignment and policy optimization \cite{yeganeh2025deep,arjona2019rudder}. Specifically, the consequences of actions are rarely observable immediately, and goal-driven tasks demand effective planning over extended temporal horizons. As a result, standard RL formulations may be insufficient to capture the delayed effects and system-wide interactions that arise at scale. This is further compounded by the limited use of advanced RL regimes: self-supervised and pretraining methods, including offline training, are rarely employed \cite{nair2020awac,kumar2020conservative}. When offline training is used, conservative value-learning objectives (e.g., CQL) and implicit policy extraction methods (e.g., IQL) are common tools to mitigate extrapolation error under distribution shift. Tailored exploration strategies remain largely unexplored, with most works relying on $\epsilon$-greedy, which is ineffective in long-horizon settings. Moreover, there are relatively few systematic comparisons across algorithmic classes under consistent settings, with many studies relying on feature engineering and heuristic baselines, which limits the available evidence on generalization and transferability. Large-scale problems are often simplified by restricting control to bottleneck tools or localized agents rather than learning scalable, system-wide policies \cite{sood2024supporting,stockermann2025scalability}. The central challenge is to develop scalable formulations, representations, reward mechanisms, and training strategies that can handle the full temporal and structural complexity of real-world systems, especially semiconductor fabs.



\section{Policy Optimization}
\label{sec:policy}

A central difficulty in RL is exploiting the temporal structure of the underlying decision process \cite{klissarov2025discovering}. In systems such as semiconductor manufacturing, the effect of a local decision may become visible only after a long and highly entangled chain of subsequent actions and equipment interactions \cite{yeganeh2025deep}. System-level KPIs are therefore cumulative outcomes of many interdependent decisions rather than immediate consequences of a single action, creating a severe long-horizon credit-assignment problem. In particular, the system-level feedback is typically observed only over sufficiently large segments encompassing thousands of actions, whereas smaller segments are dominated by stochastic fluctuations. To this end, we formulate an event-driven learning interface for asynchronous, resource-constrained systems, where local actions induce temporally extended and interdependent events, while system-level feedback is delayed and aggregated.

\begin{figure}[h]
    \centering
    \includegraphics[width=0.79\textwidth, trim=3cm 2.5cm 4.5cm 4.5cm,
        clip]{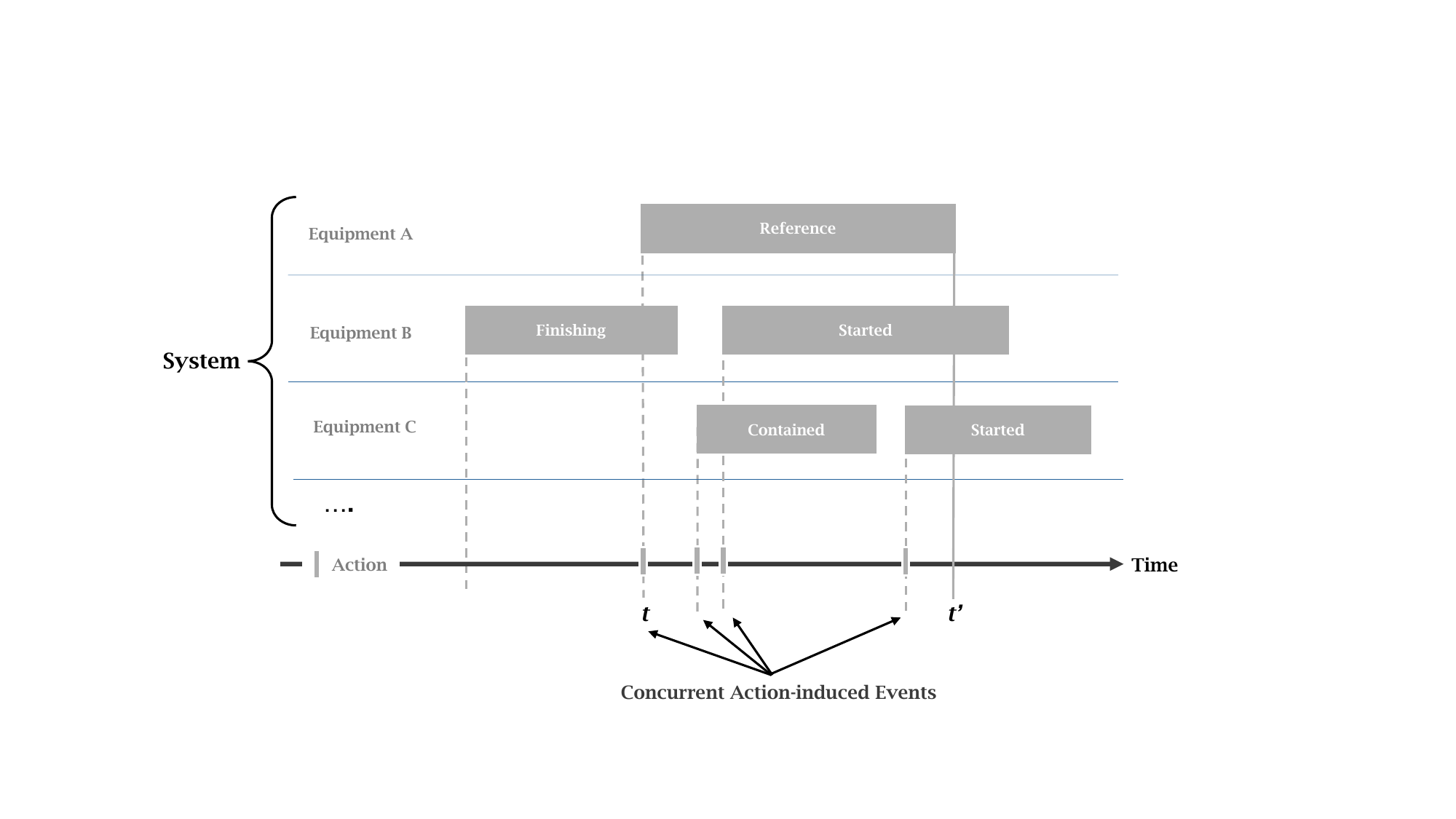}
    \caption{Actions induce discrete, temporally extended events that may overlap across the system.}
    \label{fig:temporal_structure_policy}
\end{figure}

\paragraph{Temporal Structure}
The system differs from standard episodic settings. The problem is naturally continuous and effectively infinite-horizon. Actions trigger events, such as dispatching decisions when equipment becomes available, rather than occurring at fixed time steps. Events may overlap in time. Taking one event as a \emph{reference} (as shown in Fig.~\ref{fig:temporal_structure_policy}), some previously active events may be \emph{finishing}, others may have \emph{started} while it remains active, and some may be \emph{contained} within its span. Because actions initiate events, subsequent actions taken during this span become intertwined with these processes and influence the system trajectory \cite{monch2012production,ghasemi2023deep}. This structure is lost when one considers only a sequence of actions. In particular, the next decision point may correspond to a different event and thus be uninformative for credit assignment. This view is also aligned with \emph{temporally extended} actions and abstractions studied in semi-Markov decision processes (SMDPs) and hierarchical RL \cite{machado2023temporal,kopp2022hierarchical}. The same temporal structure appears in related discrete-event control settings where actions consume resources over time and affect delayed aggregate performance.
These considerations motivate a continuing \emph{event-driven} MDP, $M=(\mathcal{S},\mathcal{A},P,r)$ with state space $\mathcal{S}$, adaptive feasible action sets $\mathcal{A}$, transition kernel $P$, and reward function $r$. For credit assignment, it is more natural to define transitions $(s,a,r,s')$ with the temporal extent of the same action-induced event rather than with the next arbitrary decision. This motivates a \emph{system-coupled} view in which overlapping events that affect overall performance are not treated as independent samples but instead contribute jointly to temporal-difference updates. In this setting, the action-value function $Q$ remains useful because it preserves action-dependent information while supporting event-level credit assignment.

\paragraph{Formulation}
Two return objectives are relevant for continuing control. Under discounted evaluation, the value of a stationary policy $\pi$ is
\begin{equation}
V_\gamma^\pi(s)
=
\mathbb{E}^\pi\!\left[
\sum_{t=1}^{\infty}\gamma^{t-1} r(s_t,a_t)
\,\middle|\, s_1=s
\right]
\label{eq:v}
\end{equation}
which optimally, 
\[
V_\gamma^*(s)=\max_\pi V_\gamma^\pi(s).
\]
An alternative is the long-run average reward,
\begin{equation}
\rho^\pi(s)
=
\lim_{T\to\infty}
\mathbb{E}^\pi\!\left[
\frac{1}{T}\sum_{t=1}^{T} r(s_t,a_t)
\,\middle|\, s_1=s
\right].  
\label{eq:rho}
\end{equation}
The average-reward formulation is arguably natural for non-terminating fab system control \cite{wan2021average}. However, to remain compatible with standard deep offline and online RL algorithms, we instantiate the tailored methods below primarily in discounted form. This choice preserves algorithmic compatibility while still enabling long-horizon reasoning through event-driven and groupwise temporal-difference constructions.

In discounted formulations, the discount factor $\gamma$ controls the effective planning horizon. Larger values of $\gamma$ better capture long-term consequences, but as $\gamma \rightarrow 1$, learning becomes more difficult: bootstrapped targets depend on increasingly distant returns, approximation errors accumulate, and Q-values may become unstable. This trade-off is especially acute in delayed systems, which require stable learning and long-horizon planning. These observations motivate temporal-difference models that couple interrelated events and propagate reward at the group level rather than only through isolated transitions.

\subsection{Control Architecture}


We consider a centralized agent that controls the entire system at the most fine-grained decision level—dispatching—to optimize global objectives. This represents the most complex setting while remaining applicable to more abstract levels. Rather than designing a multi-agent system with coordination and equilibrium formulations~\cite{yang2018mean}, we focus on a central policy with parameter sharing, enabling a more fundamental and scalable formulation.

\begin{figure}[h]
  \centering
      \includegraphics[
        width=\textwidth,
        trim=0cm 1.5cm 1cm 2cm,
        clip
    ]{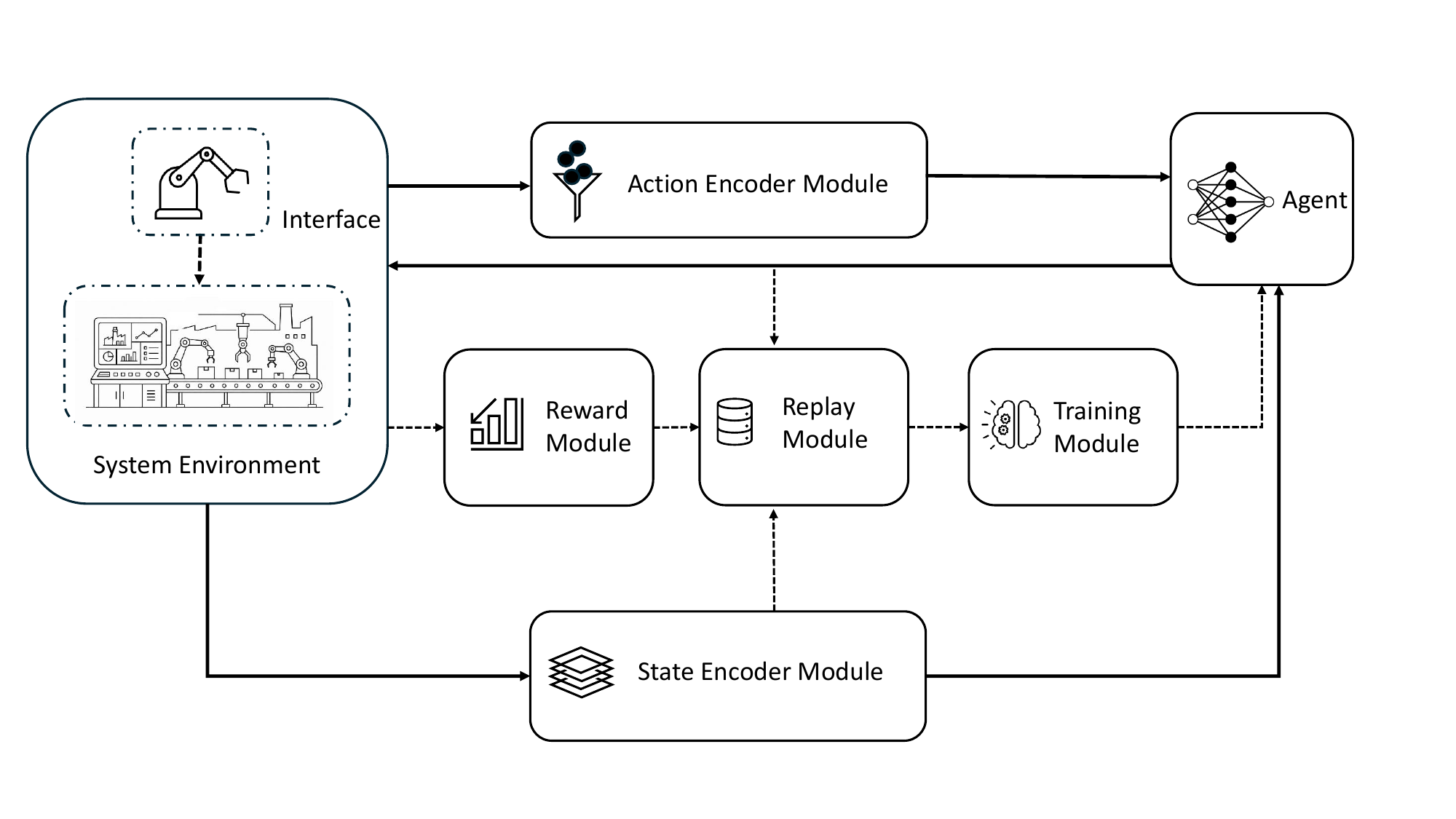}
   \caption{Overview of the modular control architecture for policy interaction with the environment through encoder modules (solid arrows) and auxiliary modules (dashed arrows) for policy optimization.}
  \label{fig:rl_framework}
\end{figure}

To support efficient environment interaction and policy optimization framework, we propose a modular architecture in Fig.~\ref{fig:rl_framework}. The policy directly selects candidates, requiring appropriate state ($s$) and action ($a$) representations to sample from $\pi_\phi(a \mid s)$. As the number of candidates can reach thousands, we adopt a score-based softmax policy:
\[
\pi_\phi(a_i \mid s)
= \frac{\exp\!\left(z_\phi(s,a_i)/\eta\right)}
{\sum_{j} \exp\!\left(z_\phi(s,a_j)/\eta\right)},
\qquad
a \sim \pi_\phi(\cdot \mid s),
\]
where \(z_\phi(s,a)\) denotes logits (e.g., Q-values), and temperature \(\eta > 0\) controls exploration. This formulation naturally adapts to varying action set sizes.

State and action encoders enable environment interactions, while policy optimization relies on reward computation, experience replay, and dedicated training modules.


The agent interacts with either a real system or a simulated environment resembling a digital twin \cite{mohammed2026digital,kuo2025deep}. When any equipment becomes available, the agent receives the current state and feasible candidates (lots in fabs), selects an action, and executes it. A reward ($r$) is then computed based on KPI-driven metrics as the system transitions to the next state ($s'$). The resulting transition $(s, a, r, s')$ is stored in replay memory and used to update the policy. This closed-loop design decouples environment interaction, representation learning, reward computation, and optimization, enabling scalable training.

\paragraph{Constrained Policy}

In practice, domain-specific constraints (e.g., corrosion avoidance in semiconductor wafers) must be enforced. We interpret $\pi_\phi(a \mid s)$ as a raw policy and define the executed policy as
\[
\mu_\phi(a \mid s) = C(\pi_\phi(\cdot \mid s), s),
\]
where $C(\cdot)$ enforces feasibility as a hard inductive bias. This guarantees valid actions during execution while the agent is encouraged, via reward, to learn constraint-compliant behavior.

\subsection{Reward Construction}
We construct a scalar reward by aggregating multi-objective feedback to improve system throughput and utilization. In semiconductor fab systems, rewards are inherently delayed and long-horizon and the impact of an action may only become observable after a substantial time lag \cite{yeganeh2025deep}. To mitigate this feedback sparsity, we shape the reward according to the structure of the problem \cite{ibrahim2024comprehensive}. In particular, we decompose the reward into an immediate event-level term and a delayed system-level term:
\begin{equation}
r = r^{e} + r^{g}.
\label{eq:r}    
\end{equation}
This decomposition captures the fact that individual decisions have immediate local effects that are correlated with global objectives, whereas the global consequences of those decisions only emerge through the longer-term evolution of the system.

\paragraph{Event-level}
This reward element provides dense feedback for each decision and measures its local alignment with the global objectives:
\begin{equation}
r^{e} = w_p \, r^{p} + w_i \, r^{i} + w_c \, r^{c},
\label{eq:r_e}
\end{equation}
where \(w_p\), \(w_i\), and \(w_c\) are weighting coefficients that control the relative importance of the different components, which are defined as
\[
r^{p} \propto -T^{p}, \qquad
r^{i} \propto -T^{i}, \qquad
r^{c} \propto -T^{c}.
\]
Here, \(T^{p}\) denotes the processing time, \(T^{i}\) denotes the equipment idle time from the decision instant until processing starts, and \(T^{c}\) denotes the corrosion-violation time associated with the selected event-action pair. This term penalizes long processing times and idle periods, while encouraging decisions that avoid process-sensitive constraint violations. Such a separation also allows us to explicitly incorporate constraints at the decision level, such as corrosion here.

\paragraph{System-level}
Local feedback alone is insufficient to capture overall system performance. We therefore include KPIs that reflect delayed and accumulated behavior across the manufacturing system. Following prior work, these signals are computed over a finite window of length \(t_s\), which emphasizes recent system dynamics while avoiding dependence on the indefinite past \cite{taheri2024active,yeganeh2025deep,sakr2023simulation}. In particular, we consider \emph{move} and \emph{load} indicators defined respectively as
\begin{equation}
x^{m}(t) = \sum_{\tau=t-t_s}^{t} N_{m}(\tau),
\label{eq:x_m}
\end{equation}
\begin{equation}
x^{\ell}(t) = \frac{1}{|\mathcal{M}|\,t_s}\sum_{i \in \mathcal{M}} \int_{t-t_s}^{t} u_i(\tau)\, d\tau.
\label{eq:x_l}
\end{equation}
Here, \(N_{m}(\tau)\) denotes the number of completed wafer moves at time \(\tau\), \(u_i(\tau)\) denotes the binary utilization of equipment \(i\) at time \(\tau\), and \(\mathcal{M}\) is the set of all equipment in the system. Thus, \(x^{m}(t)\) measures recent cumulative production activity, while \(x^{\ell}(t)\) measures the average equipment utilization over the same window.

We then define the system-level reward through temporal differences of these window-based KPIs between time $t$ and a subsequent time $t' > t$:
\begin{equation}
r^{g}(t,t') = w_m \, r^{m}(t,t') + w_l \, r^{\ell}(t,t'),    
\label{eq:r_g}
\end{equation}
where \(w_m\) and \(w_l\) are weighting coefficients that control the relative importance of the two components, which are defined as
\[
r^{m}(t,t') \propto x^{m}(t') - x^{m}(t),
\qquad
r^{l}(t,t') \propto x^{\ell}(t') - x^{\ell}(t).
\]
Accordingly, \(r^{m}(t,t')\) captures changes in production rate, while \(r^{l}(t,t')\) captures changes in average system utilization. Together, these terms provide delayed feedback that links local actions to broader system-level outcomes.

\paragraph{Normalization}
Because the reward components have different units and scales, each term is standardized prior to aggregation. In particular,
\(
r^{p},\; r^{i},\; r^{c},\; r^{m},\; r^{\ell}
\)
are transformed to have zero mean and unit variance.
All reward terms are therefore normalized and combined using weights that can vary across training phases. While the event-level terms are computed for each action, the generic form \(r^{g}(t,t')\) can be evaluated over arbitrary intervals to capture policy-induced changes, enabling the construction of temporal-difference feedback for policy optimization.

\subsection{Event-Group Temporal-Difference Learning}

A natural extension beyond one-step learning is multi-step temporal difference, in which a truncated $n$-step return is bootstrapped from the resulting state \cite{hessel2018rainbow,sutton2018reinforcement}. 
For a decision index \(k\) associated with event start time \(t_k\), we define
\begin{equation}
y_k^{(n)}
=
\sum_{j=0}^{n-1}\gamma^j r^e_{k+j}
+
r^g(t_k,t_{k+n})
+
\gamma^n \max_{a'} Q_{\bar{\theta}}(s_{k+n},a'),
\label{eq:nstep_fab_td}
\end{equation}
where $\bar{\theta}$ denotes the fixed, non-trainable target network parameters \cite{fujimoto2018addressing}. 
The first term accumulates event-level rewards along the trajectory, the second adds the system-level component, and the final bootstrap term estimates the remaining infinite-horizon tail. The corresponding critic loss is
\begin{equation}
\mathcal{L}_n(\theta) = \left(Q_\theta(s_k,a_k)-y_k^{(n)}\right)^2.
\label{eq:nstep_fab_loss}
\end{equation}

This target is appropriate when the local effect of an action is delayed but remains concentrated over a limited sequence of subsequent events. Various methods such as eligibility traces \cite{sutton2018reinforcement} and generalized advantage estimation \cite{schulman2017proximal} improve credit assignment and navigating the bias--variance trade-off over longer horizons. However, very long sequences remain challenging because exponential decays cause vanishing signals and instabilities. Prior return-decomposition work has shown that delayed rewards can make TD correction exponentially slow and Monte Carlo estimates high-variance, motivating reward redistribution for long-horizon credit assignment~\cite{arjona2019rudder}. We therefore introduce an event-group temporal-difference layer that exploits the temporal structure directly while remaining independent of the policy-optimization backbone.

\paragraph{Event-Driven Aggregation}

We model the system as a collection of asynchronously evolving, inter-dependent event--action pairs whose effects are expressed through a shared system-level reward. Rather than updating each event independently, we construct each training update over a sampled time segment
\[
I_h=[t,t+h],
\]
where \(h > 0\) denotes the segment length. We treat all events associated with this segment as a single \emph{group},
\[
 \mathcal{G} \equiv \mathcal{E}(I_h).
\]

Each event $e\in \mathcal{G}$ has a start time $\tau_e^{\mathrm{start}}$, a finish time $\tau_e^{\mathrm{finish}}$, and corresponding boundary states
\[
s_e^{\mathrm{start}}=s(\tau_e^{\mathrm{start}}), \qquad
s_e^{\mathrm{finish}}=s(\tau_e^{\mathrm{finish}}).
\]
We partition $\mathcal{G}$ into two subsets. The first is the set of events contained within the segment that both start and finish within it:
\[
\mathcal{E}_{\mathrm{contained}}(I_h)=
\left\{
e \,\middle|\,
\tau_e^{\mathrm{start}} \ge t,\;
\tau_e^{\mathrm{finish}} \le t+h
\right\}.
\]
The second contains \emph{started-only events}, which start within the segment but finish after it:
\[
\mathcal{E}_{\mathrm{started}}(I_h)=
\left\{
e \,\middle|\,
\tau_e^{\mathrm{start}}\in I_h,\;
\tau_e^{\mathrm{finish}} > t+h
\right\}.
\]
Thus,
\[
\mathcal{G}=\mathcal{E}_{\mathrm{contained}}(I_h)\cup \mathcal{E}_{\mathrm{started}}(I_h).
\]

For contained events, both boundary states are observed. For started-only events, the terminal state is outside of the segment, so we use the edge state $s_{t+h}$ as a proxy. This captures all actions initiated within the segment, while the effects of previously selected actions that complete during the interval remain encoded in the state. For each event $e\in \mathcal{G}$, we define the bootstrapped temporal-difference discrepancy
\begin{equation}
\delta_e = Q(s,a)-r^e-\gamma V(s'),
\label{eq:delta}
\end{equation}
where $r^e$ is the partially observable--shaped--event-level reward. The bootstrap term depends on the underlying algorithm, e.g., $V(s')=Q(s',a')$ for SARSA and $V(s')= \max\limits_{a'}  Q(s',a')$ for Q-learning \cite{sutton2018reinforcement}.

The key idea is that system-level performance emerges from the joint contribution of events within $\mathcal{G}$; therefore, reward propagation should also occur at the group level. Accordingly, we optimize the group-wise objective
\begin{equation}
\mathcal{L}_{\mathrm{Agg}}(\theta)=
\left(
\bigoplus_{e\in \mathcal{G}}\{\delta_e(\theta)\}-r^g(I_h)
\right)^2,
\label{eq:lagg}
\end{equation}
where \(\bigoplus\) denotes an aggregation operator over the event-level temporal-difference discrepancies \(\{\delta_e(\theta)\}_{e\in G}\), and $r^g(I_h) \equiv r^g(t, t+h)$ represents the system-level reward over the interval. Here, we use mean aggregation 
\begin{equation}
\mathcal{L}_m(\theta)=
\left(
\frac{1}{|\mathcal{G}|}\sum_{e\in \mathcal{G}}\delta_e(\theta)-r^g(I_h)
\right)^2.
\label{eq:lm}
\end{equation}
Compared to sum aggregation, which is highly expressive, the mean acts as a smoothing operator that improves optimization stability. In contrast to averaging per-event squared errors, $\mathrm{mean}((Q_i - y_i)^2)$, this objective matches aggregate system performance and enables end-to-end propagation of global reward signals across interdependent events, akin to reward modeling.

\paragraph{Weighted Aggregation}

An alternative aggregation accounts for the fact that longer events occupy resources for a larger fraction of the interval and may exert a greater influence on subsequent system dynamics. To capture this asymmetry, we introduce a weighted aggregation based on Eq.~\ref{eq:lagg}. For a sampled event $e_0$, let $\mathcal{E}(e_0)$ denote the group of event-action pairs overlapping its interval, and let $\Delta_e$ be the duration of event $e$ within that interval. We define normalized weights
\[
\omega_e(e_0)
=
\frac{\Delta_e}{\sum_{e' \in \mathcal{E}(e_0)} \Delta_{e'}}~,
\qquad
\sum_{e \in \mathcal{E}(e_0)} \omega_e(e_0)=1.
\]
The weighted objective then becomes:
\begin{equation}
\mathcal{L}_w(\theta)
=
\left(
\sum_{e \in \mathcal{E}(e_0)}\omega_e(e_0)\,\delta_e(\theta)-r^g(e_0)
\right)^2,
\label{eq:lw}
\end{equation}
where 
$
r^g(e_0)
\equiv
r^g(\tau_{e_0}^{\mathrm{start}},\tau_{e_0}^{\mathrm{finish}}).
$
This is similar to SMDP-style updates that connect actions through their temporal extent \cite{sutton1999between}.

\subsection{Algorithmic Integration}

The aggregation layer is algorithm-agnostic: it specifies how delayed system-level feedback is assigned to temporally overlapping action-induced events, independently of the policy-optimization backbone. The proposed framework therefore integrates standard RL backbones with temporal-difference constructions tailored to delayed, event-driven, and tightly coupled system dynamics, providing a shared credit-assignment interface across offline and online training. Individual algorithms primarily differ in how value functions and policies are parameterized, updated, and regularized. Offline methods support conservative and data-efficient (pre-)training, while online methods enable further improvement through constrained exploration and stable policy refinement.

\subsubsection{Offline Algorithms}

 In the offline phase, the policy is learned from a fixed dataset $\mathcal{D}$ without further environment interaction \cite{nair2020awac,kumar2020conservative}. This setting is particularly attractive in fab systems, where exploration is risky, costly, and operationally disruptive. It also enables policy pre-training from logged system data before online fine-tuning. The core challenge is to improve beyond the behavior policy in the data without direct environmental feedback~\cite{kumar2020conservative}. This gives rise to the offline model-selection problem since surrogate objectives such as Bellman error may not correlate with downstream control performance \cite{zitovsky2023revisiting}. Distributional shift is another major issue: if the learned policy selects actions that are rare or absent in $\mathcal{D}$, Q-values can become inaccurate and over-optimistic \cite{kostrikov2021offline,haarnoja2018soft}. This problem is exacerbated by the \emph{deadly triad} of function approximation, bootstrapping, and off-policy learning; therefore, conservative learning is essential \cite{van2018deep,sutton2018reinforcement}.

\paragraph{Q-Learning with Conservatism}

Double Q-Learning (DQL) \cite{van2016deep} provides a direct value-based baseline and is well suited to the event-driven critic formulations above. It uses the Bellman backup
\begin{equation}
    Q_i(s,a) = r + \gamma \, Q_j\big(s', \arg\max_{a'} Q_i(s', a')\big),
\quad i,j \in \{1,2\},\ i \ne j ,
\label{eq:dql}
\end{equation}
which can still be risky offline because the maximizing action $a'$ may be unsupported, leading to overestimation. In practice, offline methods therefore constrain the learned policy to remain close to the behavior policy \cite{kumar2020conservative}. The Q-function implicitly defines the policy through greedy or $\varepsilon$-soft action selection.

Conservative Q-Learning (CQL) \cite{kumar2020conservative} addresses this issue by augmenting the temporal-difference objective with a regularizer that lowers the values of unsupported actions. This discourages extrapolation to out-of-distribution actions and improves offline stability. We employ the discrete variant, which can be implemented either as pure Q-learning or in an actor--critic form related to discrete SAC \cite{christodoulou2019soft}. The conservative regularizer further encourages a gap between in-distribution and out-of-distribution actions, helping prevent unsupported actions from receiving spuriously large values \cite{kumar2020conservative}.

Implicit Q-Learning (IQL) completely avoids explicit evaluation of out-of-distribution actions by replacing the max backup with a learned state-value function and extracting the policy from in-dataset advantages \cite{kostrikov2021offline}. Its Bellman target is
\[
Q_{\theta}(s,a) \leftarrow r + \gamma V(s'),
\]
with the value function fitted by expectile regression ($L_2^{\varsigma}$), controlled by a parameter $\varsigma$~\cite{kostrikov2021offline}. This interpolates between SARSA style ($\varsigma = 0.5$) and Q-learning as $\varsigma \to 1$, approaching the supremum over the dataset \cite{kostrikov2021offline}:
\begin{equation}
    \mathcal{L}_V(\psi) = \mathbb{E}_{(s,a)\sim \mathcal{D}}
    \left[L_2^{\varsigma}\big(Q_{\bar{\theta}}(s,a) - V_{\psi}(s)\big)\right].
\label{eq:iqlv}
\end{equation}
This reduces bootstrapping through unsupported actions. The policy is then extracted via advantage-weighted regression \cite{kostrikov2021offline, nair2020awac},
\[
\mathcal{L}_{\pi}(\phi) = \mathbb{E}_{(s,a)\sim \mathcal{D}} \left[
\exp\big(\beta (Q_{\bar{\theta}}(s,a) - V_{\psi}(s))\big)
\log \pi_{\phi}(a \mid s)
\right],
\]
where $\beta$ is a temperature parameter.

All the three offline algorithms are combined with the proposed event-driven group aggregation and domain constraints, such as batch-machine coupling, before gradient updates. Prior work on D4RL shows that long-horizon, sparse-reward tasks such as \emph{AntMaze} are particularly informative for comparing offline algorithms \cite{fu2020d4rl}. Because these tasks emphasize delay and trajectory composition rather than imitation of expert demonstrations, they are well aligned with the long-horizon credit-assignment challenges studied here.

\subsubsection{Online Algorithms}

Online interaction remains valuable because it allows the agent to adapt beyond the support of the offline dataset. In fab systems, however, exploration must be principled and cautious, since large deviations during learning can directly affect operational KPIs. For Q-based methods, we use a Boltzmann temperature-controlled soft policy for exploration \cite{cesa2017boltzmann}. For actor--critic methods such as soft actor--critic (SAC), we use maximum-entropy regularization \cite{haarnoja2018soft}.

Although SAC was originally proposed for continuous actions, discrete SAC extends it to finite action spaces \cite{christodoulou2019soft}. The entropy-regularized objective is 
\begin{equation}
J_{\mathrm{SAC}}(\pi)
=
\mathbb{E}_{\pi}
\left[
\sum_t r(s_t,a_t)
+
\alpha \mathcal{H}(\pi(\cdot\mid s_t))
\right],
\label{eq:sac_j}
\end{equation}
where $\mathcal{H}$ is the entropy operator and $\alpha$ balances exploration \cite{haarnoja2018soft}. In the discrete setting, the induced soft value is
\[
V(s)
=
\alpha \log \sum_a \exp\!\left(\frac{Q(s,a)}{\alpha}\right),
\]
making the connection between entropy-regularized actor--critic learning and soft Q-learning explicit. Additionally, $\alpha$ is adaptively optimized to attend to target entropy $\bar{H}$,
\begin{equation}
J(\alpha) = \mathbb{E}_{a_t \sim \pi_t} \left[ -\alpha \left( \log \pi_t(a_t \mid s_t) + \bar{H} \right) \right].
\label{eq:alpha}
\end{equation}
An important advantage of SAC and Q-learning variants is off-policy learning, which allows reuse of historical data collected under different policies. This improves sample efficiency and provides a natural bridge from offline pre-training to online fine-tuning. 

\paragraph{On-Policy Learning.}
Another major class of model-free methods is policy-gradient learning, with REINFORCE as the generic foundation \cite{sutton2018reinforcement}. Direct policy optimization, however, often suffers from high variance and unstable gradients. This motivated trust-region methods such as TRPO \cite{schulman2015trust}, followed by the clipped surrogate objective of PPO \cite{schulman2017proximal}, which restricts overly large policy updates. These methods enable smoother and more stable optimization and scale well to higher-dimensional settings.

Policy-gradient updates are typically driven by an advantage estimate,
\begin{equation}
A(s,a)=Q(s,a)-V(s),
\label{eq:adv}
\end{equation}
with policy objective
\[
\mathcal{L}_{PG}(\phi) = \mathbb{E}_t \left[ \log \pi_\phi(a_t \mid s_t)\, A_t \right].
\]
Thus, in our framework, PPO is integrated primarily through the construction of the advantage signal, while the clipped policy-optimization backbone remains unchanged. In practice, PPO is often combined with GAE \cite{schulman2017proximal,mnih2016asynchronous,schulman2015high}, which is less aligned with our very long-delay setting. When PPO is implemented with both \(Q\) and \(V\) critics, the standard advantage in Eq.~\ref{eq:adv} can be used directly. Otherwise, we construct a tailored group-relative surrogate by separating the shared system-level reward from the critic-implied return. For each event \(i \in \mathcal{G}\), we define
\begin{equation}
\tilde{A}_i
\triangleq
\delta_i - r^{g}.
\label{eq:adv_sur}
\end{equation}
This decomposition-based credit-assignment heuristic removes the shared broadcast reward from the critic-implied return, leaving a signal that emphasizes the event-level contribution within each coupled group (see Appendix~\ref{sec:surrogate_advantage} for the derivation). Conceptually, this is related to group-relative policy optimization, where improvement is measured relative to a group baseline rather than a global state-value baseline \cite{shao2024deepseekmath}. In practice, the two advantage constructions allow PPO-style updates to remain compatible with the same grouped temporal-difference structure used by the critics.

\subsection{Stability in Policy Learning}

Policy optimization can be sensitive to approximation error, distributional mismatch, and instability under delayed rewards. To better stabilize bootstrapping, we adopt the standard twin-critic and target-network design used in modern actor--critic methods. Let
\[
Q_{\theta_1}(s,a), \qquad Q_{\theta_2}(s,a)
\]
denote two critics. The bootstrapped target is based on their minimum,
\[
y
=
r+\gamma \min\left(
Q_{\bar{\theta}_1}(s',a'),
Q_{\bar{\theta}_2}(s',a')
\right).
\]
which reduces positive bias from noisy function approximation and greedy selection. With target parameters $\bar{\theta}$, the target critics are updated by Polyak averaging,
\[
\bar{\theta}
\leftarrow
(1-\kappa)\bar{\theta} + \kappa \theta,
\]
where $\kappa$ is the target update rate. This slows target drift and makes optimization more stationary \cite{fujimoto2018addressing}. The same design is also compatible with all the algorithms, where double critics and slow targets improve stability.

\section{Experiments}
\label{sec:experiments}


We systematically evaluate the proposed framework for policy optimization in semiconductor fabs, used here as a representative real-world instance of event-driven complex adaptive systems, to address three key questions:
whether it can efficiently train policies that improve system-level operational performance, whether the proposed
formulation enhances temporal credit assignment in the long-horizon system, and how different model-free
algorithms behave across training stages. Using high-fidelity simulations based on real industrial data, we assess policy quality through fab-level KPIs, including throughput, saturation, and load, and perform targeted analyses of policy-optimization methods, temporal-difference formulations, model selection, exploration, algorithmic combinations, and reward design. Further details on the semiconductor fabs and the use case are provided in Appendix~\ref{sec:fab}.

To ensure a robust evaluation of generalization, we consider 31 real scenarios spanning different work-in-progress (WIP)~\cite{monch2012production} and operational configurations. To mitigate temporal leakage and assess generalization across time, we adopt a temporally separated train-test split that maximizes the time gap between selected training and test scenarios. Accordingly, 70\% of the scenarios (21 of 31) are assigned to the training set and kept fixed across all agents, whereas the remaining 30\% (10) are reserved for testing.

\subsection{Encoding}

Effective task-driven optimization requires the agent to form an implicit model of the system dynamics. Recent evidence suggests that even model-free agents acquire such internal world models \cite{richens2025general}. This makes state and action encoding central to policy quality. Given the modular nature of our framework, the encoder can incorporate either predefined features or learned representations within an end-to-end pipeline or from pretrained networks. In this study, we adopt feature-based encodings to keep the focus on the RL backbone. 

\paragraph{State Encoding.}

The state representation provides a system-level snapshot at each decision step (see Appendix~\ref{sec:fab} for information about fabrication systems and the use case). It includes workload statistics and equipment-status information across the fab. Because the environment contains a high mix and a variable number of wafer processing steps, we encode the distribution of WIP across process-flow segments (i.e., sectors) and wafer product categories. This is motivated by prior work showing that workload balance is critical for fab performance \cite{lee2002manufacturing}. We further include aggregate process-step indicators and equipment-state summaries, yielding a compact fixed-size representation of balance, resource availability, and production context.

\paragraph{Action Encoding.}

Similarly, for candidate wafer lots, feature vectors capture product, sector, technology, quantity, processing time, position, batchability, sequence number, setup status and constraint-related attributes (e.g., corrosion). This retains essential information while remaining computationally tractable. A key challenge is representing a large number of jobs (up to tens of thousands). We address this by using processing time as a compact proxy, combined with other lot-level features (e.g., product type and process sequence), enabling the agent to learn process characteristics effectively.

\paragraph{Pre-processing} 

During encodings, categorical variables are one-hot encoded, and numerical variables are normalized to $\mathcal{N}(0,1)$. Processing time is first log-transformed and then normalized. The resulting feature dimension exceeds 500 per action and 5000 for the state, totaling over 5500 for the policy input. For reward, to ensure consistent scaling across heterogeneous reward terms, we estimate normalization statistics from trajectories generated by a random policy. We retain events over 90 shifts following policy interaction to mitigate transient effects, and fix the resulting means and variances to standardize all reward components to zero mean and unit variance during training. Event-level rewards are normalized at the granularity of individual events, whereas global components are computed over temporal horizons with these adaptive event lengths, ensuring consistency with their underlying temporal structure.

\subsection{Training}

Training was conducted in two phases: an offline phase using data collected under a random policy and an online phase in which either random or pretrained policies were optimized using simulator interaction. To better approximate practical deployment conditions, we constrained the amount of data available for learning in both phases, particularly during online interaction. Specifically, the training budget was limited to interaction corresponding to several weeks of fab operation, equivalent to 90 shifts for offline and 50 shifts for online.

\begin{figure}[!h]
  \centering
  \begin{subfigure}[t]{0.48\linewidth}
    \centering
     \includegraphics[width=\linewidth]{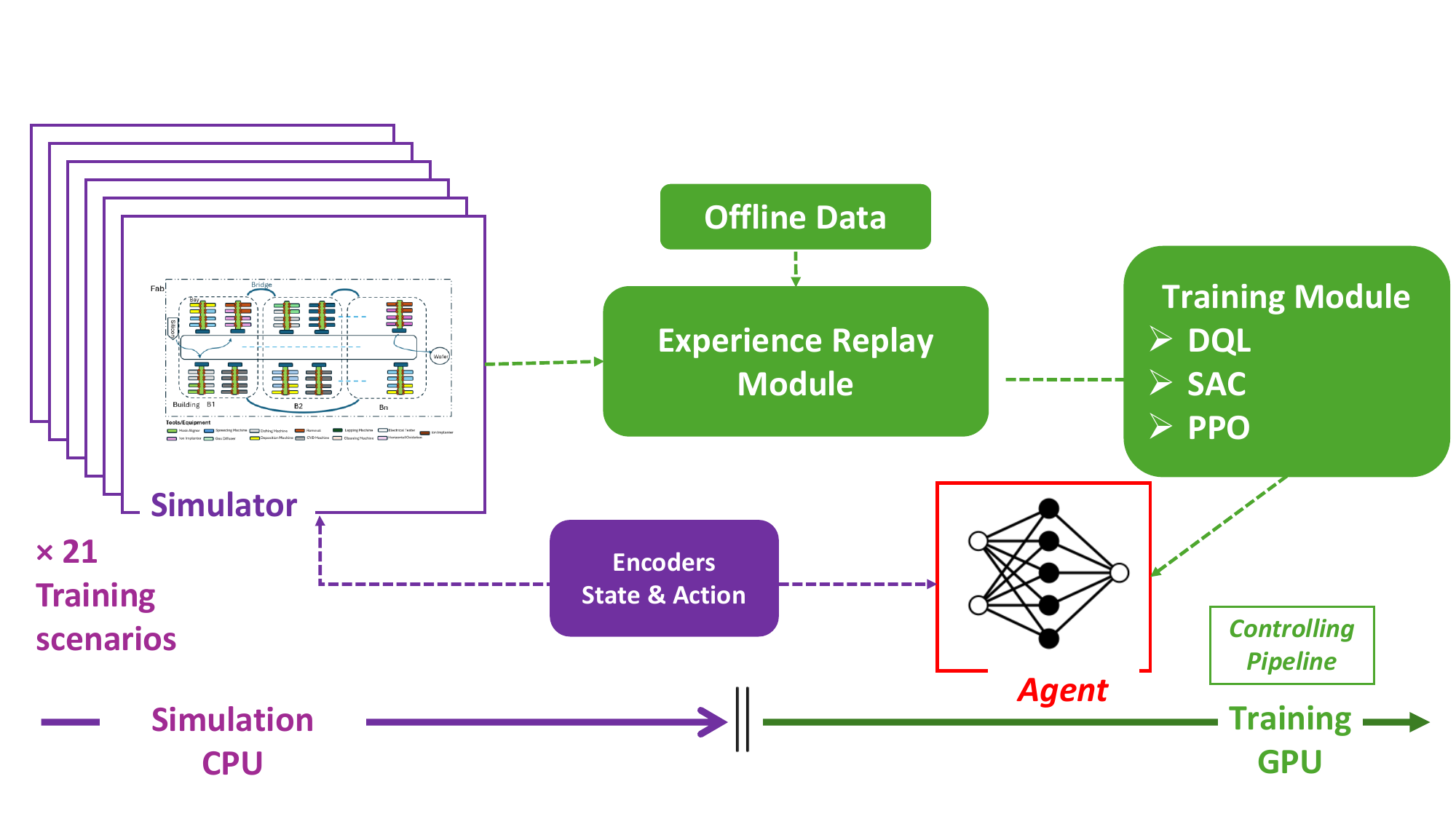}
    \caption{Offline.}
    \label{fig:offline_pipeline}
  \end{subfigure}\hfill
  \begin{subfigure}[t]{0.48\linewidth}
    \centering
    \includegraphics[width=\linewidth]{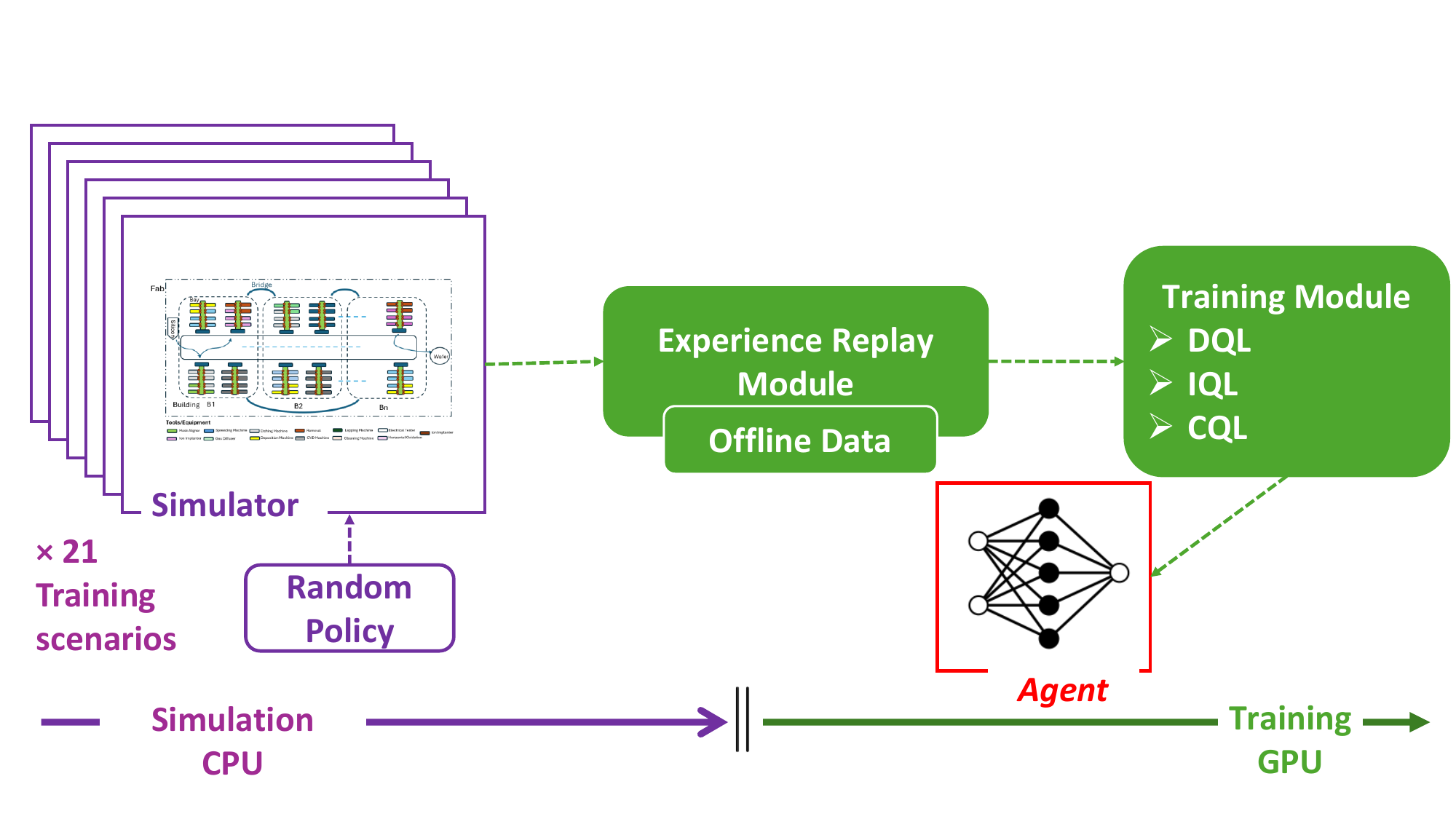}
    \caption{Online.}
    \label{fig:online_pipeline}
  \end{subfigure}
  \caption{Training pipelines, with simulations on CPU and training on GPU.}
  \label{fig:training_pipelines}
\end{figure}

The overall training pipelines are illustrated in Fig.~\ref{fig:training_pipelines}. In the offline setting, training was limited to 100,000 gradient steps. In the online setting, interaction was restricted to 50 training shifts, corresponding to a total budget of 15,000 gradient steps across all agents. The temporal aggregation scheme also differed between stages. During offline training, the aggregation horizon was adaptive and defined by sampling individual events with only event-level (shaped~\cite{ibrahim2024comprehensive}) reward. During online training, it was fixed to 360-minute segments, which were sampled from replay for off-policy methods such as DQL and SAC, and aligned with the most recent interaction data for the on-policy method PPO. To further reflect realistic computational constraints, we also imposed a maximum GPU-time budget for all training of 2 days offline and 5 days online.

\subsection{Numerical Evaluation}

We followed a rigorous evaluation protocol consisting of 30 scenario–seed evaluation instances, obtained from 10 independent test scenarios and three initialization seeds per scenario. This helps obtain an aggregated estimate of each agent's performance over diverse operating conditions. All experiments were conducted using a high-fidelity simulator based on real industrial data. For referencing performance, the random policy is included as a no-control, shortest processing time (SPT) is included as a dispatching heuristic aligned with the processing-time component of the reward, and first-in-first-out (FIFO) is used as the deterministic baseline for computing relative KPI gains and aggregating performance across test scenarios. See Appendix~\ref{sec:valid} for details of the evaluation procedure, performance-metric computation, and statistical reporting.

\subsubsection{Offline}
\label{subsubsec:offline}

\begin{figure*}[!h]
    \centering
    \begin{subfigure}[t]{0.49\textwidth}
        \centering
        \includegraphics[width=\textwidth]{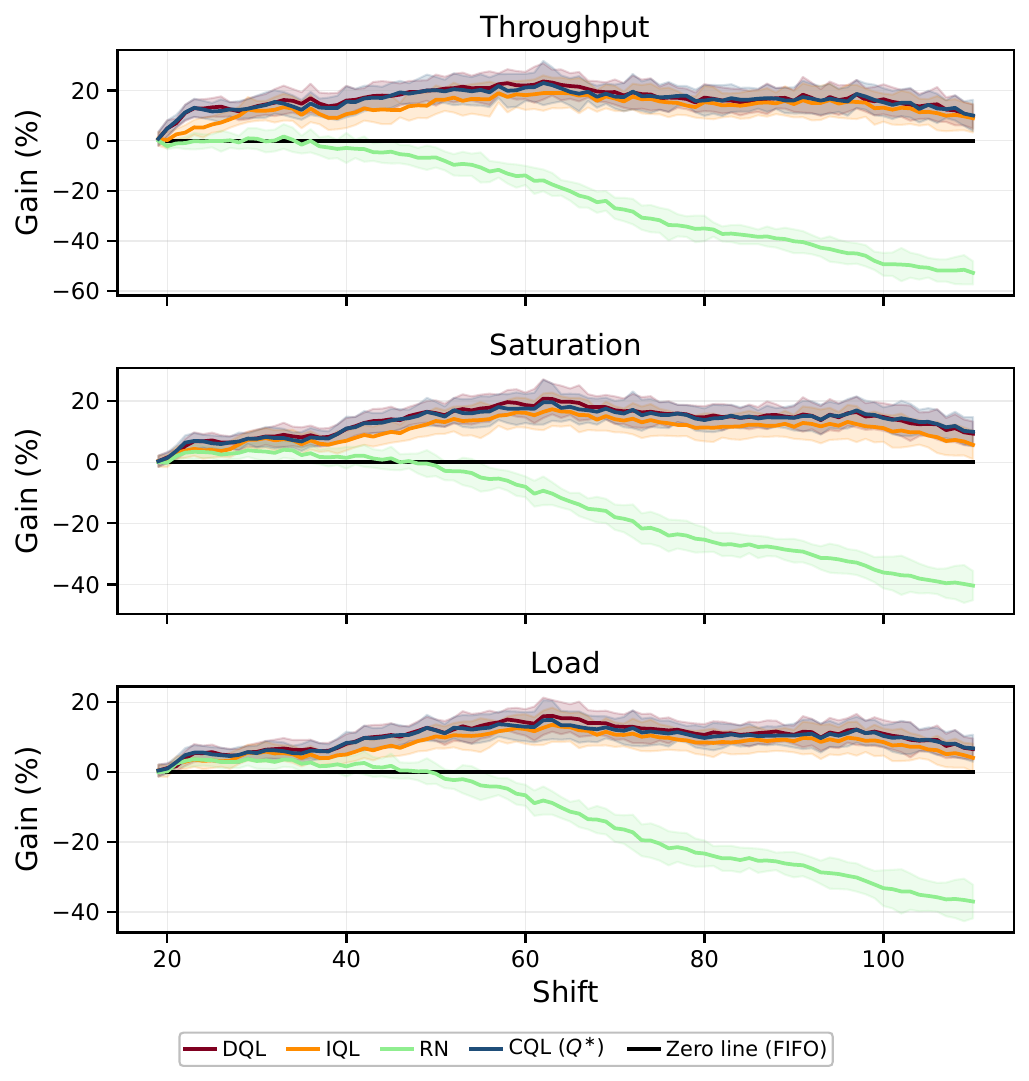}
        \caption{Aggregated gains against the Random policy.}
        \label{fig:rn-combined-kpi}
    \end{subfigure}
    \hfill
    \begin{subfigure}[t]{0.49\textwidth}
        \centering
        \includegraphics[width=\textwidth]{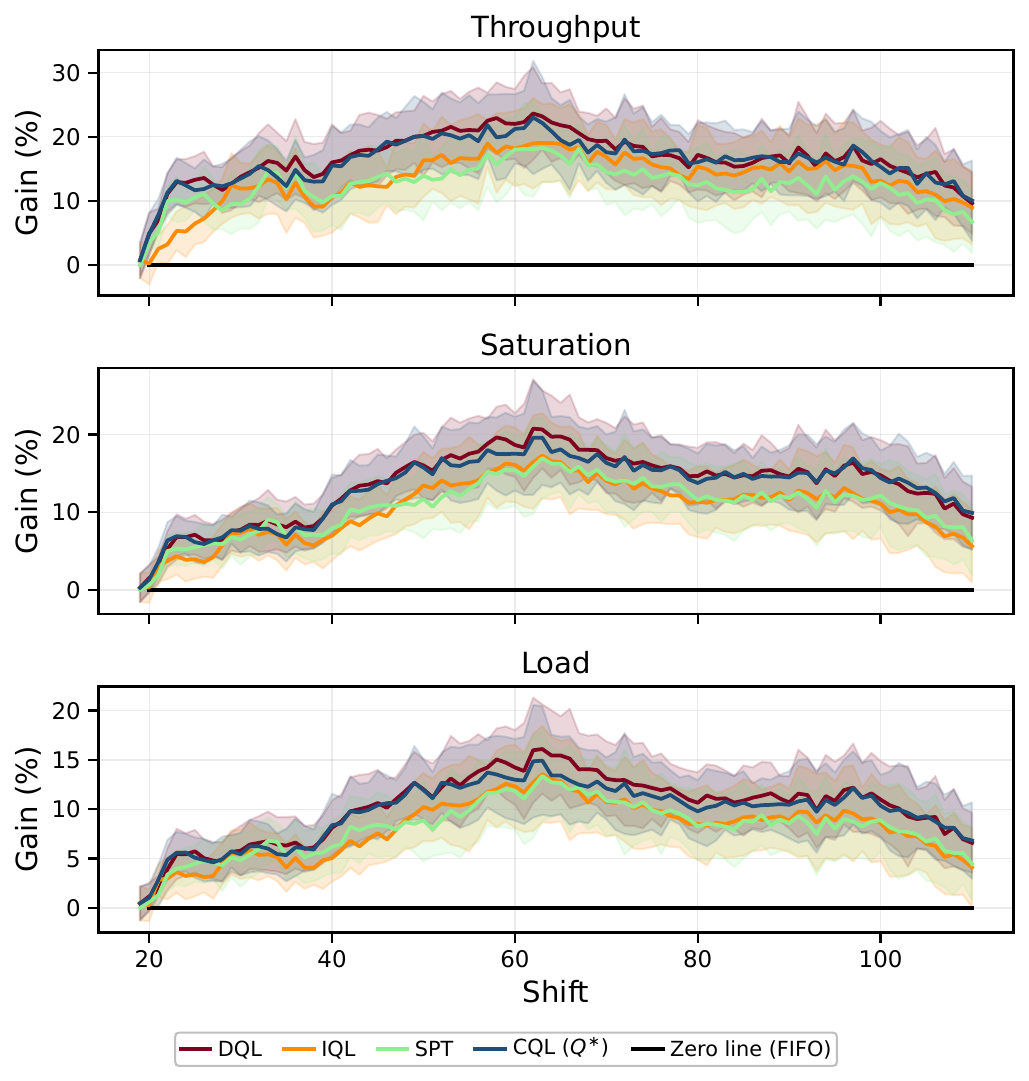}
        \caption{Aggregated gains against the SPT policy.}
        \label{fig:offline_spt-combined-kpi}
    \end{subfigure}
    \caption{Comparison of KPI gains for offline agents relative to the FIFO baseline against the random and SPT policies. The black horizontal line denotes the FIFO reference. Colours identify the policies as shown in the in-panel legends. Solid lines denote means and shaded regions denote Bonferroni-adjusted 95\% confidence intervals across evaluation runs.}
    \label{fig:offline_spt-rn-side-by-side}
\end{figure*}

We consider a challenging offline control setting in which the dataset is collected under the random policy, and training relies only on event-level reward components. Because random-policy trajectories remain relatively stable, provide only weak behavioral supervision, and include no system-level temporal-difference reward, this setting makes policy optimization particularly difficult. Offline experience collected from the training scenarios yields approximately 7 million transitions, which are used to train agents for 100,000 gradient steps with the event-driven mean aggregation objective in Eq.~\ref{eq:lm}. Figure~\ref{fig:offline_spt-rn-side-by-side} and Table~\ref{tab:20_offline_agents} summarize the performance of the offline agents relative to the FIFO as a deterministic baseline, alongside the Random and SPT reference policies. For CQL, we report the Q-learning variant, denoted as CQL(\(Q^*\)); See Appendix~\ref{sec:offline_performance_scenarios} for their comparison across different scenarios and Appendix~\ref{sec:cql_Variants} for comparison among CQL variants. 

\begin{table}[!h]
\caption{KPI gains for offline agents relative to FIFO. Values are percentage gains in throughput, saturation and load, reported as mean $\pm$ Bonferroni-adjusted 95\% confidence-interval across evaluation runs.}
\label{tab:20_offline_agents}
\centering
\begin{tabular}{lccc}
\toprule
\textbf{Agent} & \textbf{Throughput} & \textbf{Saturation} & \textbf{Load} \\
\midrule
Random & $-21.2 \pm 3.0$ & $-12.3 \pm 2.6$ & $-10.8 \pm 2.4$ \\
SPT & $13.9 \pm 4.1$ & $13.3 \pm 3.5$ & $10.7 \pm 2.9$ \\
\midrule
IQL & $14.7 \pm 4.3$ & $13.0 \pm 3.8$ & $10.6 \pm 3.1$ \\
CQL($Q^*$) & $17.5 \pm 6.3$ & $15.7 \pm 5.0$ & $12.3 \pm 4.1$ \\
DQL & $18.0 \pm 5.3$ & $16.0 \pm 4.2$ & $12.8 \pm 3.5$ \\
\bottomrule
\end{tabular}
\vspace{2mm}
\end{table}

\paragraph{TD Formulation} 
\label{sec:offline_td}

To examine the effect of temporal-difference formulation, we compare four alternatives built from the same event-sampled segments. First, we consider a \textit{truncated discounted-sequence} objective, corresponding to the standard \(n\)-step discounted target in Eq.~\ref{eq:nstep_fab_td}. Second, we evaluate a \textit{stacked event-wise} objective, in which events within a segment and their rewards are treated independently, analogous to a per-event squared TD loss, \(\mathrm{mean}\!\left((Q_i-y_i)^2\right)\). Third, we consider an \textit{event-averaged} objective, in which rewards are averaged within the segment while the value terms are computed only from the parent event states \(v(s)\) and \(v(s')\). Finally, we evaluate the proposed \emph{event-aggregated} objective (Eq.~\ref{eq:lm}) by using the mean of event-level rewards as a single group-level target, skipping individual event rewards. We compare these four formulations under both IQL and DQL to isolate the effect of temporal modeling and credit assignment. See Appendix~\ref{sec:td_abl_iql} for IQL performances under different hyperparameter settings.

\begin{table}[!h]
\caption{KPI gains for different TD formulations for selected offline agents. Values are percentage gains in throughput, saturation and load relative to FIFO, reported as mean $\pm$ Bonferroni-adjusted 95\% confidence-interval across evaluation runs. Higher values indicate larger gains.}
\label{tab:kpi_td_by_agent}
\centering
\footnotesize
\begin{tabular}{llccc}
\toprule
\textbf{Agent} & \textbf{TD formulation} & \textbf{Throughput} & \textbf{Saturation} & \textbf{Load} \\
\midrule
\multirow{4}{*}{DQL}
 & Truncated discounted-sequence & $-3.4 \pm 3.6$ & $0.4 \pm 3.2$  & $0.9 \pm 2.7$ \\
 & Event-averaged                & $5.6 \pm 5.8$  & $5.7 \pm 4.3$  & $4.5 \pm 3.6$ \\
 & Stacked event-wise            & $16.2 \pm 5.3$ & $14.8 \pm 4.6$ & $11.9 \pm 3.8$ \\
 & Event-aggregated              & $18.0 \pm 5.3$ & $16.0 \pm 4.2$ & $12.8 \pm 3.5$ \\
\midrule
\multirow{4}{*}{IQL}
 & Truncated discounted-sequence & $-5.1 \pm 6.0$ & $-1.2 \pm 4.2$ & $-0.5 \pm 3.6$ \\
 & Event-averaged                & $13.2 \pm 4.3$ & $12.4 \pm 3.6$ & $10.3 \pm 3.1$ \\
 & Stacked event-wise            & $10.1 \pm 4.1$ & $9.8 \pm 3.6$  & $8.3 \pm 3.1$ \\
 & Event-aggregated              & $14.7 \pm 4.3$ & $13.0 \pm 3.8$ & $10.6 \pm 3.2$ \\
\bottomrule
\end{tabular}
\vspace{2mm}
\end{table}

\paragraph{Offline model selection}
\label{sec:ofm}

\begin{figure}[!h]
    \centering
    \includegraphics[trim=0 0 0 0,clip,width=0.65\textwidth]{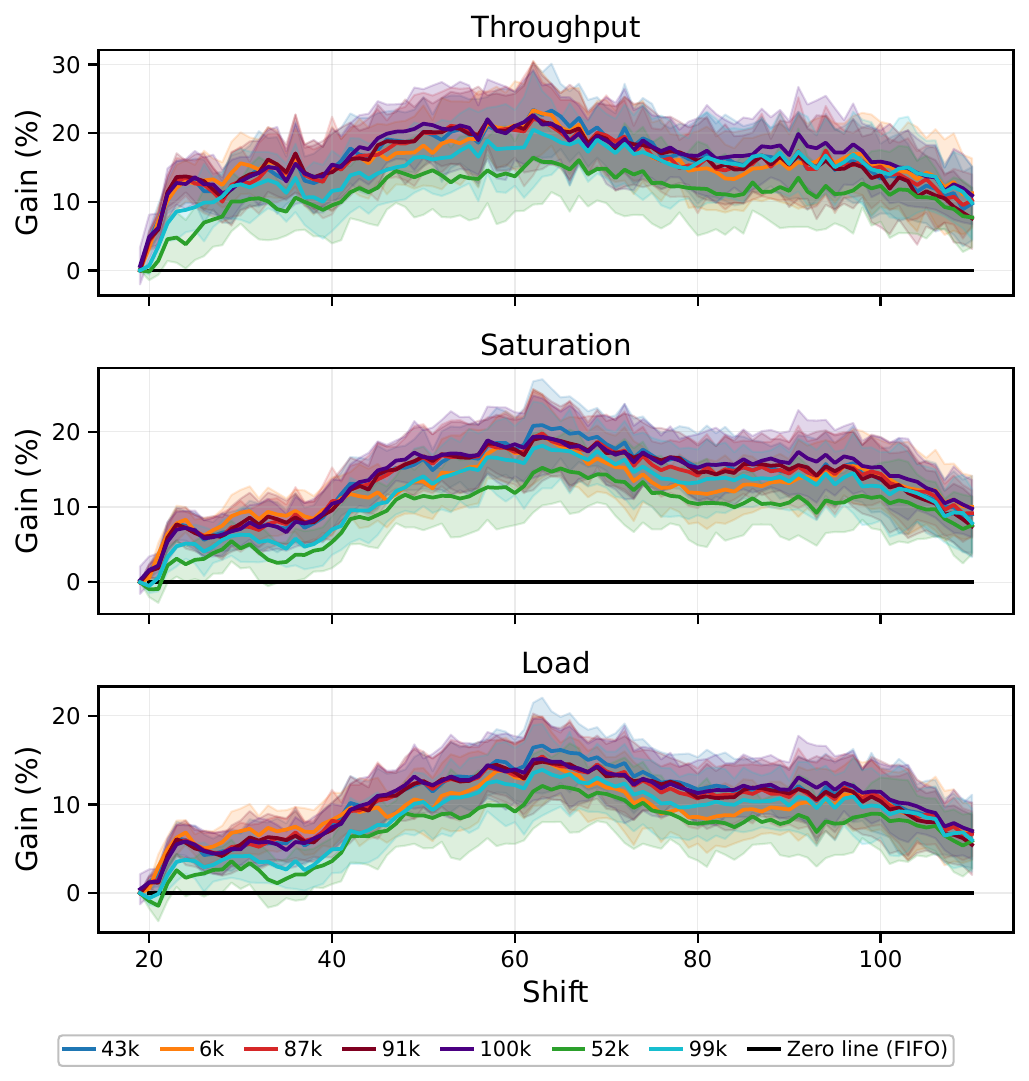}
    \caption{KPI gains (\%) of selected offline-trained DQL checkpoints relative to FIFO across shifts. The black horizontal line denotes the FIFO reference. Colours identify the policies as shown in the in-panel legends. Solid lines show means and shaded regions show Bonferroni-adjusted 95\% confidence intervals.}
    \label{fig:all_selected_checkpoints_online}
\end{figure}

To study offline model selection, selected checkpoints from the offline training trajectory were evaluated by comparing offline TD loss with test-time performance across different KPIs. As summarized in Table~\ref{tab:offline_model_selection_ckpt_all}, the relationship between TD loss and policy performance is weak and non-monotonic. This analysis was intended to assess whether TD loss can serve as a proxy for downstream policy quality. Checkpoint-wise KPI trajectories and the TD-loss curve over training are provided in Appendix~\ref{sec:app_offline_model_selection}, with Fig.~\ref{fig:td_loss_selected_checkpoints} showing TD loss and Fig.~\ref{fig:all_selected_checkpoints_online} showing KPI gains during the test.

\begin{table}[!h]
\caption{KPI gains and TD loss for selected offline DQL checkpoints. Values report the average TD loss during offline training and percentage KPI gains relative to FIFO. KPI gains are reported as mean $\pm$ Bonferroni-adjusted 95\% confidence-interval across evaluation runs.}
\label{tab:offline_model_selection_ckpt_all}
\centering
\begin{tabular}{ccccc}
\toprule
\textbf{Iteration} & \textbf{TD loss} & \textbf{Throughput} & \textbf{Saturation} & \textbf{Load} \\
\midrule
6,000  & 1.8 & $17.2 \pm 6.2$ & $14.9 \pm 5.0$ & $11.9 \pm 4.3$ \\
43,000  & 3.9 & $17.3 \pm 5.2$ & $15.9 \pm 4.3$ & $13.0 \pm 4.1$ \\
52,000  & 4.7 & $12.6 \pm 5.6$ & $11.7 \pm 4.6$ & $9.6 \pm 3.9$ \\
87,000  & 6.4 & $17.1 \pm 5.7$ & $15.7 \pm 4.7$ & $12.6 \pm 3.9$ \\
91,000  & 4.7 & $17.1 \pm 5.5$ & $15.7 \pm 4.3$ & $12.6 \pm 3.6$ \\
99,000  & 4.8 & $15.8 \pm 4.6$ & $14.0 \pm 4.1$ & $11.0 \pm 3.4$ \\
100,000 & 5.8 & $17.9 \pm 5.4$ & $16.2 \pm 4.4$ & $12.9 \pm 3.6$ \\
\bottomrule
\end{tabular}
\vspace{2mm}


\end{table}

\subsubsection{Online}
\label{subsec:online}

In the online setting, agents learn directly from simulator interaction using system-level temporal-difference rewards (Eq.~\ref{eq:r_g}), without event-level reward components; event-level rewards are considered only in the reward-design ablation. Training is therefore more challenging, as feedback must be propagated over long horizons: each sampled segment spans 360 minutes and typically contains thousands of actions, making credit assignment difficult. We evaluate DQL, SAC, and two PPO-style variants initialized from pretrained offline agents. During online fine-tuning, the pretrained policies are softened for exploration. The performance of different online agents is summarized in Table~\ref{tab:online} and Fig.~\ref{fig:online_spt-rn-side-by-side}; see Appendix~\ref{sec:online_performance_scenarios} for their comparison across different scenarios and Appendix~\ref{sec:online_agg} for a comparison of different aggregators in DQL.

\begin{table}[!h]
\caption{KPI gains for online agents relative to FIFO. Values are percentage gains in throughput, saturation and load, reported as mean $\pm$ Bonferroni-adjusted 95\% confidence-interval across evaluation runs.}
\label{tab:online}
\centering
\begin{tabular}{lccc}
\toprule
\textbf{Agent} & \textbf{Throughput} & \textbf{Saturation} & \textbf{Load} \\
\midrule
Random & $-21.2 \pm 3.0$ & $-12.3 \pm 2.6$ & $-10.8 \pm 2.4$ \\
SPT    & $13.9 \pm 4.1$  & $13.3 \pm 3.5$  & $10.7 \pm 2.9$ \\
\midrule
PPO    & $16.3 \pm 6.5$  & $15.3 \pm 4.8$  & $12.1 \pm 3.9$ \\
DQL    & $19.7 \pm 4.8$  & $16.6 \pm 3.9$  & $13.2 \pm 3.4$ \\
SAC    & $20.7 \pm 5.1$  & $17.5 \pm 4.3$  & $13.8 \pm 3.6$ \\
\bottomrule
\end{tabular}
\vspace{2mm}
\end{table}

\begin{figure*}[!h]
    \centering
    \begin{subfigure}[t]{0.49\textwidth}
        \centering
        \includegraphics[width=\textwidth]{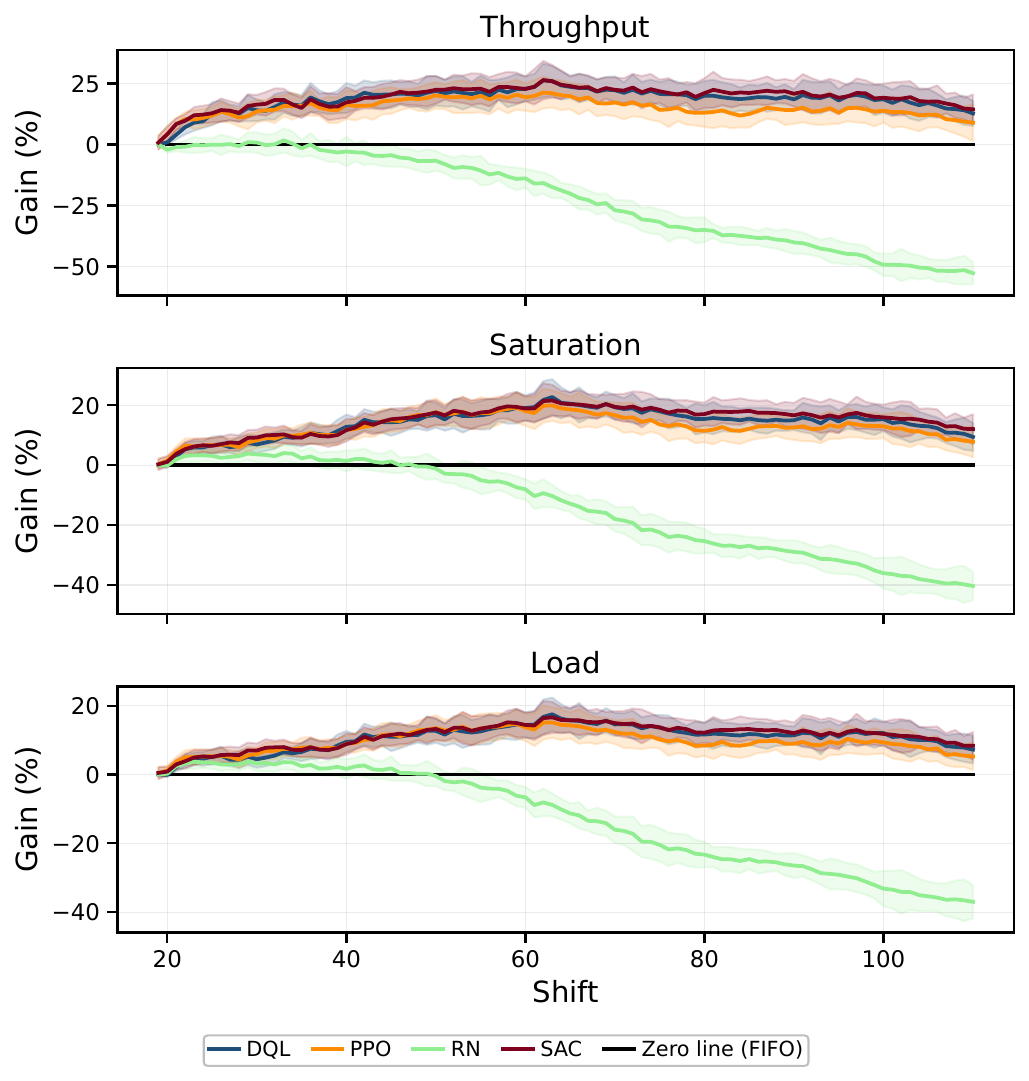}
        \caption{Aggregated gains against the random policy.}
        \label{fig:rn_online-combined-kpi}
    \end{subfigure}
    \hfill
    \begin{subfigure}[t]{0.49\textwidth}
        \centering
        \includegraphics[width=\textwidth]{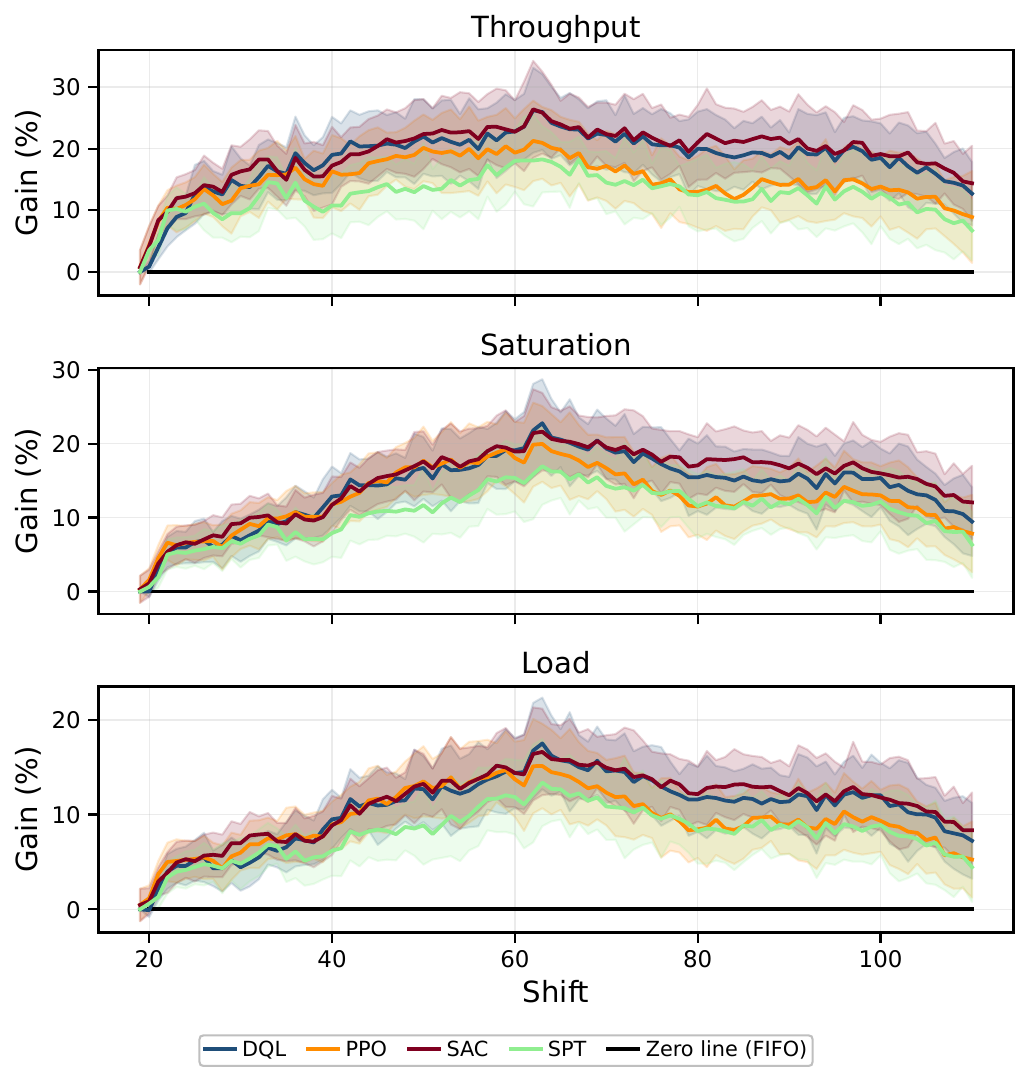}
        \caption{Aggregated gains against the SPT policy.}
        \label{fig:spt_online-combined-kpi}
    \end{subfigure}
    \caption{Comparison of KPI gains for online agents relative to the FIFO baseline against the random and SPT policies. The black horizontal line denotes the FIFO reference. Colours identify the policies as shown in the in-panel legends. Solid lines denote means and shaded regions denote Bonferroni-adjusted 95\% confidence intervals across evaluation runs. }
    \label{fig:online_spt-rn-side-by-side}
\end{figure*}

\paragraph{PPO advantage architecture}
\label{sec:ppo_adv}

We compared two PPO variants on how the advantage is estimated. In the first, the clipped policy objective uses a surrogate advantage computed directly from the conservative twin-critic estimate, \(\min(Q_{\theta_1}, Q_{\theta_2})\), following Eq.~\ref{eq:adv_sur}. In the second, we additionally train a value function, \(V(s)\), using segment-level rewards based on the event-driven aggregation in Eq.~\ref{eq:lm}, and then compute the advantage using the exact form in Eq.~\ref{eq:adv}. As summarized in Table~\ref{tab:ppo_advantage_variants}, the exact-advantage achieves better performance.

\begin{table}[!h]
\caption{Effect of PPO advantage formulation on KPI gains. The first block reports percentage KPI gains relative to FIFO (first block) as mean $\pm$ Bonferroni-adjusted 95\% confidence-interval across evaluation runs. The second block is reported values indicate mean differences, and Holm–Bonferroni-adjusted $p$-values compare exact and surrogate advantages for each KPI using different statistical tests ($n = 10$).}
\label{tab:ppo_advantage_variants}
\centering
\begin{tabular}{llccc}
\toprule
\textbf{Category} & \textbf{Entry} & \textbf{Throughput} & \textbf{Saturation} & \textbf{Load} \\
\midrule
\multirow{2}{*}{Variant}
& Surrogate advantage & $13.9 \pm 5.9$ & $13.7 \pm 4.4$ & $10.8 \pm 3.9$ \\
& Exact advantage     & $16.3 \pm 6.5$ & $15.3 \pm 4.8$ & $12.1 \pm 3.9$ \\
\midrule
\multirow{2}{*}{Test}
& Mean difference
& \(2.386\) & \(1.586\) & \(1.285\) \\
& Permutation \(p\)-value
& \(5.859 \times 10^{-3}\) & \(5.859 \times 10^{-3}\) & \(5.859 \times 10^{-3}\) \\
& Wilcoxon \(p\)-value
& \(5.859 \times 10^{-3}\) & \(5.859 \times 10^{-3}\) & \(5.859 \times 10^{-3}\) \\
\bottomrule
\end{tabular}
\vspace{2mm}
\end{table}

\paragraph{Exploration}
\label{sec:exp}

\begin{figure}[!h]
    \centering
    \includegraphics[width=0.65\linewidth]{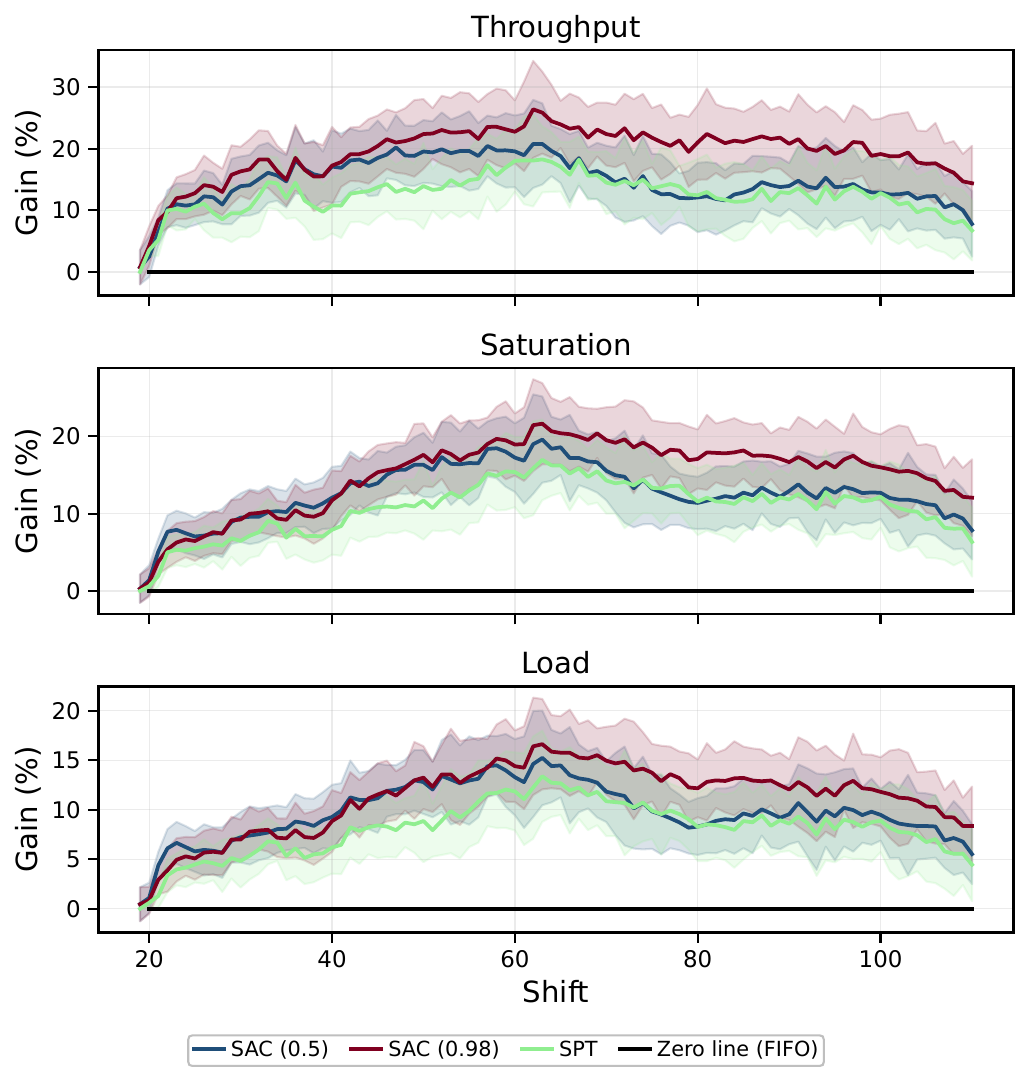}
    \caption{Comparison of KPI gains for two SAC agents with different target-entropy coefficients, relative to the FIFO baseline against SPT. The black horizontal line denotes the FIFO reference, whereas burgundy and dark blue correspond to SAC with \(c_h=0.98\) and  SAC with \(c_h=0.5\) respectively. The light green curve represents the SPT.Solid lines denote means and shaded regions denote Bonferroni-adjusted 95\% confidence intervals across evaluation runs.}
    \label{fig:online_spt-rn-side-by-side-twosac}
\end{figure}

In SAC, exploration is controlled through the target entropy, which determines the strength of entropy regularization. In our implementation, the target entropy is defined as
\[
-\, c_h \log\!\left(\frac{1}{|\mathcal{A}|}\right),
\qquad c_h \in \{0.98,\,0.5\},
\]
and the temperature parameter is adapted according to Eq.~\ref{eq:alpha}. Larger values of \(c_h\) encourage higher policy entropy and  therefore, stronger exploration during training. We compare two SAC variants with \(c_h=0.98\) and \(c_h=0.5\). As summarized in Table~\ref{tab:sac_exploration} and illustrated in 
Fig.~\ref{fig:online_spt-rn-side-by-side-twosac}, the higher-entropy setting consistently outperforms the lower-entropy setting across all three KPIs.

\begin{table}[!h]
\caption{Effect of SAC target entropy on KPI gains. The first block reports percentage KPI gains relative to FIFO as mean $\pm$ Bonferroni-adjusted 95\% confidence-interval across evaluation runs. The second block reports mean differences and Holm--Bonferroni-adjusted permutation and Wilcoxon signed-rank $P$ values comparing $c_h = 0.98$ and $c_h = 0.5$ for each KPI using $n = 10$ paired test scenarios.}
\label{tab:sac_exploration}
\centering
\begin{tabular}{llccc}
\toprule
\textbf{Category} & \textbf{Entry} & \textbf{Throughput} & \textbf{Saturation} & \textbf{Load} \\
\midrule
\multirow{2}{*}{Variant}
& \(c_h = 0.5\)  & $16.1 \pm 5.6$ & $15.2 \pm 4.4$ & $12.4 \pm 3.7$ \\
& \(c_h = 0.98\) & $20.7 \pm 5.1$ & $17.5 \pm 4.3$ & $13.8 \pm 3.6$ \\
\midrule
\multirow{3}{*}{Test}
& Mean difference
& \(4.604\) & \(2.277\) & \(1.447\) \\
& Permutation \(p\)-value
& \(2.930 \times 10^{-3}\) & \(2.930 \times 10^{-3}\) & \(2.930 \times 10^{-3}\) \\
& Wilcoxon \(p\)-value
& \(5.859 \times 10^{-3}\) & \(5.859 \times 10^{-3}\) & \(5.859 \times 10^{-3}\) \\
\bottomrule
\end{tabular}
\vspace{2mm}
\end{table}

\paragraph{Reward design}
\label{sec:reward}
We further study the effect of reward design in the twin-Q PPO variant, when not pretrained, as summarized in Table~\ref{tab:ppo_reward}. We compare training with both event-level (shaped) and segment-level rewards against training with segment-level rewards only.
\begin{table}[!h]
\caption{Effect of reward design on plain PPO performance. Values are percentage KPI gains relative to FIFO for the random policy and for plain PPO trained with event-plus-segment rewards or segment-level rewards only, reported as mean $\pm$ Bonferroni-adjusted 95\% confidence-interval across evaluation runs.}
\label{tab:ppo_reward}
\centering
\begin{tabular}{lccc}
\toprule
\textbf{Reward elements} & \textbf{Throughput} & \textbf{Saturation} & \textbf{Load} \\
\midrule
Random          & $-21.2 \pm 3.0$ & $-12.3 \pm 2.6$ & $-10.8 \pm 2.4$ \\
\midrule
Event + segment & $14.0 \pm 4.2$ & $14.0 \pm 3.2$ & $11.6 \pm 2.9$ \\
Segment only    & $15.9 \pm 3.9$ & $15.4 \pm 3.0$ & $12.3 \pm 2.4$ \\
\bottomrule
\end{tabular}
\vspace{2mm}
\end{table}
Additional scenario-wise plots for the segment-only PPO setting are reported in Appendix~\ref{app:ppo_segment_only_plots}, where the same configuration is compared against the Random and SPT reference policies across throughput, saturation, and load.

\FloatBarrier 
\section{Discussion}
\label{sec:discussion}


The results highlight the effectiveness of the proposed framework for policy optimization across different settings. We evaluated the agents on 30 industry-level test instances over 90 shifts. In the offline setting, training was performed without interaction and relied only on data generated by a random policy across separate scenarios. This is important from both safety and data-efficiency perspectives, as it enables policy learning without interfering with the live system. Even under these constraints, all algorithms achieved substantial percentage-point gains, with up to 39\% in throughput and 28\% in saturation compared with the random policy (no control). Results also underline the importance of offline model selection \cite{zitovsky2023revisiting}, which helps extract a policy at its best stage of training.

In the online setting, interaction with the environment allowed the agent to learn from system dynamics and achieve further improvements. Relative to the no-control setting, performance percentage-point gains reached approximately 42\% in throughput and 30\% in saturation. We also analyzed the impact of control on the system state, particularly the balance of the manufacturing system and how it changed after applying the learned policies (see Appendix~\ref{sec:sector_analysis}). This analysis further demonstrates the effectiveness of the trained policies. Notably, the framework improved all three KPIs associated with the multi-objective goal of increasing production and utilization. In addition, training remained computationally feasible on GPU due to the modular and pipeline-oriented design, demonstrating the scalability and operational efficiency of the proposed formulation.

\paragraph{Formulation}

The proposed approach addresses the main challenges of the problem, namely delayed feedback and long-horizon decision making. In the offline setting, we employed shaped rewards~\cite{ibrahim2024comprehensive} for each event (action). However, the results in Section~\ref{sec:offline_td}, where the truncated discounted-sequence approach performed poorly, highlight the importance of the credit assignment issue in the long-horizon \cite{yeganeh2025deep}. This suggests that event-level rewards are meaningful, as reflected in the stacked results, which effectively treat events independently. At the same time, the proposed aggregation method proved helpful by refining event-level information even without a global reward. The results also support the motivation for an average-reward perspective, which is consistent with the long-run system objective. In the online setting, event-level rewards were not used. Instead, training relied on scalar rewards assigned to 360-minute segments that covered many actions (often thousands). Through the aggregation mechanism, these coarse rewards were propagated to individual action-event pairs, enabling effective policy improvements, as evidenced by the results. This is also clearly shown in Table~\ref{tab:ppo_reward}: even without a pretrained agent and using only global reward training, the method achieved strong performance after just 7,500 gradient steps. Including the event reward also performed worse there, although it can accelerate online training in some cases.

\paragraph{Algorithmic Perspective}

The results obtained with different algorithms tailored to the proposed formulation provide several insights. Rather than focusing on exhaustive tuning of algorithm-specific settings, this study investigates the proposed framework and how RL conceptualization can support effective and scalable policy optimization. A strict comparison of algorithmic superiority would require careful and method-specific hyperparameter tuning, which is beyond the scope of this work. To maintain fairness, we kept the main hyperparameters and training settings, including learning rates and gradient steps, as consistent as possible across methods.
For offline policy optimization, conservatism is important. CQL and IQL achieved relatively close performance compared to DQL, which is more prone to overestimation. It is worth noting that both CQL and IQL are more complex to train, and IQL involves more network components than DQL, so improved training strategies may further enhance their performance. These findings also reinforce the importance of model selection during offline training. Among the offline methods, the Q-learning version of CQL performed well, while its entropy-regularized version may be a promising candidate for a pretrained high-entropy policy for online fine-tuning with respect to exploration. IQL, which learns a value function via expectile regression \cite{kostrikov2021offline} and performs policy extraction, is more sensitive to hyperparameters than DQL and CQL, but provides stronger theoretical grounding and enables more conservative policy extraction.
In the online setting, off-policy methods, namely DQL and SAC, have higher sample efficiency, whereas the on-policy methods (PPO) are more sensitive to data for optimization. DQL remained the simplest yet effective value-based approach. PPO, on the other hand, provides a stable policy-gradient framework, although its performance depends strongly on architectural choices and advantage estimation. The results suggest that combining Q- and V-based architectures can improve PPO performance. SAC, which combines sample efficiency with entropy regularization, offers a natural formulation for stochastic policy learning and exploration control. This is reflected in its strong performance, particularly when the target entropy was adapted. Keeping the entropy objective higher to encourage exploration resulted in a notable performance gain. Overall, both PPO and SAC show strong theoretical promise, and the results support their applicability within the proposed framework.


\section{Conclusion}
\label{sec:conclusion}

We addressed the core challenge of applying RL to semiconductor manufacturing control in a real large-scale setting. The proposed framework was designed to cope with computational bottlenecks, delayed feedback, long-horizon credit assignment, and the need for adaptive decision-making in a dynamic system. By combining an event-driven formulation, a centralized learning architecture, and a tailored optimization strategy, the framework enables agents to optimize multiple system-relevant KPIs in a high-fidelity digital twin across different algorithms. The numerical results show that the learned policies achieve consistent gains over conventional baselines, confirming that both careful problem formulation and algorithm design are critical for efficient agents. The study also answers the research questions of how reinforcement learning can be formulated in a scalable and effective way for semiconductor production control and applied to similar complex adaptive systems. Overall, the proposed policy-optimization strategy combines standard deep RL backbones with tailored temporal-difference constructions for event-driven, delayed, and tightly coupled system dynamics. Offline training provides conservative and data-efficient pretraining, while online training supports further improvement through constrained exploration and stable learning, resulting in a general and flexible framework that can support diverse optimization objectives under different constraints.

\paragraph{Future Work}

Several methodological directions follow from this work. A limitation of the present study is that state and action representations are feature-based; learned encoders, graph-based representations, and foundation-model-assisted representations may further improve transferability and are left for future work. The framework, therefore can be extended along multiple axes, including representation learning (e.g., graph-based models to better capture system structure), average-reward formulations, stability-enhancing optimization methods, alternative aggregation mechanisms, and improved training procedures and regularization. Systematic hyperparameter tuning and benchmarking are also needed, with additional algorithms adjusted to the formulation. A promising direction is the use of model-based agents, as well as incorporating self-supervised and generative models to better learn system dynamics \cite{hafner2025training,yeganeh2025deep}. While the proposed event-group temporal-difference objective improves empirical credit assignment, it remains a surrogate aggregation objective and requires further theoretical grounding, particularly regarding whether and when it admits, approximates, or preserves a Bellman-consistent fixed point. A key direction for future work is to analyze when such aggregation preserves policy improvement or introduces bias, including bias–variance trade-offs, gradient-dispersion effects, and connections to return decomposition and reward redistribution methods ~\cite{arjona2019rudder}.

From an application perspective, reward design and evaluation remain critical. In particular, studying KPIs in multi-objective settings is important to understand their effects on policy behavior, convergence, and trade-offs among system-level objectives. Since the current study is based on approximately one month of production data, future work should evaluate the framework over longer and more diverse time horizons to improve robustness and generalization across operating conditions. Depending on operational objectives and constraints, additional system-level KPIs can be incorporated into the optimization framework. This is closely related to how training scenarios are constructed and how performance is aggregated; for example, satisfying lot due dates could be explicitly introduced as an additional constraint.

\section*{Data Availability}
The data that were used to support the findings of this study are not publicly available due to their proprietary nature and association with real industrial case scenarios. Access to these data is therefore restricted.

\section*{Code availability}
The code for the reinforcement learning framework, independent of the proprietary data and simulator, is publicly available on GitHub under the MIT License:
\href{https://github.com/YavarYeganeh/EventDrivenTD-RL}{https://github.com/YavarYeganeh/EventDrivenTD-RL}.

\section*{Acknowledgments}
The work presented in this paper was supported by HiCONNECTS, which has received funding from Chips Joint Undertaking under grant agreement number 101097296. The authors also thank Antonio Maria Stivala, Elias El Achkar, Rosario Messina, Antonio Costa, Roberto Corsini, Leonardo Longo, Daniele Vinciguerra, Marco Stefano Scroppo. 

\section*{Declaration}
The authors have no competing interests.
LLMs were used solely for grammatical and language correction, as well as for search assistance.

\bibliographystyle{unsrtnat}
\bibliography{references}  

\newpage
\appendix

\label{sec:append}

\section{Semiconductor Fab}
\label{sec:fab}

The manufacturing environment under consideration is a semiconductor front-end fabrication facility (fab) within the integrated-circuit production chain, distinct from downstream assembly and packaging (back-end). The fab exhibits high operational complexity, as device structures are created through long sequences of technologically constrained and interdependent processing steps on silicon wafers. Consequently, production planning and control (PPC) must operate under strict precedence constraints and limited resource availability, with decisions propagating across a tightly coupled process flow.

The facility consists of interconnected buildings subdivided into areas and bays, where equipment is grouped and typically connected via automated material handling systems. Material transport is performed using both automated and manual mechanisms; transfer times are significant for inter-building moves but generally negligible within local areas. The fab produces a diverse mix of technologies and products, each following a predefined sequence of operations referred to as a \emph{recipe} or process flow. Routing commonality is high within a technology and lower across technologies. Wafers are processed in fixed-capacity carriers (``lots''), including special-purpose lots such as engineering lots, and are assigned internal due dates even when not linked to specific customer orders. A distinction is made between operations (ordered steps in a route) and processes (underlying transformations), where the same process type may appear multiple times within a route.

PPC in such environments is structured as a hierarchical decision process with progressively shorter time horizons, ranging from aggregate planning to real-time shop-floor control~\cite{monch2012production}. Higher levels define feasible production targets and regulate work-in-process (WIP), while lower levels allocate and sequence operations and react to system variability. In particular, dispatching operates at the lowest PPC level, selecting in real time the next job for an available resource, with local decisions potentially impacting global performance.

\subsection{Decision Levels}

\begin{itemize}
    \item Aggregate planning (long-term): translates demand into a feasible master plan and regulates WIP inflow.
    \item Process capacity planning (mid-term): sets throughput targets per process based on the current fab status.
    \item Scheduling (short-term): defines the time-phased allocation and sequencing of operations on equipment.
    \item Dispatching (real-time): selects the next lot for each piece of equipment, adapting to variability.
\end{itemize}

An additional decision layer, equipment-level control (seconds to minutes), executes immediate actions (e.g., batching, chamber assignment) required to run specific processes. This layer lies outside the scope of PPC and is typically managed independently.

This hierarchical structure motivates the use of learning-based control methods at the dispatching level while ensuring alignment with higher-level planning objectives and system constraints.

\newpage
\subsection{Use Case}
\label{subsubsec:use-case}

\begin{wrapfigure}{r}{0.55\linewidth}
  \centering
    \includegraphics[width=\linewidth, trim=3cm 3cm 3cm 3cm, clip]{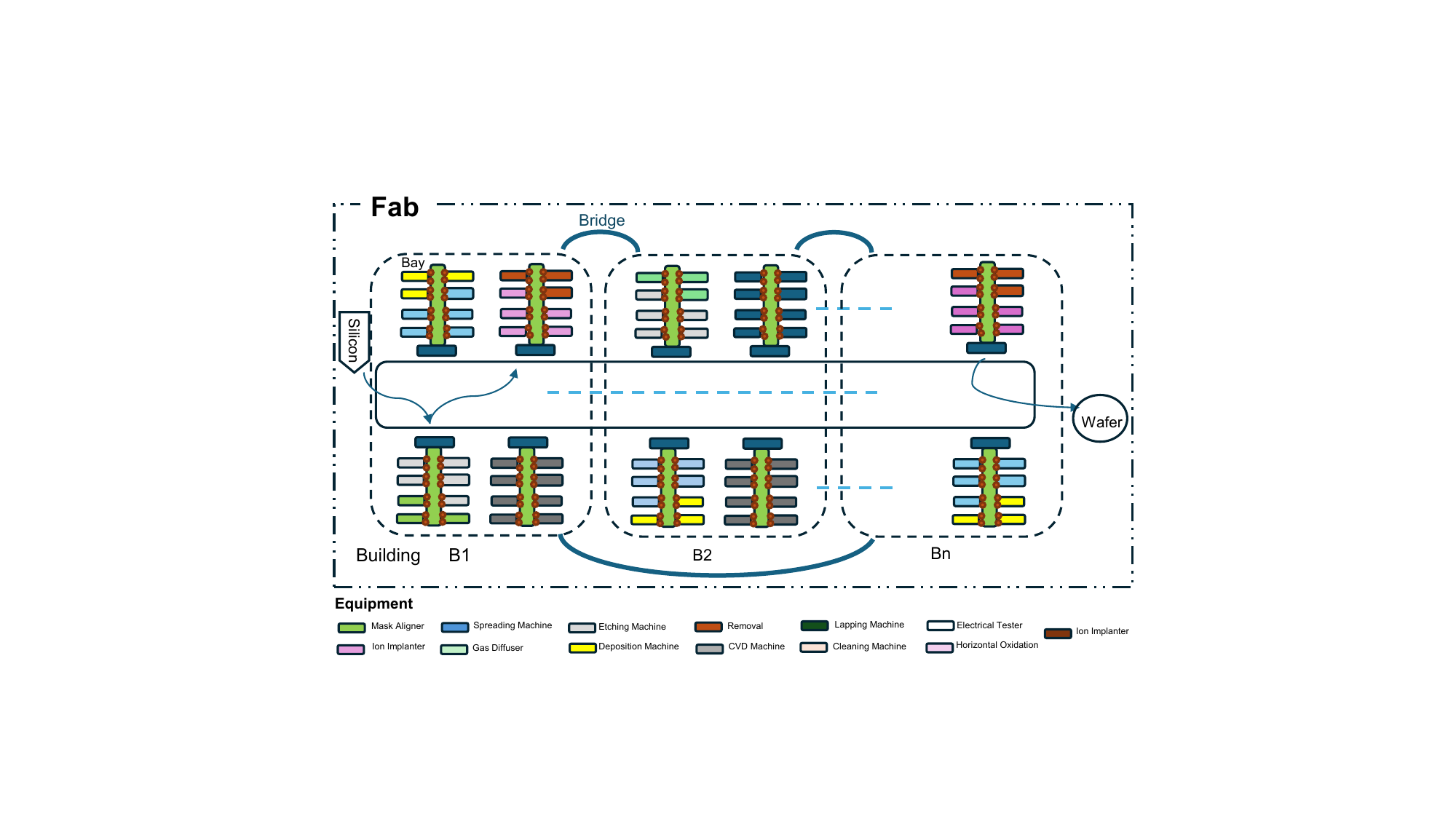}
    \caption{General illustrative layout of the considered semiconductor fabrication facility.}
  \label{fig:sti_layout}
\end{wrapfigure}

The industrial setting considered in this work is a semiconductor front-end fabrication facility within the HiCONNECTS project, inspired by a real STMicroelectronics use case. The fab is organized into multiple interconnected buildings, areas, and bays, where equipment is distributed across different process families and material transport may require non-negligible transfer times. Production is lot-based: wafers are grouped into carriers (lots) that follow predefined process flows and are managed under due-date and cycle-time requirements.

The considered fab operates in a high-mix environment with many technologies and products sharing partially overlapping resources. Process flows are long, technologically constrained, and re-entrant, meaning that the same resource family may be visited multiple times during production. As a result, local dispatching decisions may propagate through the system and affect fab-level performance, including throughput, equipment utilization, and workload distribution.

Production planning and control is structured hierarchically. Higher decision levels define production targets and release decisions, whereas lower levels determine the detailed allocation and sequencing of lots on resources. In this hierarchy, dispatching corresponds to the real-time selection of the next lot for an available resource, while equipment-level decisions such as chamber assignment or batching are handled at a finer operational level. This structure motivates the study of reinforcement learning for dispatching, since the policy must react online to local system conditions while remaining consistent with fab-level operational objectives.

\begin{figure}[!h]
  \centering
    \includegraphics[trim=40 20 40 20,clip,width=0.70\textwidth]{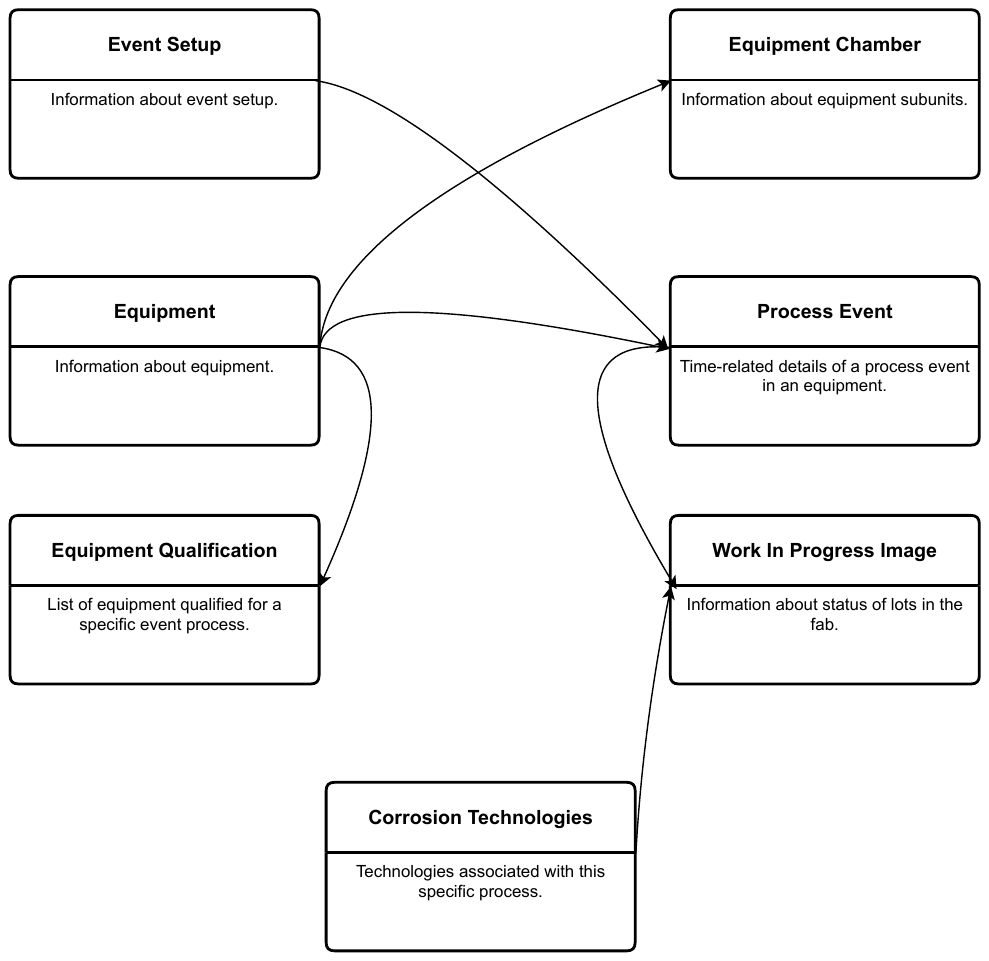}
    \caption{Conceptual ERD of the industrial use case, showing the main entity types and their relationships.}
    \label{fig:erd_conceptual}
\end{figure}

\paragraph{Data representation through a conceptual ERD.}
To summarize the main information objects involved in the industrial use case and their interactions, the system is represented through a conceptual Entity--Relationship Diagram (ERD). In this work, only the conceptual level is used, since it is sufficient to illustrate the main entities and their relationships without introducing implementation-specific database details. The conceptual ERD shown in Fig.~\ref{fig:erd_conceptual} provides a compact representation of the principal information structures underlying the simulator and the dispatching environment.

In this representation, rectangles denote entity types, whereas arrows denote relationships between entities. The ERD includes the main conceptual entities used to describe the production environment as presented in Table~\ref{tab:erd_entities}. These entities capture the core structural information needed to describe process definitions, equipment structure, equipment eligibility, process execution, and fab-level lot status.

\begin{table}[htbp]
\centering
\caption{Entity names used in the conceptual ERD. The table lists the entity types used to structure the industrial fab data model and support feature construction.}
\label{tab:erd_entities}
\vspace{2mm}
\begin{tabular}{l}
\hline
\textbf{Entity Name} \\ \hline
Event Setup \\
Equipment \\
Equipment Qualification \\
Process Event \\
Equipment Chamber \\
Work In Progress Image \\
Corrosion Technologies \\ \hline
\end{tabular}
\end{table}

\newpage
\subsection{Key Performance Indicators}
\label{subsubsec:kpi}
Key performance indicators (KPIs) in semiconductor manufacturing are quantitative measures used to monitor and evaluate the effectiveness of manufacturing operations and associated decision-making processes. Typical KPI families capture performance in areas such as cycle time, yield, equipment utilization, throughput, schedule adherence, manufacturing cost, labor productivity, downtime, and supplier reliability. Systematic measurement of these indicators enables manufacturers to identify improvement opportunities, assess progress over time, and enhance operational efficiency. Ultimately, KPI-driven monitoring supports higher productivity and cost reduction by guiding targeted process optimization.

The KPIs emphasized in practice depend on an organization’s strategic priorities and operational objectives. In semiconductor production, several indicators are particularly common. \textbf{Cycle time} measures the duration required to complete a production cycle from start to end and directly affects production responsiveness and capacity~\cite{waschneck2018deep,liu2022dynamic,kovacs2022customizable,dennis2017lean}. \textbf{Throughput} captures the production rate achieved over a given time horizon and is central to assessing whether the fab sustains productive flow and meets production demand~\cite{lee2022deep,kuhnle2019autonomous,kovacs2022customizable,stevenson2014operations}. \textbf{Equipment utilization} reflects the fraction of available resource time that is effectively used for production~\cite{stricker2018reinforcement,sakr2023simulation,kovacs2022customizable,arnold1998introduction}. Other widely used indicators include \textbf{on-time delivery}, which evaluates adherence to due dates~\cite{liu2022dynamic,stevenson2014operations}; \textbf{production downtime}, which measures interruptions and lost productive time~\cite{kovacs2022customizable,heizer2020operations}; \textbf{setup time}, which captures the effort needed to reconfigure resources for new products or events~\cite{van2004microchip,lee2019simulation,stevenson2014operations}; \textbf{tardiness}, which measures the extent of delayed job completion~\cite{luo2020dynamic,zhou2022reinforcement,nahmias2015production}; \textbf{queue-related measures}, which indicate congestion and waiting pressure in the system~\cite{sakr2023simulation,nahmias2015production}; \textbf{lead time}, which covers the total elapsed time from release to delivery~\cite{stricker2018reinforcement,stevenson2014operations}; and \textbf{yield}, which quantifies the proportion of usable output and is closely linked to process quality and scrap reduction~\cite{jacobs2018manufacturing}.

In the present work, the operational objective is defined primarily in terms of \textbf{throughput}
and \textbf{utilization}. Throughput is used as the main system-level indicator of productive
performance. Utilization is evaluated through two complementary KPIs: \textbf{saturation}, which
captures the instantaneous use of equipment capacity, and \textbf{load}, which measures utilization
over a time window. Together, these indicators describe both the immediate and accumulated use
of available capacity, and are therefore used to assess the effectiveness of the proposed dispatching
policies under capacity-constrained fab operation.

At the same time, the proposed framework is not restricted to these two objectives. Depending on the decision context and the operational priorities of the fab, the same framework can be extended to incorporate additional KPIs, including cycle time, tardiness, due-date adherence, setup-related indicators, queue-related measures, or yield. This flexibility is particularly important for future work, where alternative reward definitions and broader multi-objective formulations may be required.

\subsection{Glossary}
\label{sec:glossary}


\paragraph{Batch:} A group of lots loaded simultaneously in an equipment to be processed all at once, e.g. during diffusion and cleaning processes. 

\paragraph{Chamber:}
An equipment might have several chambers to perform different events (i.e., processes) or speed up a single event. Chambers may all do the same process, even on different products. If one chamber stops working, the others can keep going.

\paragraph{Cycle time:} The time of the entire process flow of a specific product (also named \emph{system time}). In other words, the amount of time elapsed from the release of a lot in the plant (new lot, step $i=1$) to its move out per completion of the recipe (last step $i=N$). It is composed of \emph{processing time}, needed to perform the recipe, \emph{waiting time}, and \emph{transportation time}.

\paragraph{Cluster Equipment:}An equipment cluster in semiconductor manufacturing refers to a group of interconnected equipment that perform multiple sequential processes on semiconductor wafers without needing to manually move the wafers between steps. The cluster equipment system allows for the automatic transfer of a substrate between process modules via a transfer module to deposit different layers using different processes, or to process similar layers in parallel, without exposing the substrates to air before or after processing. 

\paragraph{Configuration:} A physical change (or setup) to the equipment in order to enable it to perform different processes. Configuration changes primarily occur when the process needs to be altered. However, they may also be necessary for different batches requiring distinct processing conditions, such as temperature, pressure, or gas composition for deposition or etching processes. Additionally, changes in the lithography mask necessitate reconfiguring the equipment to properly hold and align the new mask. 

\paragraph{Entity:}
It is equivalent to \textit{equipment}.

\paragraph{Equipment:}
In the semiconductor field, equipment consists of three components or groups of components: a number of process/cleaning/cooling chambers, chambers for loading and unloading, and multiple robots for transferring wafers between chambers. Therefore, a piece of \emph{equipment} refers to a resource, machine, entity, or tool.

\paragraph{Event:}
A manufacturing process in the production of semiconductors. 
It is equivalent to \textit{process}.

\paragraph{Fab:} The semiconductor manufacturing factory.

\paragraph{Load lock:} A container, where the lot is placed, to enable the transfer of wafers from the external environment into the processing chamber's controlled environment. This transfer is done by a robotic arm.

\paragraph{Lot:} A group of wafers (that are placed in a carrier) at a certain step of the product recipe  (i.e., process flow). A lot is considered \emph{on hold}, when its processing (flow) is stopped.

\paragraph{Move out:} A move out generally consists in the completion of an operation for a lot/batch. \textit{1 Move out} is considered (performance) metric, when one wafer moved from one operation to another.

\paragraph{Operation:} An identifier of an event (i.e.,the same event can have different number of operations so if an event happens multiple times it is assigned different operation IDs at each time). 
\paragraph{Process:} It is equivalent to \textit{event}. 

\paragraph{Process flow:} Sequence of operations that must be processed to have the final product (ref. recipe).

\paragraph{Processing time:} The time required to perform a process (event).

\paragraph{Product:} The output of semiconductor manufacturing, specifically denoting the state of wafers after completing the entire process flow. Once wafers undergo all necessary processing steps, they are considered products.

\paragraph{Product Type:} It is determined based on the electronic output (i.e., characteristics) and process flow.

\paragraph{Qualification:} An equipment (or a chamber) is considered qualified to perform an operation over a certain product. 

\paragraph{Recipe :} The product process flow, i.e., the sequence of processes required by a product to be completed.

\paragraph{Seq\_num:} A step ($op=1...N$) in the recipe (process flow) of a product.

\paragraph{Technology:} Semiconductor technology can include different products. This technology is determined based on the electrical output characteristics. Products of the same technology share around 90\% of process flow; products of different technology share around 50\% of process flow.

\paragraph{Tool:} 
It is equivalent to \textit{equipment}.

\paragraph{Wafer:} A thin (disc-shape) slice of semiconductor material, usually silicon, used as the foundation for the fabrication of microelectronic devices.

\paragraph{Wafer Carrier:} Wafer are handled in cassettes or boxes of fixed capacity to avoid wafer contamination or breakages.

\paragraph{WIP:} The WIP is defined as the total number lots in the fab (between step 1 and step $N$-th of the sequence). 

\newpage
\section{Acronym list} \label{sec:acro}
\begin{acronym}[ABCDEFG] 
\acro{AI}{Artificial Intelligence}
\acro{CT}{Cycle Time} 
\acro{CR}{Critical Ratio}
\acro{DDQN}{Double Deep Q-Network}
\acro{DL}{Deep Learning}
\acro{DNN}{Deep Neural Network}
\acro{DRL}{Deep Reinforcement Learning}
\acro{DQN}{Deep Q-Network}
\acro{EDD}{Earliest Due Date}
\acro{ERD}{Entity Relationship Diagram}
\acro{FIFO}{First In First Out}
\acro{FCFS}{First Come First Serve}
\acro{GNNs}{Graph Neural Networks}
\acro{KPIs}{Key Performance Indicators}
\acro{LPT}{Longest Processing Time}
\acro{MDP}{Markov Decision Processes}
\acro{ML}{Machine Learning}
\acro{MOR}{Most Operation Remaining}
\acro{MRT}{Most Remaining Processing Time}
\acro{NN}{Neural Network}
\acro{ODD}{Operation Due Date}
\acro{PPO}{Proximal Policy Optimization}
\acro{RL}{Reinforcement Learning}
\acro{SPT}{Shortest Processing Time}
\acro{SPTSSU}{Shortest sum of Processing Time and Setup Time}
\acro{SR}{Slack Ratio}
\acro{SSU}{Shortest Setup Time}
\acro{TRPO}{Trust Region Policy Optimization}
\end{acronym}

\section{Literature}
\label{sec:appedn.literature}

Semiconductor manufacturing systems are characterized by large scale, stochasticity, and stringent operational constraints, with lots traversing complex, re-entrant process flows across heterogeneous equipment networks. Efficient scheduling and dispatching are critical for achieving high throughput, low cycle time, and stable resource utilization. Despite these requirements, fabs predominantly rely on dispatching heuristics (such as FIFO, SPT, and rules based on due dates or setups) due to their ease of implementation and integration into manufacturing execution system (MES) logic. However, the effectiveness of these heuristics often diminishes in the presence of disruptions, including tool failures, variable processing times, and changes in product mix \cite{stricker2018reinforcement,kuhnle2019autonomous}

Recent research increasingly formulates scheduling and dispatching as reinforcement learning (RL) problems, where decision policies are learned through interaction with discrete-event simulators or from historical data. Although the term \emph{deep RL} is sometimes used broadly to refer to RL with neural function approximation, scheduling applications typically employ specific algorithm families that differ in stability, scalability, and inductive bias. The dominant approaches in the reviewed literature include (i) \emph{value-based} methods derived from Q-learning (such as DQN variants), (ii) \emph{policy-gradient and actor--critic} methods (such as PPO and A3C), and (iii) \emph{multi-agent value-decomposition} methods (such as QMIX) for distributed control.

\paragraph{Value-based methods (DQN, DDQN, D3QN).}
The Deep Q-Network (DQN) family represents the most prevalent approach, utilizing neural networks to approximate the action-value function and deriving policies through greedy or softened action selection. DQN-based controllers are widely applied to semiconductor dispatching and scheduling, frequently incorporating replay and target networks to enhance stability \cite{waschneck2018deep,lee2022deep,sakr2023simulation}. To address value overestimation in large-scale or non-stationary scheduling environments, several studies employ Double DQN (DDQN) \cite{luo2020dynamic,chang2022deep}. For dynamic flexible job shop problems with added complexity, such as sequence-dependent setup times, advanced DQN variants like D3QN have been adopted to further improve learning stability and performance \cite{yan2025deep}. In semiconductor applications, tabular or simplified Q-learning formulations are sometimes used when action spaces are intentionally kept compact to facilitate real-time deployment \cite{stricker2018reinforcement}.

\paragraph{Policy-gradient / actor--critic methods (PPO, A3C).}
When smoother policy updates or increased robustness in optimization are required, scheduling research often adopts policy-gradient and actor--critic methods. Proximal Policy Optimization (PPO) has been implemented in job-shop scheduling and related manufacturing contexts, where stable policy improvement through clipped updates is advantageous \cite{zhang2022dynamic}. More recently, adaptive scheduling frameworks have integrated an A3C backbone with supplementary mechanisms to refine RL design elements. For instance, a large language model (LLM)-guided framework incorporates A3C while dynamically adjusting state and reward definitions based on training feedback \cite{hong2025large}.

\paragraph{Multi-agent and distributed scheduling (QMIX).}
To enhance scalability and facilitate decentralized decision-making, certain studies decompose control across workcenters or shop-floor cells. A representative approach employs QMIX, in which each manufacturing cell is managed by a local agent, and a mixing network aggregates local action-values to optimize a global objective. This structure enables distributed scheduling while maintaining coordination \cite{xing2023distributed}. In this formulation, the decision abstraction is often implemented through weighted composite dispatching rules instead of direct job-to-machine assignments. This approach maintains compact local action spaces while optimizing overall shop-floor performance.

\paragraph{Reward design, multi-objective optimization, and simulation environments.}
In both semiconductor and job-shop research, optimization objectives encompass multiple metrics. Studies address makespan, tardiness, throughput, cycle time, utilization, and other key performance indicators (KPIs), typically through weighted reward components or multi-objective reward formulations \cite{zhou2022reinforcement,lee2019simulation,zhang2024adaptive}. Since fab-level KPIs are delayed and result from numerous interdependent events, shaped intermediate rewards are commonly employed to stabilize learning and enhance credit assignment \cite{sakr2023simulation,stockermann2025scalability}. High-fidelity simulation is crucial for evaluation: configurable simulation platforms and benchmarking environments enable systematic comparisons under realistic stochasticity and re-entrant flows \cite{kovacs2022customizable}. Additionally, explainability modules are increasingly introduced to interpret RL-based dispatching behavior in semiconductor manufacturing \cite{immordino2025explainable}.

\subsection{Problem Formulation in RL-Based Fab Scheduling}
Most reinforcement learning (RL)-based approaches model scheduling and dispatching in semiconductor manufacturing as a discrete-event or event-driven Markov decision process (MDP). In this framework, decisions are triggered at operational epochs such as machine-idle events, lot arrivals, or other discrete control points \cite{waschneck2018deep,kuhnle2019autonomous}. Variations in action abstraction, scope of control, and temporal feedback structure significantly influence scalability and learning stability.

A key distinction lies in the definition of the action space. Some formulations employ fine-grained actions, such as explicit job-to-machine assignments or detailed sequencing decisions. While these designs offer high control resolution, the action space expands rapidly with increased system scale and routing flexibility. Studies on dynamic flexible job shops address this complexity using value-based methods such as Deep Q-Network (DQN) and Double DQN (DDQN) \cite{luo2020dynamic,chang2022deep}, and employ advanced variants like Dueling Double DQN (D3QN) when additional factors, such as sequence-dependent setup times, introduce further instability \cite{yan2025deep}. In these scenarios, tractability is typically achieved through algorithmic stabilizers, such as Double-Q designs, and meticulous state and action engineering, rather than relying solely on generic deep RL approaches.

To manage decision complexity, many studies reformulate scheduling as a dispatching-rule selection problem, where the agent selects from a predefined set of priority rules or rule combinations instead of choosing individual jobs. This abstraction maintains a compact and stable action set, making it well-suited for value-based learning methods such as DQN in manufacturing and fab dispatching contexts \cite{waschneck2018deep,alexopoulos2024deep,marques2025dynamic}. These formulations are particularly attractive for industrial integration because they preserve explicit feasibility logic and can be aligned with operational constraints while supporting adaptive control \cite{kuhnle2019autonomous}.

An intermediate strategy employs priority-based formulations, where policies generate priority scores for jobs or job-machine pairs, and a deterministic selection mechanism determines the final decision, such as through sorting or invalid-action masking. Proximal Policy Optimization (PPO) is a representative optimizer for this class of formulations in job-shop scheduling, offering stable policy updates and gradual adjustments in dispatching behavior \cite{zhang2022dynamic}. Beyond static formulations, recent research investigates adaptive redesign of RL components. For instance, a large language model (LLM)-guided framework integrates an Asynchronous Advantage Actor-Critic (A3C) backbone with dynamic refinement of state and reward definitions based on training feedback \cite{hong2025large}. This approach demonstrates that robustness can be enhanced by modifying the formulation layer surrounding the RL optimizer.

Problem formulation also defines the scope of control allocated to the agent. Some studies focus on optimizing global objectives at the shop or fab level \cite{zhou2022reinforcement,chang2022deep}, while others limit learning-based control to bottleneck tools or selected workcenters to reduce state dimensionality and training costs \cite{stockermann2025scalability}. Another approach decomposes control spatially through multi-agent learning. For example, QMIX facilitates distributed scheduling by assigning local agents to each manufacturing cell and employing a mixing network to align local decisions with a global objective \cite{xing2023distributed}. This comparison underscores a key trade-off: centralized control simplifies coordination but increases representational complexity, whereas multi-agent decompositions enhance scalability but necessitate explicit coordination mechanisms.

Formulation choices are also closely linked to reward design and credit assignment. Global key performance indicators (KPIs), such as throughput and cycle time, are delayed and result from numerous interdependent events. This delay motivates the use of shaped rewards based on intermediate signals or weighted KPI combinations to stabilize learning in event-driven environments \cite{sakr2023simulation,stockermann2025scalability}. Evaluation further depends on simulation fidelity and scenario diversity. Configurable fab simulators enable systematic benchmarking under realistic stochastic conditions \cite{kovacs2022customizable}, and explainability modules have been developed to enhance the interpretability of learned policies \cite{immordino2025explainable}.

Collectively, these studies demonstrate that scalability and performance in RL-based scheduling are influenced as much by problem formulation choices—particularly action abstraction, scope of control, and temporal structure—as by the choice of optimizer. This perspective supports the development of formulations that explicitly integrate event-level decision-making with structured global performance feedback, and it establishes a basis for comparing specific algorithmic frameworks (such as DQN, DDQN, D3QN, PPO, QMIX, and A3C) under consistent evaluation conditions.

\begin{table}[htbp]
\centering
\resizebox{\textwidth}{!}{%
\begin{tabular}{llllll}
\hline
\textbf{Work} & \textbf{Problem Type} & \textbf{Decision Abstraction} & \textbf{Scope of Control} & \textbf{RL Method} & \textbf{Objective(s)} \\ \hline
Waschneck et al. (2018)\cite{waschneck2018deep} & Semiconductor fab & Dispatching-rule selection & Fab/workcenter scheduling & DQN & Throughput, cycle time \\
Kuhnle et al. (2019)\cite{kuhnle2019autonomous} & Semiconductor fab & Dispatching-rule selection & Semiconductor dispatching & Q-learning & Cycle time \\
Park et al. (2019)\cite{park2019reinforcement} & Semiconductor fab & Multi-agent rule selection & Facility-level entities & MA-DQN & Throughput \\
Luo et al. (2020)\cite{luo2020dynamic} & DFJSP & Job-machine assignment & Full shop & DDQN & Makespan \\
Chang et al. (2022)\cite{chang2022deep} & Semiconductor fab & Job selection & Full fab & DDQN-style value learning & Throughput \\
Zhou et al. (2022)\cite{zhou2022reinforcement} & Job shop & Job selection & Full shop & DQN & Multi-KPI \\
Zhang et al. (2022)\cite{zhang2022dynamic} & Job shop & Priority-based & Full shop & PPO & Tardiness \\
Kovács et al. (2022)\cite{kovacs2022customizable} & Semiconductor fab & Rule selection & Simulator platform & N/A (simulator platform) & Cycle time \\
Xing et al. (2023)\cite{xing2023distributed} & Job shop & Distributed rule selection & Distributed cells & QMIX & Makespan \\
Alexopoulos et al. (2024)\cite{alexopoulos2024deep} & Manufacturing system & Rule selection & Full system & Value-based RL(DQN family) & Tardiness \\
Zhang et al. (2024)\cite{zhang2024adaptive} & Multi-task manufacturing & Hierarchical rule selection & Full system & Hierarchical DQN & Time, cost \\
Yan et al. (2025)\cite{yan2025deep} & DFJSP & Job-machine assignment & Full shop & D3QN & Makespan \\
Jiaxin et al. (2025)\cite{jiaxin2025adaptive} & Semiconductor fab & Hybrid priority-based control & Wafer fab & Model-free RL & Throughput \\
Immordino et al. (2025)\cite{immordino2025explainable} & Semiconductor fab & Rule selection & Semiconductor fab & CMA-ES-trained attention policy + XAI & Throughput, tardiness  \\
Stockermann et al. (2025)\cite{stockermann2025scalability} & Semiconductor fab & Rule selection & Fab dispatching & Hybrid RL--EA & Throughput \\
Hong \& Li (2025)\cite{hong2025large} & FJSP & Dynamic formulation refinement & Full shop & LLM + A3C & Makespan, tardiness \\
\hline
\textbf{This work} & \textbf{Semiconductor fab} & \textbf{Job selection} & \textbf{Semiconductor Fab} & \textbf{Multiple algorithms} & \textbf{Throughput, utilization} \\ \hline
\end{tabular}%
}
\end{table}

\section{Analyses}

\subsection{ Analyzing Collected Experiences}
\label{sec:analyzing_col_exp}

In real semiconductor fabs, products follow long and diverse routes, and decision-related operational attributes can exhibit strongly skewed, long-tailed behavior. These characteristics induce highly heterogeneous decision contexts across time and equipment, motivating the decision-level distributional analysis of collected experiences presented in this subsection.

This section adopts a decision-focused analysis that characterizes the system conditions observed by the control policy at each dispatching decision, rather than tracking individual lots from release to completion.
The analysis window is defined by the agent call time (i.e., the instant at which the agent observes the environment and selects an action); therefore, decisions are included if they occur within a predefined steady-state interval (e.g., shifts 20–110). This intentionally decouples the analysis from job arrival (\texttt{queue\_time}) and completion (\texttt{out\_time}) timestamps, which are more appropriate when evaluating job-centric metrics such as flow time (\texttt{out\_time} - \texttt{in\_time} ) or throughput over complete trajectories. Consequently, each row of the resulting dataset corresponds to a single decision at a specific simulation time within the steady-state window, enabling direct inspection of policy behavior under comparable operating conditions. To summarize the decision context, the system state tensor $\mathbf{s}_{a,0}$ is aggregated across all lots currently present in the fab, and mean statistics are computed for lot quantity, anonymized timing-related attributes, batching capability, LCR-related indicators, sequencing features, and sector-level activity indicators.

Two complementary distributional analyses are then conducted on the decision-level dataset. 
First, selected metrics are discretized using empirical quartiles computed over all decisions. Each decision is assigned to one of four quartile groups (Q1–Q4) for each metric, and counts and proportions are summarized overall and by scenario. This yields an interpretable scenario-level comparison of how frequently the policy encounters low, typical, and extreme regions of the state space, while abstracting from within-quartile variation. 
Second, continuous distributions are compared using kernel density estimates (KDEs) and empirical cumulative distribution functions (CDFs), allowing assessment of scenario-dependent shifts in distributional shape, skewness, and tail behavior for anonymized decision-level metrics.
\begin{figure*}[h]
    \centering
    \includegraphics[width=\textwidth]{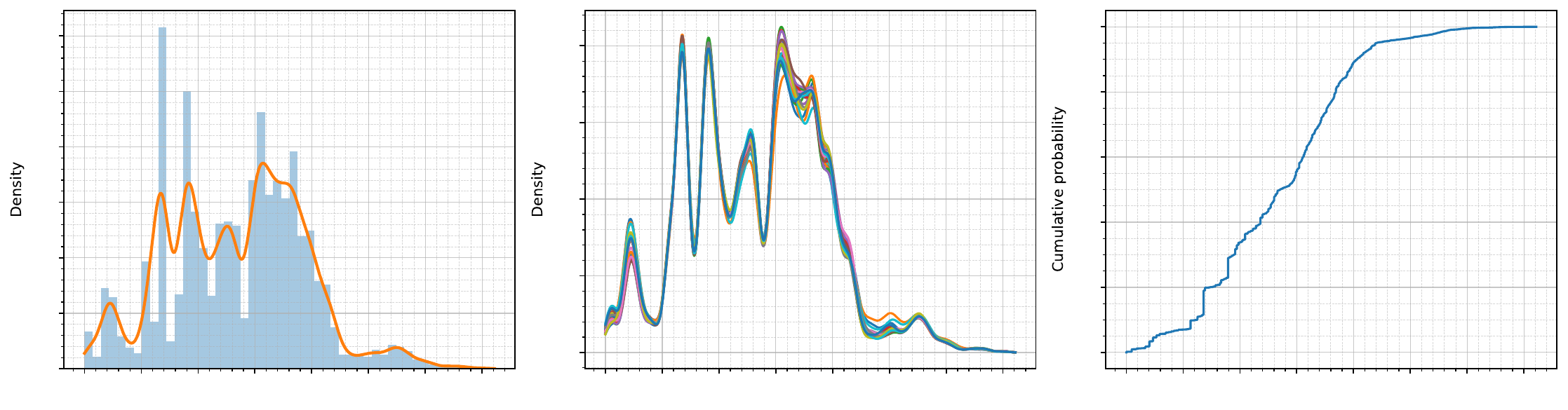}
    \caption{Scenario-dependent distribution of an anonymized decision-time metric. The figure reports (a) the overall histogram with a global KDE computed over all decisions, (b) KDE curves overlaid by scenario, and (c) the empirical CDF over all decisions. All distributions are computed from decision instances within the steady-state call-time window, and  a monotone transformation is applied to improve comparability and reduce tail dominance.}
    \label{fig:kde_by_scenario}
\end{figure*}

For a subset of metrics, the analysis is further conditioned on queue pressure by stratifying decisions according to quartiles of a congestion indicator (e.g., number of queue moves) and computing KDEs/CDFs within each congestion group. This conditional view reveals how the anonymized metric distribution changes as the system transitions from low- to high-pressure operating regimes, and complements the quartile summaries with finer-grained distributional evidence.

\begin{figure*}[h]
    \centering
        \includegraphics[width=\textwidth]{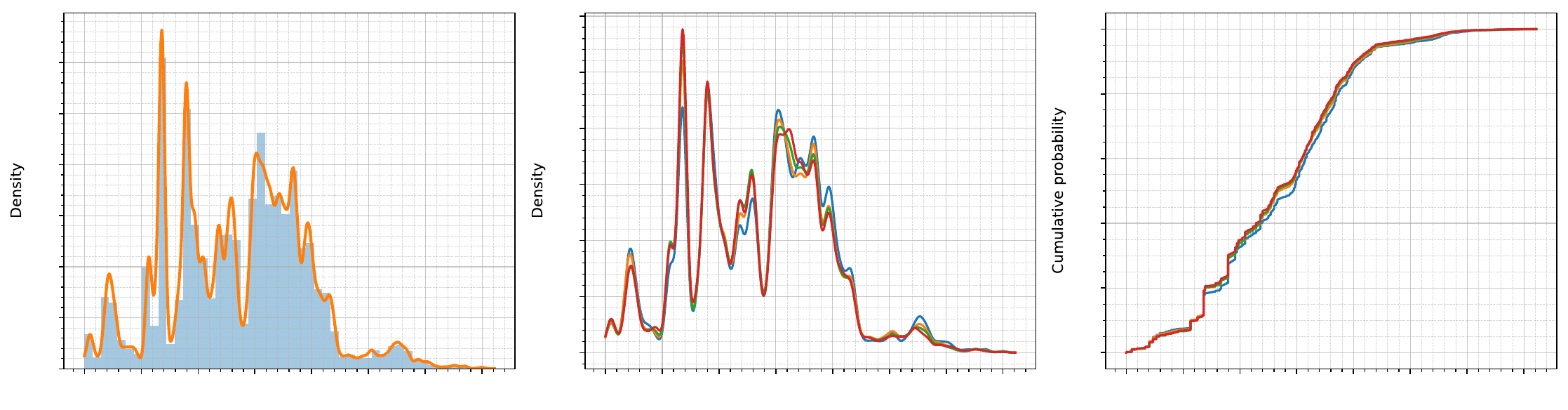}
    \caption{Distributions of an anonymized timing-related metric conditioned on queue pressure. Decisions are stratified into quartiles (Q1--Q4) of queue pressure, measured by the number of queue moves observed at the decision point. The figure shows (a) the overall histogram with global KDE, (b) KDE curves within each queue-pressure quartile, and (c) the corresponding empirical CDFs. Differences between quartiles indicate how the metric distribution shifts as the system transitions from low to high congestion.}
    
    \label{fig:kde_by_queue_quartiles}
\end{figure*}

Based on the distributional behavior observed (e.g., differences in shape, skewness, and tail behavior across decision contexts), we further examine route-length variability across the product mix as a plausible structural source of this heterogeneity.
\begin{figure*}[h]
    \centering
    \includegraphics[width=\textwidth]{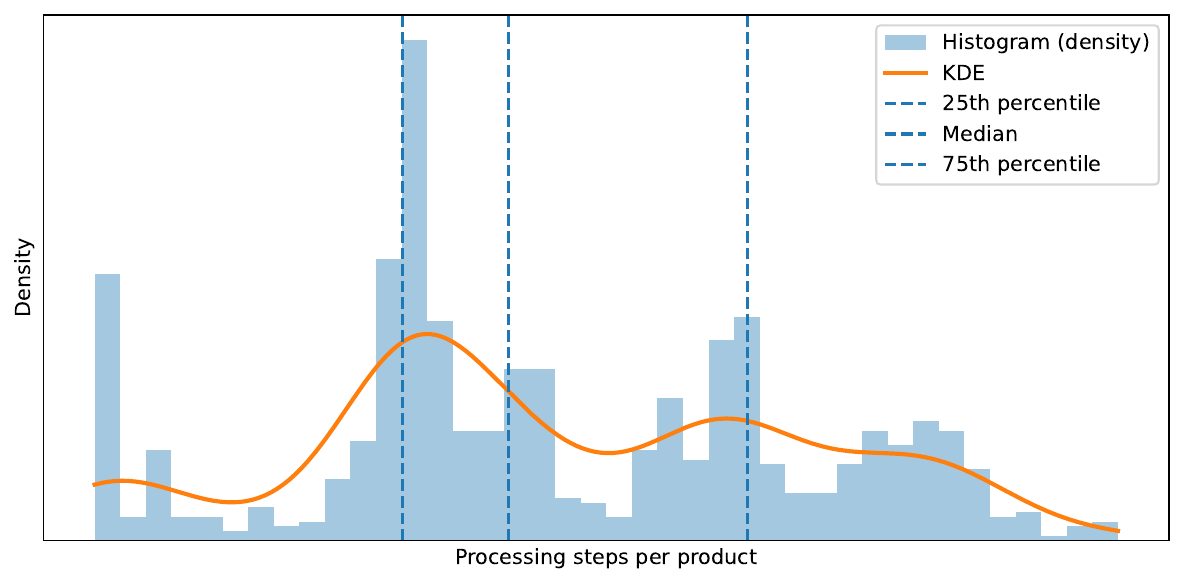}
    \caption{Distribution of processing steps per product in the dataset. The density histogram and kernel density estimate (KDE) are shown, together with the 25th percentile, median, and 75th percentile.}
    \label{fig:steps_per_product}
\end{figure*}
The distribution of route-length indicators highlights substantial variability: while many instances concentrate around a typical number of steps, the distribution remains broad across percentile ranges, indicating a non-trivial portion of cases with substantially longer (or shorter) routes. This variability is consistent with the presence of qualitatively different decision contexts in the collected experiences, since decisions encountered along shorter versus longer routes can differ in congestion exposure and time-scale effects.

\subsubsection{ Policy Fluctuation and Bottleneck Diagnostics}
\label{app:policy_fluc}
In addition to the decision-level state distribution analysis (Section \ref{sec:analyzing_col_exp}), complementary diagnostics are performed to (i) quantify how dispatching behavior changes over time and (ii) relate performance degradations to bottlenecks, sector congestion, and WIP structure. The objective is to distinguish behavioral nonstationarity (shifts in the policy’s realized decision preferences) from system-side stress (changes in congestion and capacity usage reflected in resource-level, sector-level, and WIP-related signals). This separation supports interpretation of why certain scenarios exhibit unstable KPIs even when evaluated under comparable steady-state conditions.
\paragraph{Shift-wise policy fluctuation via KL divergence.}
Consecutive-shift KL divergence was used instead of a fixed baseline comparison in order to track local behavioral drift directly and avoid dependence on an arbitrary reference shift. The evaluation horizon is partitioned into fixed production shifts. For each shift, an empirical distribution over decision outcomes is estimated from all decisions observed within that shift. Shift-to-shift behavioral change is quantified using the consecutive Kullback–Leibler (KL) divergence between adjacent shift distributions. Larger KL values indicate stronger changes in realized behavior, suggesting reduced temporal consistency. Using consecutive shifts avoids dependence on an arbitrary baseline and directly tracks behavioral drift.

\begin{figure}[h]
    \centering
    \includegraphics[width=0.65\linewidth]{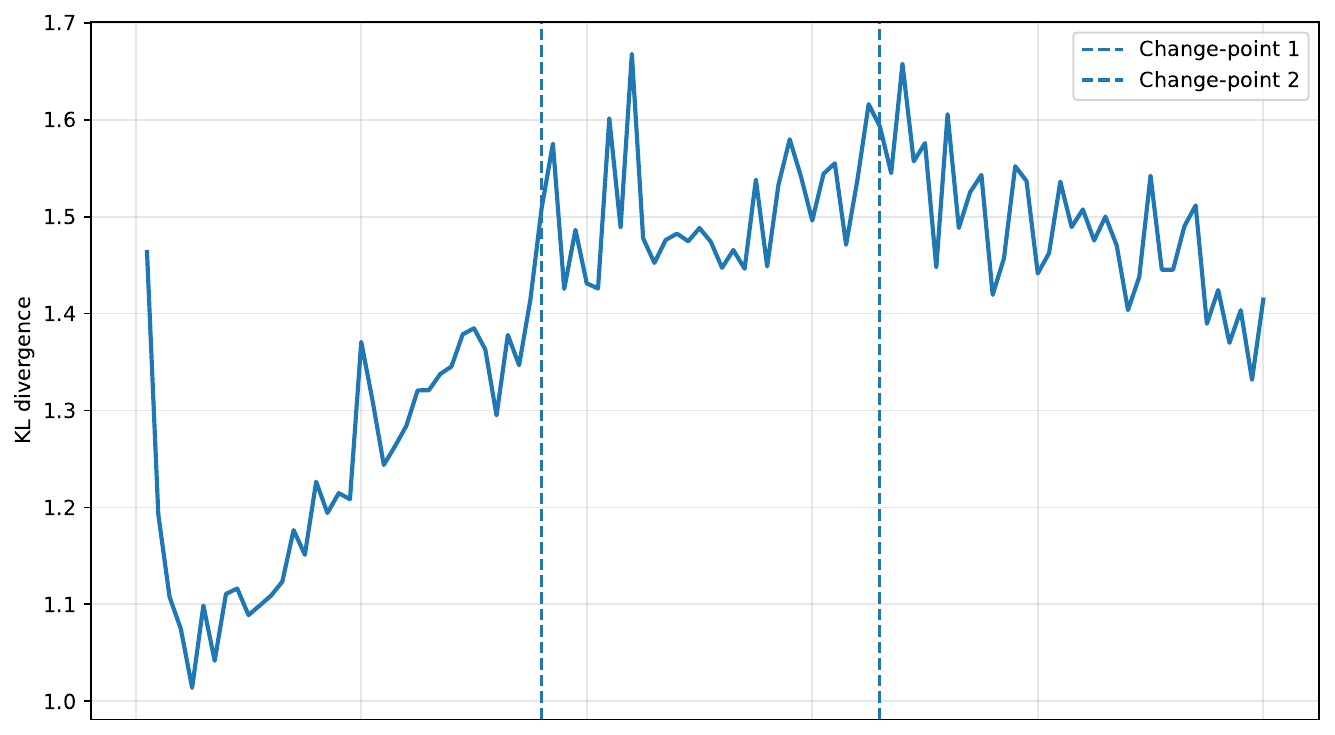} 
    \caption{Shift-wise policy fluctuation measured by the mean KL divergence between consecutive action distributions (\texttt{OPER}), aggregated across scenarios. The dashed vertical lines mark detected change-points, indicating phases of structural change in the realized dispatching behavior.}
    \label{fig:policy-fluctuation-changepoints}
\end{figure}

To further interpret the temporal behavior of the learned policy, we analyze shift-wise fluctuations in the realized action distribution. Figure~\ref{fig:policy-fluctuation-changepoints} reports the mean KL divergence between consecutive shift-level action distributions, aggregated across scenarios. The detected change-points indicate phases in which the policy behavior changes structurally over time.

The KL profile in Figure~\ref{fig:policy-fluctuation-changepoints} provides the basis for identifying persistent changes in policy behavior over time.

\paragraph{Detecting persistent change points.} To identify structural (non-transient), the KL time series is examined for change points. Candidate change points are flagged when the KL series exhibits a sustained increase in level and/or variability relative to preceding shifts, thereby filtering out short-lived fluctuations consistent with noise. The resulting change points identify shift locations where the policy enters a qualitatively different behavioral regime.

\paragraph{Linking behavior changes to KPI variability with delays.} Because downstream KPI effects may appear with delay, lagged associations are evaluated between the KL divergence series and KPI variability (e.g., variance of throughput, saturation, or window load). Correlations are computed under integer shift delays and summarized by reporting the delay that maximizes association strength. This provides an interpretable estimate of how long it takes for behavioral instability to manifest as KPI instability.

To complement the behavioral analysis, additional diagnostics are used to connect KPI degradations to plausible physical and structural causes in the fab.

\begin{figure}[h]
    \centering
    \includegraphics[width=0.88\linewidth]{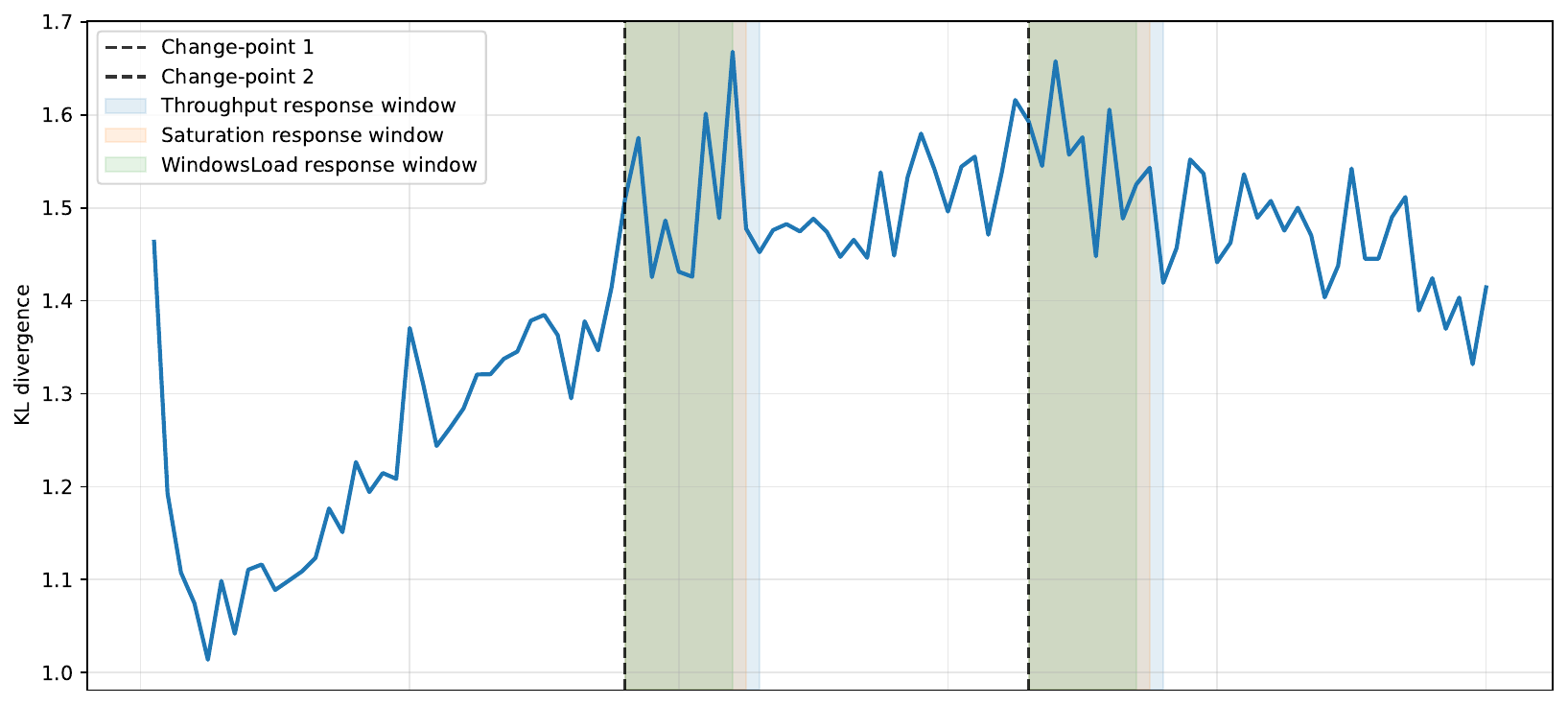} 
    \caption{Timeline of detected policy change-points and delayed KPI response windows. The vertical markers indicate absolute shift locations at which persistent changes in the realized dispatching behavior are detected from the KL-divergence series, whereas the KPI-specific windows indicate the relative delay after which instability is expected to become visible in throughput, saturation, and window load. Thus, each KPI response window represents the combined effect of the absolute change-point location and the estimated lag to KPI manifestation.}
    \label{fig:policy-kpi-delay-timeline}
\end{figure}

To further support interpretation, Figure~\ref{fig:policy-kpi-delay-timeline} combines the detected policy change point with KPI-specific response delays obtained from the lagged association analysis. In this representation, the change-point location gives the absolute shift at which the policy behavior changes structurally, while the estimated delay gives the relative number of shifts after which the instability is expected to appear in a given KPI. The resulting timeline therefore provides an interpretable estimate of when behavioral instability is likely to manifest in throughput, saturation, and window load.

For example, if a change-point is detected at a given shift and the lag analysis indicates a KPI-specific delay of several shifts, then the expected KPI response window is obtained by combining the absolute change-point location with that relative delay.

Once delay KPI instability has been temporally localized, the remaining diagnostics are used to examine its possible physical causes within the fab. 
\paragraph{Bottleneck and sector coupling analysis}
To connect KPI degradation to physical constraints, bottleneck candidates are derived from resource-level activity by ranking resources according to a pressure indicator that captures sustained load and persistent busy-state behavior. Periods of elevated resource load are then related to sector-level congestion by identifying the sector-level indicators that increase most under stress conditions. Aggregating these results across scenarios yields a scenario-level view of which sectors are most consistently associated with stress periods.

\paragraph{WIP structure and concentration diagnostics} High-dimensional WIP signals are summarized using compact indicators that capture both volume and concentration. In addition to total WIP, measures such as the number of active WIP components and concentration/spread indicators (e.g., entropy) are computed. These indicators are aligned with KPI trajectories to assess whether performance drops coincide with increasing WIP, rising WIP concentration into fewer dimensions, or shifts in the mix of products/bins contributing to congestion.

Overall, these diagnostics provide a consistent chain of evidence linking (i) temporal drift in dispatching behavior to (ii) delayed KPI instability and (iii) concurrent bottleneck, sector, and WIP signatures, thereby supporting a mechanistic interpretation of scenario differences.


\subsubsection{Process Mining for Routing Diagnostics} 

Semiconductor manufacturing processes are highly complex, multi-step, and data-intensive. In order to validate and summarize the routing behavior observed in the industrial use case, process mining was applied to transform heterogeneous manufacturing data into an event log and to discover the dominant control-flow structure. Because the original data was large and highly detailed, a small representative sample was constructed to enable rapid iteration and visualization while preserving the main routing characteristics of the process. This use of process mining follows the same objective adopted in the initial report, namely to reconstruct, visualize, and interpret the actual production flow from available fab data. 

The event log was derived by integrating four complementary manufacturing data sources: qualification and routing information, a Work-in-Progress (WIP) snapshot, an equipment registry, and an event library containing nominal processing-time information. In this representation, each lot was treated as a case and each routed step was treated as an event. Since nominal processing times were used instead of execution timestamps, timestamps were generated sequentially along the route. In this way, event ordering was preserved and a consistent basis for process discovery was obtained, while timing information remained suitable only for relative diagnostic interpretation rather than exact cycle-time measurement. 

To comply with industrial confidentiality constraints, all proprietary activity identifiers were anonymized before analysis. Each operation was mapped to a generic label (\texttt{Op\_001}, \texttt{Op\_002}, \ldots, \texttt{Op\_00n}), and non-informative or placeholder events were handled during preprocessing rather than being retained as explicit nodes in the final visualization. In addition, low-support edges were filtered so that only the most frequent transitions were preserved. In this way, the dominant routing backbone was retained while rare, potentially identifying routing signatures were suppressed. 

\begin{figure}[t]
    \centering
    \includegraphics[width=0.85\linewidth]{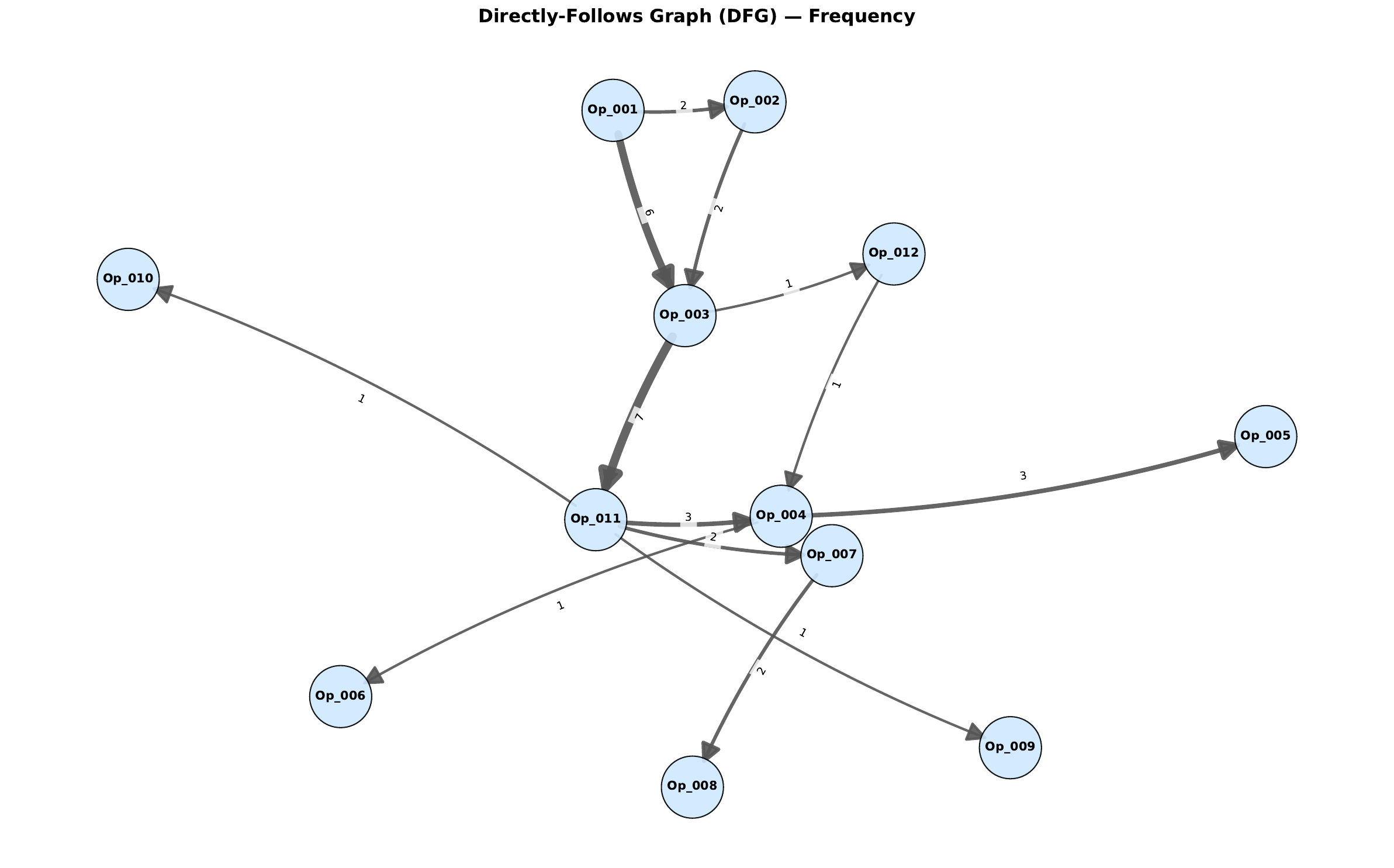}
    \caption{Anonymized Directly-Follows Graph (DFG) based on transition frequency. Nodes represent anonymized operations (\texttt{Op\_001}, \texttt{Op\_002}, \ldots, \texttt{Op\_012}), while directed edges indicate directly-following relations in the constructed nominal event log. Edge labels denote transition frequency.}
    \label{fig:dfg_frequency}
\end{figure}

Process discovery was then performed on the anonymized event log to obtain the Directly-Follows Graph (DFG) shown in Fig.~\ref{fig:dfg_frequency}. In this graph, nodes represent anonymized operations and directed edges represent directly-following relations observed in the sampled lot traces. Edge labels indicate the frequency with which each transition was observed. Since the purpose of this subsection is structural routing diagnosis, only the anonymized frequency-based DFG was retained in the paper. 

As shown in Fig.~\ref{fig:dfg_frequency}, the discovered process structure was found to be predominantly semi-linear, with a limited number of branches around a central routing backbone. A small set of transitions carries most of the observed flow, while a few lower-frequency paths reflect alternative handling patterns or product-family differences. This interpretation is consistent with the initial report, in which the manufacturing process was also characterized as mostly linear with limited branching and with a dominant route accompanied by a small number of less frequent variants. 

The anonymized visualization therefore preserves the essential structural conclusions of the original process-mining analysis without exposing fab-specific operation names. It can still be observed that the sampled manufacturing flow progresses through a dominant sequence of operations before diverging into a small number of alternative downstream paths. In this sense, the DFG provides a compact routing diagnostic that confirms both the existence of a main process backbone and the presence of limited but meaningful routing variation within the sampled data. 

\newpage 

\subsection{Impact of sector-level information.}
\label{sec:sector_analysis}

\begin{figure}[h]
\centering
\begin{minipage}{0.55\textwidth}
    \centering
    \includegraphics[width=\linewidth]{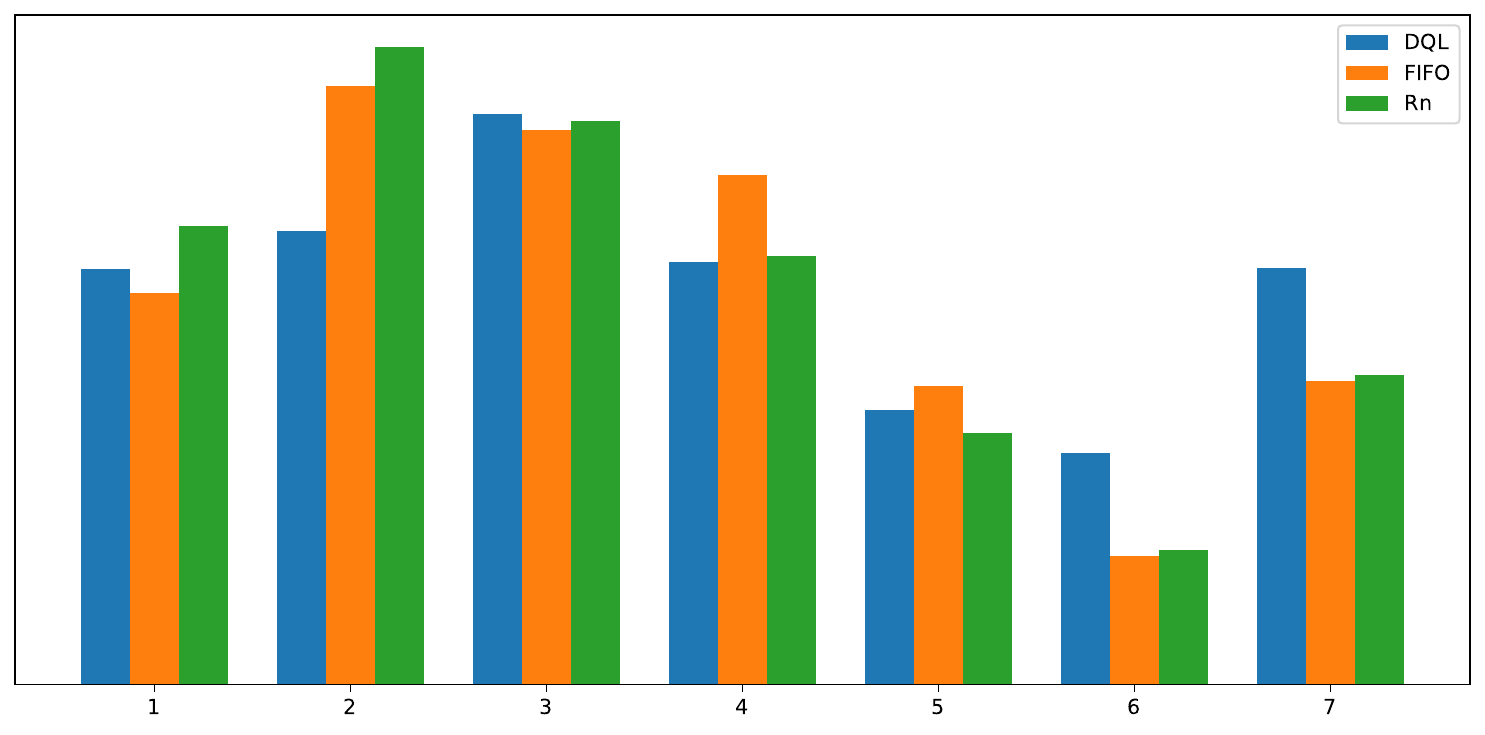}
    \caption{Sectors distribution across scenarios for DQL, FIFO, and Random. The comparison highlights differences in workload allocation across fab sectors relative to the FIFO baseline.}
    \label{fig:sector_distribution_threeagents}
\end{minipage}
\hfill
\begin{minipage}{0.40\textwidth}
    \centering
    \begin{tabular}{lcc}
    \hline
    \textbf{Sector} & \textbf{Random (\%)} & \textbf{DQL (\%)} \\
    \hline
    Sector 1 & 17.24 & 6.11 \\
    Sector 2 & 6.52 & -24.25 \\
    Sector 3 & 1.72 & 2.89 \\
    Sector 4 & -15.99 & -17.15 \\
    Sector 5 & -15.87 & -7.94 \\
    Sector 6 & 5.08 & 80.48 \\
    Sector 7 & 2.08 & 37.37 \\
    \hline
    \end{tabular}
    \captionof{table}{Percentage change in sector-level mean distribution for Random and DQL relative to FIFO.}
    \label{tab:sector_gain_fifo_threeagents}
\end{minipage}
\end{figure}

\paragraph{DQL} To assess the role of sector-level information in the state representation, the distribution of lots across sectors is compared under DQL, Random, and the FIFO baseline. The resulting comparison is summarized in Figure~\ref{fig:sector_distribution_threeagents} and Table~\ref{tab:sector_gain_fifo_threeagents}. In particular, the largest positive changes under DQL are observed in Sector~6 and Sector~7, while smaller gains appear in Sector~1 and Sector~3. Negative changes are observed in Sectors~2, 4, and 5. Random exhibits more limited deviations from FIFO across most sectors. Although this comparison does not constitute a full ablation study, it provides supportive evidence that sector-level information contributes useful structure to the state representation by reflecting workload differences across the fab. 

\begin{figure}[h]
\centering
\begin{minipage}{0.55\textwidth}
    \centering
    \includegraphics[width=\linewidth]{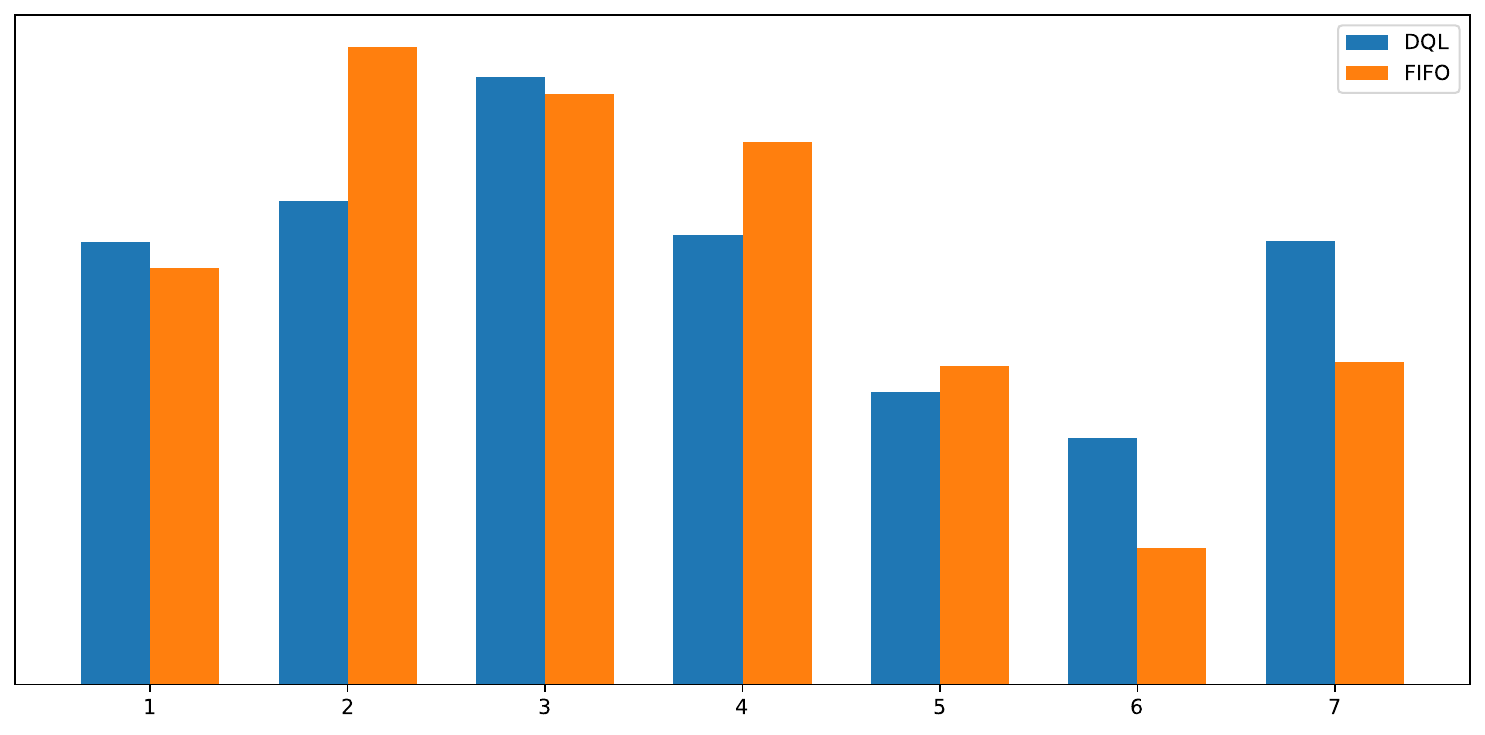}
    \caption{Sectors distribution across scenarios for DQL and FIFO. The comparison highlights how the learned policy reshapes workload allocation across fab sectors relative to the FIFO baseline.}
    \label{fig:sector_distribution}
\end{minipage}
\hfill
\begin{minipage}{0.40\textwidth}
    \centering
    \label{tab:sector_gain_fifo}
    \begin{tabular}{lc}
    \hline
    \textbf{Sector} & \textbf{Difference (\%)} \\
    \hline
    Sector 1 & 6.11 \\
    Sector 2 & -24.25 \\
    Sector 3 & 2.89 \\
    Sector 4 & -17.15 \\
    Sector 5 & -7.94 \\
    Sector 6 & 80.48 \\
    Sector 7 & 37.37 \\
    \hline
    \\~\\
    \end{tabular}
    \captionof{table}{Percentage change in sector-level mean distribution for DQL relative to FIFO.}
    \label{tab:sector_gain_fifo}
\end{minipage}
\end{figure}

\newpage

\begin{figure}[h]
\centering
\begin{minipage}{0.55\textwidth}
    \centering
    \includegraphics[width=\linewidth]{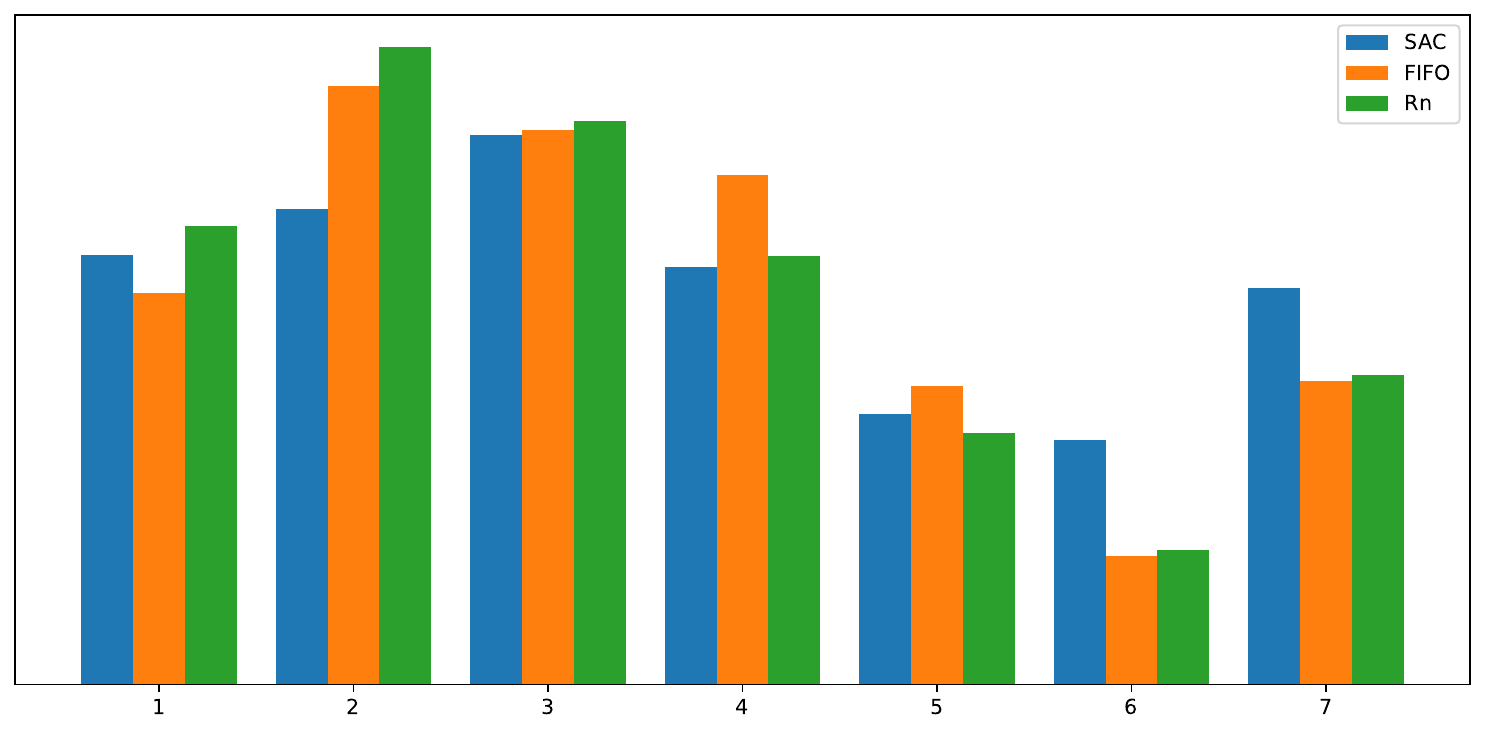}
        \caption{Sector distribution across scenarios for SAC, FIFO, and Random. The comparison highlights differences in workload allocation across fab sectors relative to the FIFO baseline.}
    \label{fig:sector_distribution_sac_threeagents}
\end{minipage}
\hfill
\begin{minipage}{0.40\textwidth}
    \centering
    \begin{tabular}{lcc}
    \hline
    \textbf{Sector} & \textbf{Random (\%)} & \textbf{SAC (\%)} \\
    \hline
    Sector 1 & 17.24 & 9.76 \\
    Sector 2 & 6.52 & -20.62 \\
    Sector 3 & 1.72 & -0.82 \\
    Sector 4 & -15.99 & -18.07 \\
    Sector 5 & -15.87 & -9.50 \\
    Sector 6 & 5.08 & 91.10 \\
    Sector 7 & 2.08 & 30.89 \\
    \hline
    \end{tabular}
    \captionof{table}{Percentage change in sector-level mean distribution for Random and SAC relative to FIFO.}
    \label{tab:sector_gain_fifo_sac_threeagents}
\end{minipage}
\end{figure}

\paragraph{SAC} To further examine the role of sector-level information in the state representation, we also compare the distribution of lots across sectors under SAC with \(c_h=0.98\), Random, and the FIFO baseline. The resulting comparison is summarized in Fig.~\ref{fig:sector_distribution_sac_threeagents} and Table~\ref{tab:sector_gain_fifo_sac_threeagents}. Similar to DQL, the largest positive changes under SAC are observed in Sectors~6 and 7, with a smaller positive change in Sector~3, whereas Sector~1 remains close to the FIFO baseline. Negative changes are observed in Sectors~2, 4, and 5. Compared with Random, SAC induces a stronger redistribution of workload across sectors, particularly in Sector~6. Although this comparison does not constitute a full ablation study, it provides further evidence that sector-level information offers useful structure for learned dispatching policies by capturing workload differences across the fab.

\begin{figure}[h]
\centering
\begin{minipage}{0.55\textwidth}
    \centering
    \includegraphics[width=\linewidth]{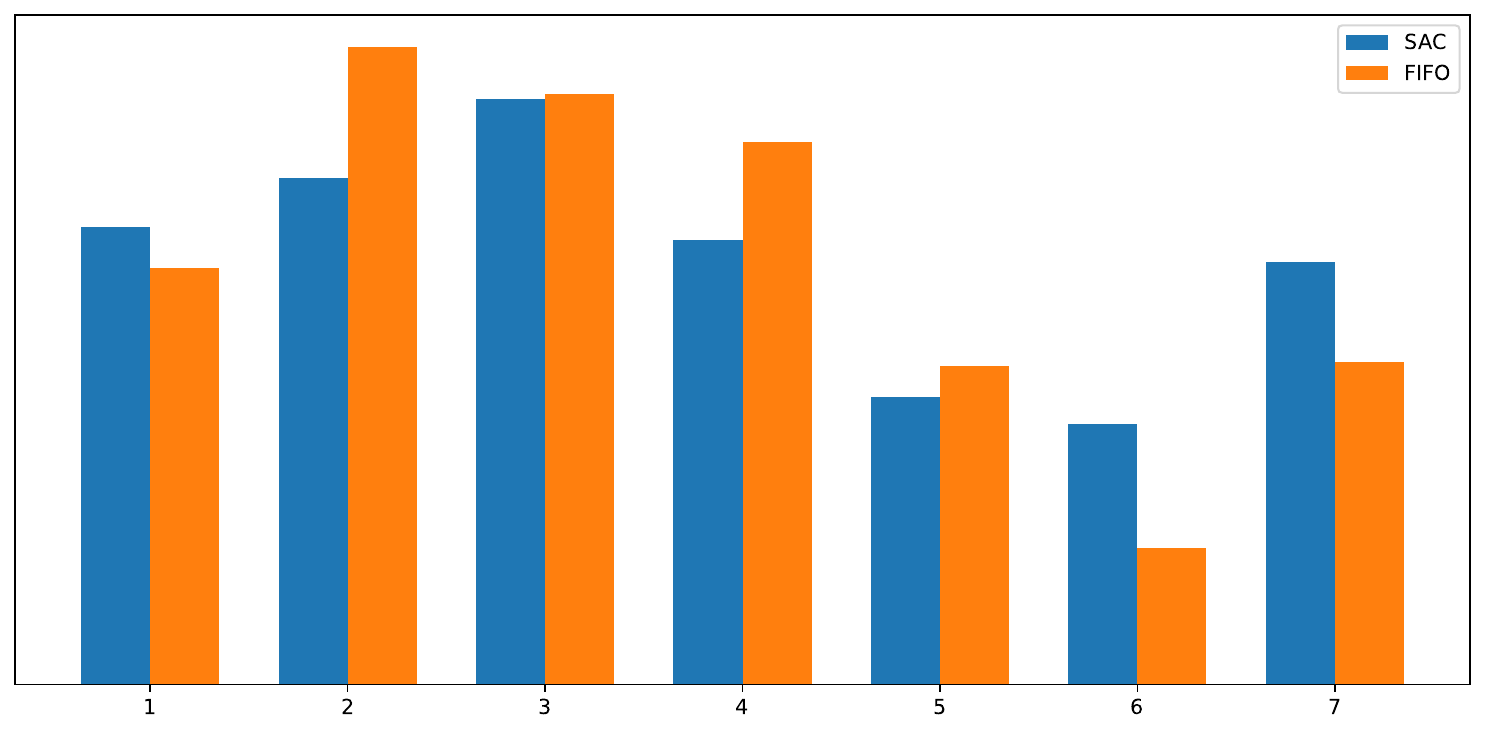}
    \caption{Sector distribution across scenarios for SAC and FIFO. The comparison highlights how the learned SAC policy reshapes workload allocation across fab sectors relative to the FIFO baseline.}
    \label{fig:sector_distribution_sac}
\end{minipage}
\hfill
\begin{minipage}{0.40\textwidth}
    \centering
    \begin{tabular}{lc}
    \hline
    \textbf{Sector} & \textbf{Difference (\%)} \\
    \hline
    Sector 1 & 9.76 \\
    Sector 2 & -20.62 \\
    Sector 3 & -0.82 \\
    Sector 4 & -18.07 \\
    Sector 5 & -9.50 \\
    Sector 6 & 91.10 \\
    Sector 7 & 30.89 \\
    \hline
    \end{tabular}
    \captionof{table}{Percentage change in sector-level mean distribution for SAC relative to FIFO.}
    \label{tab:sector_gain_fifo_sac}
\end{minipage}
\end{figure}

\newpage

\section{Policy Optimization}

\subsection{Derivation and Interpretation of the Group-Relative Surrogate Advantage}
\label{sec:surrogate_advantage}

In the event-group temporal-difference framework, and in the absence of an explicit state-value critic, we can construct a group-relative advantage signal for PPO. Recall that, for an event-action pair $(e \in \mathcal{G})$, the event-level temporal-difference discrepancy is defined as
\begin{equation}
    \delta_e =
    Q(s_e,a_e) - r_e^e - \gamma V(s'_e),
    \label{eq:app_delta_def}
\end{equation}
where $s_e$ and $s'_e$ denote the boundary states associated with event $e$, $a_e$ is the selected action,
$r_e^e$ is the event-level (shaped) reward, and $V(s'_e)$ denotes the bootstrap value. For a sampled segment
\begin{equation}
    I_h = [t,t+h],
\end{equation}
we denote by
\begin{equation}
 \mathcal{G}  = \mathcal{E}(I_h),
\end{equation}
the group of event-action pairs associated with the segment. Under mean aggregation, the event-group
temporal-difference objective is
\begin{equation}
    \mathcal{L}_m(\theta)
    =
    \left(
    \frac{1}{|\mathcal{G}|}
    \sum_{e\in \mathcal{G}}\delta_e(\theta)
    -
    r^g(I_h)
    \right)^2,
    \label{eq:app_mean_agg}
\end{equation}
where $r^g(I_h) = r^g(t,t+h)$ is the system-level reward over the segment. Hence, when the grouped
critic objective is optimized, the mean event-level discrepancy is encouraged to match the segment-level
system reward:
\begin{equation}
    \frac{1}{|\mathcal{G}|}
    \sum_{e\in \mathcal{G}}\delta_e
    \approx
    r^g(I_h).
    \label{eq:app_group_mean_delta}
\end{equation}

The standard PPO policy-gradient update is driven by an advantage estimate of the form
\begin{equation}
    A(s_i,a_i)
    =
    Q(s_i,a_i)-V(s_i).
    \label{eq:app_standard_adv}
\end{equation}
Using the event-level discrepancy in Eq.~\eqref{eq:app_delta_def}, for event $i\in G$ we have
\begin{equation}
    \delta_i
    =
    Q(s_i,a_i) - r_i^e - \gamma V(s_i'),
    \label{eq:app_delta_i}
\end{equation}
which, rearranging, gives it
\begin{equation}
    Q(s_i,a_i)
    =
    \delta_i + r_i^e + \gamma V(s_i').
    \label{eq:app_q_from_delta}
\end{equation}
Substituting Eq.~\eqref{eq:app_q_from_delta} into the standard advantage in
Eq.~\eqref{eq:app_standard_adv} yields
\begin{align}
    A(s_i,a_i)
    &=
    \delta_i + r_i^e + \gamma V(s_i') - V(s_i)
    \nonumber\\
    &=
    \delta_i
    -
    \left[
    V(s_i) - r_i^e - \gamma V(s_i')
    \right].
    \label{eq:app_adv_decomp}
\end{align}
Thus, the exact advantage can be written as the event-level discrepancy $\delta_i$ minus an event-specific
baseline,
\begin{equation}
    b_i
    =
    V(s_i) - r_i^e - \gamma V(s_i').
    \label{eq:app_event_baseline}
\end{equation}
When a separate state-value critic is trained, this exact form can be evaluated directly through
$Q(s_i,a_i)-V(s_i)$. However, in the surrogate PPO variant, the policy update is 
constructed from the event-group TD critic without introducing an additional value network. In that case, we
replace the event-specific baseline $b_i$ by the shared group-level baseline induced by the aggregation
objective:
\begin{equation}
    b_i
    =
    V(s_i) - r_i^e - \gamma V(s_i')
    \approx
    r^g(I_h).
    \label{eq:app_baseline_approx}
\end{equation}
This gives the group-relative surrogate advantage
\begin{equation}
    \tilde{A}_i
    \triangleq
    \delta_i - r^g(I_h).
    \label{eq:app_surrogate_advantage}
\end{equation}
For compactness, the main text writes this as
\begin{equation}
    \tilde{A}_i \triangleq \delta_i - r^g.
\end{equation}

An equivalent interpretation follows directly from Eq.~\eqref{eq:app_group_mean_delta}. Since the
aggregation objective encourages
\begin{equation}
    r^g(I_h)
    \approx
    \frac{1}{|\mathcal{G}|}\sum_{j\in \mathcal{G}}\delta_j,
\end{equation}
the surrogate advantage can be viewed as an approximately centered event-level residual:
\begin{equation}
    \tilde{A}_i
    =
    \delta_i - r^g(I_h)
    \approx
    \delta_i
    -
    \frac{1}{|\mathcal{G}|}
    \sum_{j\in \mathcal{G}}\delta_j.
    \label{eq:app_centered_residual}
\end{equation}

Therefore, $\tilde{A}_i$ measures whether event $i$ has a larger or smaller critic-implied
contribution than the average event in the coupled segment. Positive values indicate event-action pairs whose
TD discrepancy lies above the group baseline, while negative values indicate event-action pairs whose
discrepancy lies below it. This construction should be interpreted as a group-relative surrogate rather than an exact identity with the
standard PPO advantage. During the PPO actor update, $\tilde A_i$ is treated as a fixed advantage estimate; gradients are not propagated through the critic terms used to construct $\delta_i$.

\section{Experiments}

\subsection{Hyperparameter}

\begin{table}[H]
\caption{Shared hyperparameters used for the reported experiments. The table lists the training and model hyperparameters used to generate the reported offline and online results.}
\label{tab:hyperparameters}
\centering
\begin{tabular}{|l|l|}
\hline
\textbf{Hyperparameter} & \textbf{Value} \\ \hline
Q and V and $\pi$ Layers & 4 fully connected layers \\ \hline
Hidden dimension & [1024, 512, 256, 128] \\ \hline
Dropout & 0.1 \\ \hline
Segment length &  360 minutes \\ \hline
Replay buffer capacity &  100,000 \\ \hline
Discount rate & 0.99 \\ \hline
Learning rate (offline) & 1e-5 \\ \hline
Learning rate (online: Critic/$Q$) & 5e-5 \\ \hline
Learning rate (online: Actor) & 5e-5 \\ \hline
Entropy target & $-~ c_h ~*~ \log(1 / |\mathcal{A}|) ~s.t. ~ c ~ \in ~ \{0.98, 0.5\} $ \\ \hline
Optimizer & Adam \\ \hline
Gradient clipping & 5.0 \\ \hline
Replay ratio (off-policy) & 3 \\ \hline
Update epochs (on-policy) & 10 \\ \hline
Gradient steps (offline) & 100,000 \\ \hline
Gradient steps (online) &  up to 15,000 \\ \hline
\end{tabular}
\vspace{2mm}

\end{table}

\subsection{Validation}
\label{sec:valid}

To evaluate agent performance, we simulated each policy in a high-fidelity semiconductor manufacturing simulation and recorded trajectories of the KPIs over time. Let \(x^{(a)}_{s,r,k,j}\) denote the value of KPI \(j\) produced by agent \(a\) in scenario \(s\), random seed \(r\), and shift \(k\), where \(j \in \{\text{\ Throughput}, \text{Saturation}, \text{Load}\}\). Load is computed as a 60-minute moving average according to Eq.~\ref{eq:x_l}, whereas saturation is defined as the instantaneous ratio of equipment utilization at the sampling time. Validation was performed on 10 test scenarios spanning different work-in-progress and input configurations. For each test scenario, the simulator was first warmed up for 20 shifts under a random policy using three fixed random seeds, in order to remove dependence on the previous policy state and generate distinct initial conditions. This procedure yielded 30 evaluation instances in total. We then defined aggregate performance measures across these instances to quantify policy generalization. Because KPI trajectories at fine temporal resolution are noisy, observations were further aggregated within shifts before computing summary statistics.

To compare policies on a common scale, we express performance as percentage gain relative to a baseline:
\begin{equation}
g^{(a)}_{s,r,k,j}
=
100 \cdot
\frac{x^{(a)}_{s,r,k,j} - b_{s,r,k,j}}
{b_{s,r,k,j}}.
\label{eq:gain_general}
\end{equation}
We use the FIFO policy, denoted by \(f\), as the reference baseline because of its deterministic and operationally meaningful behavior. FIFO performance is evaluated under the same scenario and seed as the agent being compared. We consider two baseline definitions. In the first, a steady-state FIFO baseline is computed by averaging FIFO performance over the post-warm-up window \(B=\{20,\ldots,110\}\):
\begin{equation}
b_{s,r,j}
=
\frac{1}{|B|}
\sum_{k \in B} x^{(f)}_{s,r,k,j}.
\label{eq:fifo_window_baseline}
\end{equation}
In the second, a shift-wise FIFO baseline is defined directly at the matched shift:
\begin{equation}
b_{s,r,k,j}
=
x^{(f)}_{s,r,k,j}.
\label{eq:fifo_shift_baseline}
\end{equation}
The steady-state FIFO baseline provides a stable reference for overall performance, whereas the shift-wise FIFO baseline enables direct comparison against FIFO at matched operating times.

We summarize performance using two complementary aggregation measures. First, for shift-wise analysis, percentage gains are averaged across seeds for each scenario and shift:
\begin{equation}
\tilde{g}^{(a)}_{s,k,j}
=
\frac{1}{|R|}
\sum_{r \in R}
g^{(a)}_{s,r,k,j},
\end{equation}
where \(R\) denotes the set of random seeds. These values are then aggregated across scenarios:
\begin{equation}
\mu^{(a)}_{k,j}
=
\frac{1}{|S|}
\sum_{s \in S}
\tilde{g}^{(a)}_{s,k,j},
\end{equation}
\begin{equation}
\sigma^{(a)}_{k,j}
=
\sqrt{
\frac{1}{|S|}
\sum_{s \in S}
\left(
\tilde{g}^{(a)}_{s,k,j}
-
\mu^{(a)}_{k,j}
\right)^2
},
\end{equation}
where \(S\) denotes the set of evaluation scenarios. These shift-wise statistics are used to visualize the aggregated gain of each agent relative to FIFO over time.

Second, to obtain a single summary value for steady-state performance, we average the shift-wise mean gains over the evaluation horizon \(K\):
\begin{equation}
\mu^{(a)}_{j}
=
\frac{1}{|K|}
\sum_{k \in K}
\mu^{(a)}_{k,j},
\end{equation}
\begin{equation}
\sigma^{(a)}_{j}
=
\sqrt{
\frac{1}{|K|}
\sum_{k \in K}
\left(
\mu^{(a)}_{k,j}
-
\mu^{(a)}_{j}
\right)^2
}.
\end{equation}
This steady-state aggregate provides a single quantitative measure for comparing agents across evaluation instances, while the shift-wise statistics capture how performance evolves during those simulations. We can further compute additional summary statistics, including confidence intervals, based on the samples obtained from each aggregation approach.

\paragraph{Statistical reporting.}
All KPIs are reported as mean percentage gains relative to the FIFO baseline. Each method was evaluated on 10 test scenarios with 3 random seeds used for initialization per scenario, yielding 30 scenario--seed evaluation instances. For each method, gains were first computed at the scenario--seed--shift level relative to the matched FIFO baseline. These shift-level gains were then averaged over the evaluation window to obtain one value per scenario--seed pair. Next, the three seed-level values were averaged within each scenario, yielding one scenario-level estimate per method and KPI. We use scenario-level aggregation for the main summary statistics because scenarios represent the primary independent evaluation units. Averaging seeds within each scenario avoids artificially narrowing uncertainty estimates due to correlations among runs that share the same scenario, and directly quantifies generalization across test scenarios. Finally, the reported mean, sample standard deviation, and confidence interval were computed across the 10 scenario-level estimates. 
Thus, the values reported after the \(\pm\) symbol in the tables denote the half-width of Bonferroni-adjusted two-sided confidence intervals computed over \(n=10\) scenario-level means. For aggregate plots, shaded regions are interpreted according to the corresponding figure caption, since some plots report variability bands rather than confidence intervals.
Because the uncertainty is estimated across the $|S|=10$ scenario-level values, confidence intervals were computed using the Student $t$ distribution rather than the normal approximation. For each agent $a$ and KPI $j$, let $\bar{g}^{(a)}_{s,j}$ denote the scenario-level mean gain. The reported mean is \[
\hat{\mu}^{(a)}_j =
\frac{1}{|S|}
\sum_{s\in S}
\bar{g}^{(a)}_{s,j},
\] and the sample standard deviation across scenarios is\[
\hat{s}^{(a)}_j =
\sqrt{\frac{1}{|S|-1}\sum_{s\in S}\left(\bar{g}^{(a)}_{s,j}-\hat{\mu}^{(a)}_j\right)^2}.
\]
The reported confidence-interval half-width is \[
h^{(a)}_j =
t_{1-\alpha/(2m),\,|S|-1}
\frac{\hat{s}^{(a)}_j}{\sqrt{|S|}},
\] where $m=3$ is the number of KPIs and $|S|=10$. With $\alpha=0.05$, this corresponds to a Bonferroni-adjusted two-sided interval with $9$ degrees of freedom.

\paragraph{Pairwise statistical tests.}
Statistical significance was assessed using paired scenario-level comparisons, with each scenario treated as the independent unit of inference (\(n=10\) matched scenarios). For each pair of agents \(a\) and \(a'\), and for each KPI \(j\), we computed paired scenario-level differences
\[
d_{s,j}^{(a,a')}
=
\bar{g}^{(a)}_{s,j}
-
\bar{g}^{(a')}_{s,j},
\qquad s \in S,
\]
where \(\bar{g}^{(a)}_{s,j}\) denotes the mean percentage gain of agent \(a\) in scenario \(s\) for KPI \(j\). The primary test was a non-parametric paired sign-flip permutation test on the mean paired difference
\[
\bar{d}^{(a,a')}_{j}
=
\frac{1}{|S|}
\sum_{s\in S}
d_{s,j}^{(a,a')}.
\]
The null distribution was obtained by enumerating all possible sign flips of the paired differences and comparing the observed \(|\bar{d}^{(a,a')}_{j}|\) against this distribution using a two-sided test. As robustness checks, we also report Wilcoxon signed-rank tests, which test for a non-zero median paired difference without assuming normality, and paired \(t\)-tests for completeness. To control for multiple testing across the three KPIs within each agent-pair comparison, \(P\)-values were adjusted using the Holm--Bonferroni procedure, and results were considered statistically significant when the Holm-adjusted \(P < 0.05\). Directional conclusions such as \(a>a'\) were made only when the two-sided corrected test was significant and the estimated paired effect \(\bar{d}^{(a,a')}_{j}\) was positive.

\subsection{Offline Experiments}

\subsubsection{IQL TD Ablation with Different Hyperparameters}
\label{sec:td_abl_iql}

For the IQL results in Section~\ref{sec:offline_td}, we used \(\varsigma = 0.8\) and  \(\beta = 3\) . Increasing these values to \(\varsigma = 0.9\) and \(\beta = 10\) in an attempt to extend the effective horizon and improve performance policy extraction resulted in unstable training and degraded performance for the event-aggregated formulation. The results for the remaining TD formulations are summarized in Table~\ref{tab:iql_td_new_params}.

\begin{table}[h]
\caption{IQL KPI gains under $\varsigma = 0.8$ and $\gamma = 0.999$. Values are percentage gains in throughput, saturation and load relative to FIFO, reported as mean $\pm$ Bonferroni-adjusted 95\% confidence-interval across evaluation runs.}
\label{tab:iql_td_new_params}
\centering
\begin{tabular}{lcccc}
\toprule
\textbf{Agent} & \textbf{TD formulation} & \textbf{Throughput} & \textbf{Saturation} & \textbf{Load} \\
\midrule
IQL & Truncated discounted-sequence  & $-11.4 \pm 5.2$ & $-5.9 \pm 3.9$ & $-4.7 \pm 3.5$ \\
IQL & Event-averaged                 & $9.8 \pm 4.7$   & $9.9 \pm 4.1$  & $8.2 \pm 3.3$ \\
IQL & Stacked event-wise            & $15.4 \pm 5.8$  & $14.4 \pm 4.8$ & $11.3 \pm 4.0$ \\
\bottomrule
\end{tabular}
\vspace{2mm}
\end{table}

\begin{figure}[h]
  \centering
  \includegraphics[trim=0 0 0 0,clip,width=0.65\textwidth]{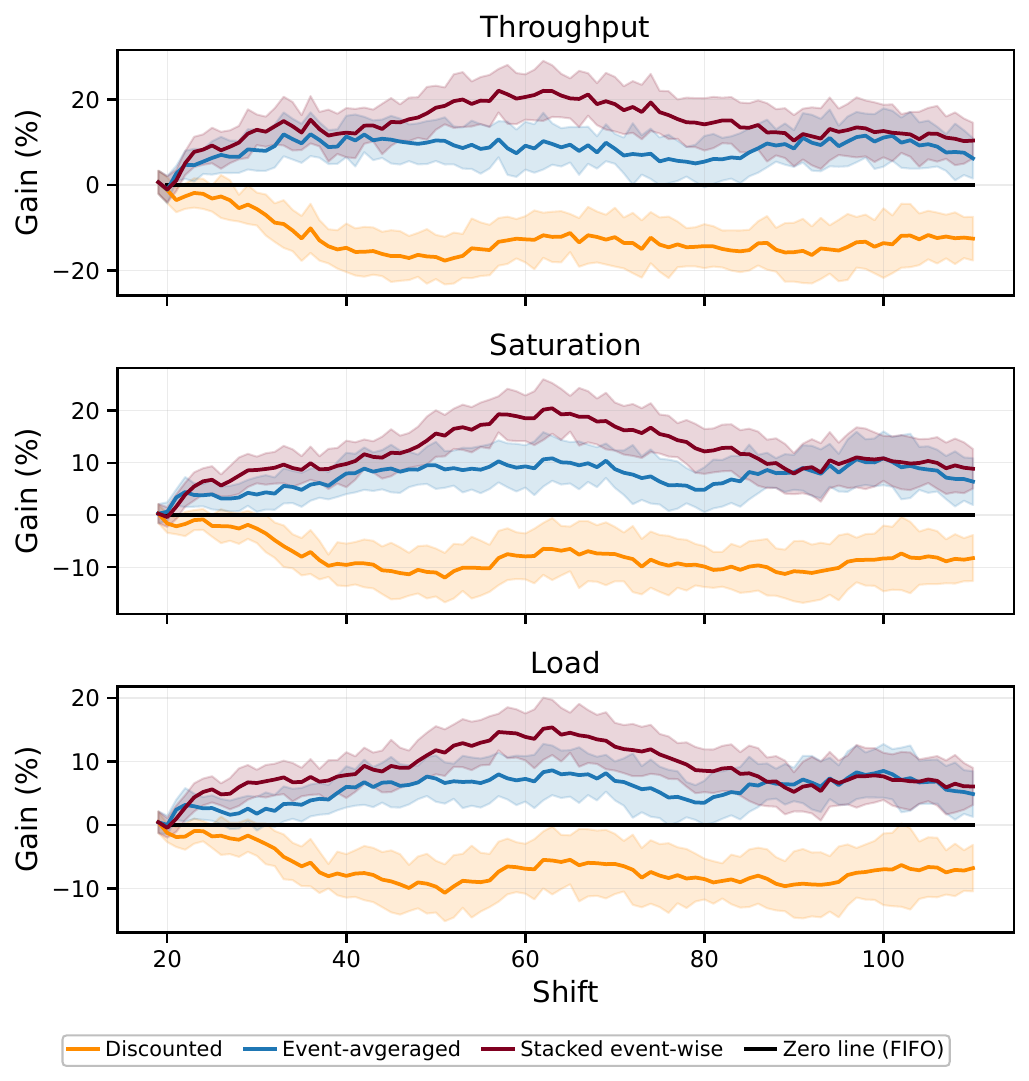}
  \caption{KPI gains (\%) across IQL agents relative to FIFO. Solid lines indicate the mean and shaded regions indicate the standard deviation. The black horizontal line denotes the FIFO reference (zero gain); orange, blue, and dark red correspond to truncated discounted-sequence, event-averaged, and stacked event-wise TD, respectively under \(\varsigma = 0.8\) and \(\gamma = 0.999\).Solid lines denote means and shaded regions denote Bonferroni-adjusted 95\% confidence intervals across evaluation runs.}
  \label{fig:plot_3iql}
\end{figure}

\subsubsection{Offline model selection}
\label{sec:app_offline_model_selection}

\begin{figure}[H]
    \centering
    \includegraphics[width=0.65\textwidth]{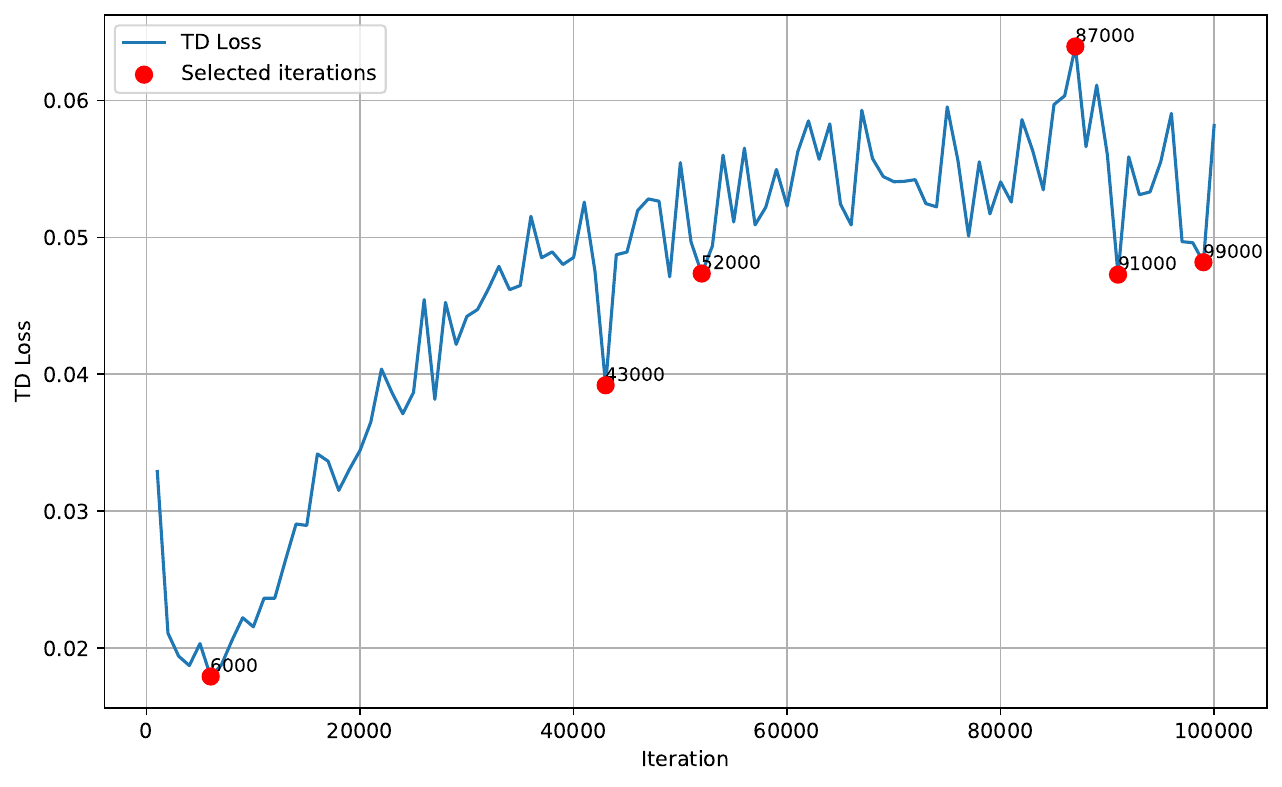} \caption{TD loss across training iterations with red bullets showing the selected checkpoints used for comparison.}
    \label{fig:td_loss_selected_checkpoints}
\end{figure}

\begin{figure}[H]
    \centering
    \includegraphics[trim=0 0 0 0,clip,width=0.65\textwidth]{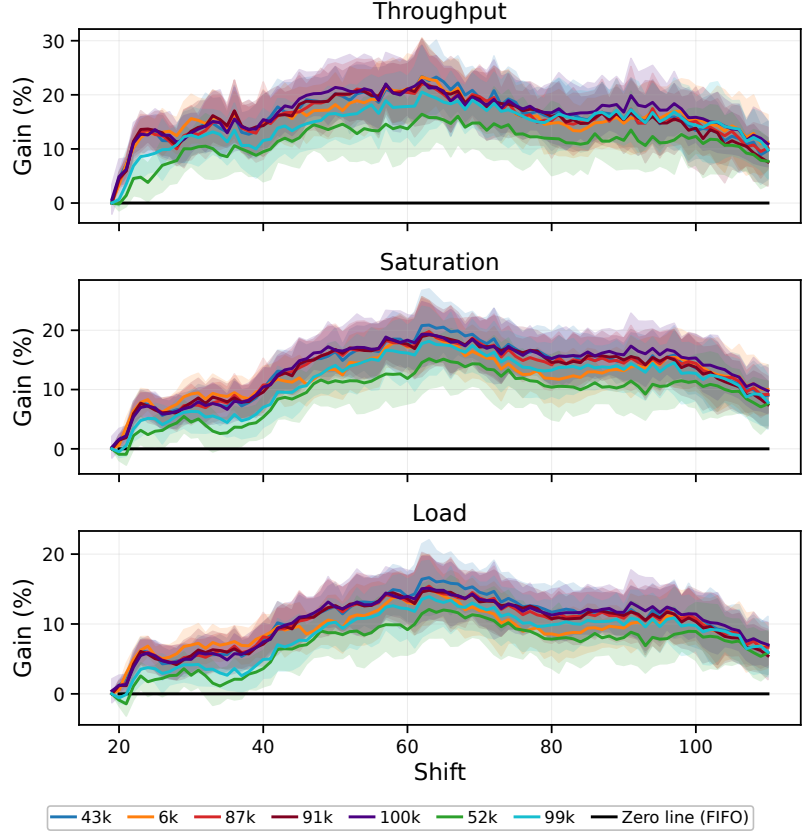}
    \caption{KPI gains (\%) of selected offline-trained DQL checkpoints relative to FIFO across shift. The black horizontal line denotes the FIFO reference. Colours identify the policies as shown in the in-panel legends. Solid lines show means and shaded regions show Bonferroni-adjusted 95\% confidence intervals. The black horizontal line denotes the FIFO reference. Colours identify the policies as shown in the in-panel legends.} 
\end{figure}

\subsubsection{CQL Variants}
\label{sec:cql_Variants}

\begin{figure}[h]
    \centering
    \includegraphics[trim=0 0 0 05,clip,width=0.65\textwidth]{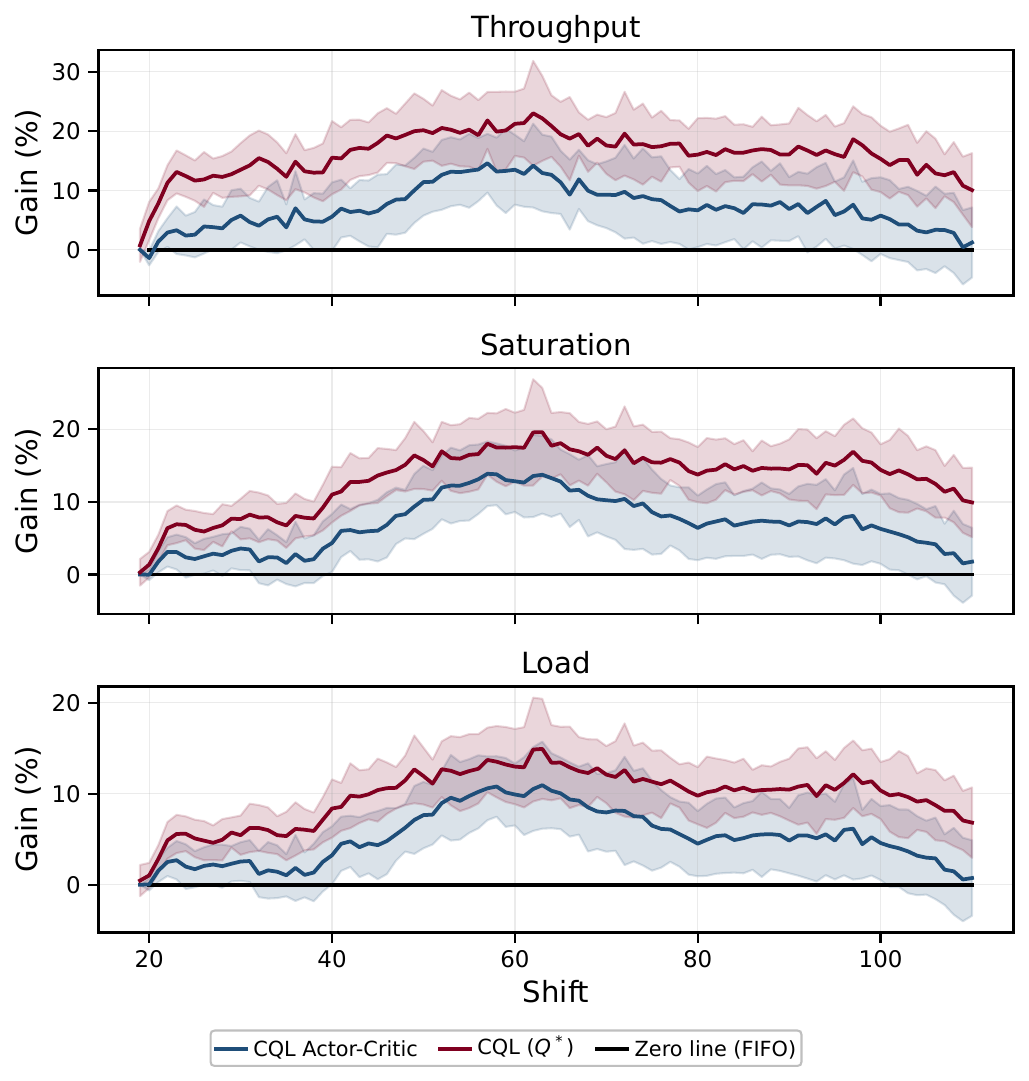}
    \caption{Comparison of the two CQL variants across shifts. The dark blue curve corresponds to Entropy-based, whereas the burgundy curve corresponds to Q-learning. Solid lines denote means and shaded regions denote Bonferroni-adjusted 95\% confidence intervals across evaluation runs.}
    \label{fig:cql_variants}
\end{figure}

To isolate the effect of the policy extraction mechanism in conservative Q-learning, we compared two offline CQL variants: an entropy-based version and a Q-learning version. As shown in Fig.~\ref{fig:cql_variants} and Table~\ref{tab:cql_variants}, the Q-learning version consistently outperforms the entropy-based version across throughput, saturation, and load. It achieves larger gains and more stable performance over the evaluation horizon, whereas the entropy-based version yields smaller improvements. This difference is expected, as the entropy-based version optimizes a maximum-entropy objective that is generally more useful during online fine-tuning, where improved exploration is beneficial. Overall, in our offline semiconductor scheduling setting, the Q-learning version provides a more effective learning signal than the entropy-based version.

\begin{table}[h]
\caption{Effect of CQL objective choice on KPI gains. Values are percentage gains in throughput, saturation and load relative to FIFO for CQL variants trained with different objectives, reported as mean $\pm$ Bonferroni-adjusted 95\% confidence-interval evaluation runs.}
\label{tab:cql_variants}
\centering
\begin{tabular}{lccc}
\midrule
 \textbf{Variant} & \textbf{Throughput} & \textbf{Saturation} & \textbf{Load} \\
\midrule
 $Q$ Learning & 17.5 $\pm$ 6.3 & 15.7 $\pm$ 5.0 & 12.3 $\pm$ 4.1 \\
 Actor Critic (Entropy-Based) & 9.5 $\pm$ 5.8 & 10.2 $\pm$ 4.8 & 8.3 $\pm$ 4.0 \\
\hline
\end{tabular}
\vspace{2mm}
\end{table}

\subsection{Online}

\subsubsection{Event-driven aggregation}
\label{sec:online_agg}

We further examine the effect of the aggregation rule used in online learning. For DQL, we compare the mean aggregation in Eq.~\ref{eq:lm} with the weighted aggregation in Eq.~\ref{eq:lw} shown in Table~\ref{tab:dql_sum_vs_weighted}. For reference, we also report the performance of SAC with mean aggregation.

\begin{table}[h]
\caption{Online aggregation variants for DQL and SAC. Values are percentage KPI gains relative to FIFO for online aggregation variants, reported as mean $\pm$ Bonferroni-adjusted 95\% confidence-interval across evaluation runs. DQL with weighted aggregation is compared with DQL with mean aggregation, and SAC with mean aggregation is included as a reference.}
\label{tab:dql_sum_vs_weighted}
\centering
\begin{tabular}{lcccc}
\toprule
\textbf{Agent} & \textbf{Aggregation} & \textbf{Throughput} & \textbf{Saturation} & \textbf{Load} \\
\midrule
DQL & Mean     & $14.9 \pm 5.8$ & $14.8 \pm 4.2$ & $11.7 \pm 3.6$ \\
DQL & Weighted & $19.7 \pm 4.8$ & $16.6 \pm 3.9$ & $13.2 \pm 3.4$ \\
\midrule
SAC & Mean     & $20.7 \pm 5.1$ & $17.5 \pm 4.3$ & $13.8 \pm 3.6$ \\
\bottomrule
\end{tabular}
\vspace{2mm}
\end{table}

Although the weighted formulation achieves better performance in this setting, it exhibits less stable behavior in practice and can lead to training failure. Moreover, its performance is highly sensitive to the training procedure and hyperparameter choices. We therefore adopt mean aggregation for the online agents (i.e., PPO and SAC) because it provides a more reliable learning signal.

\newpage

\subsection{Performance Across Different Scenarios}
\begin{figure}[H]
    \centering
    \begin{subfigure}[t]{0.49\textwidth}
        \centering
        \includegraphics[width=\textwidth]{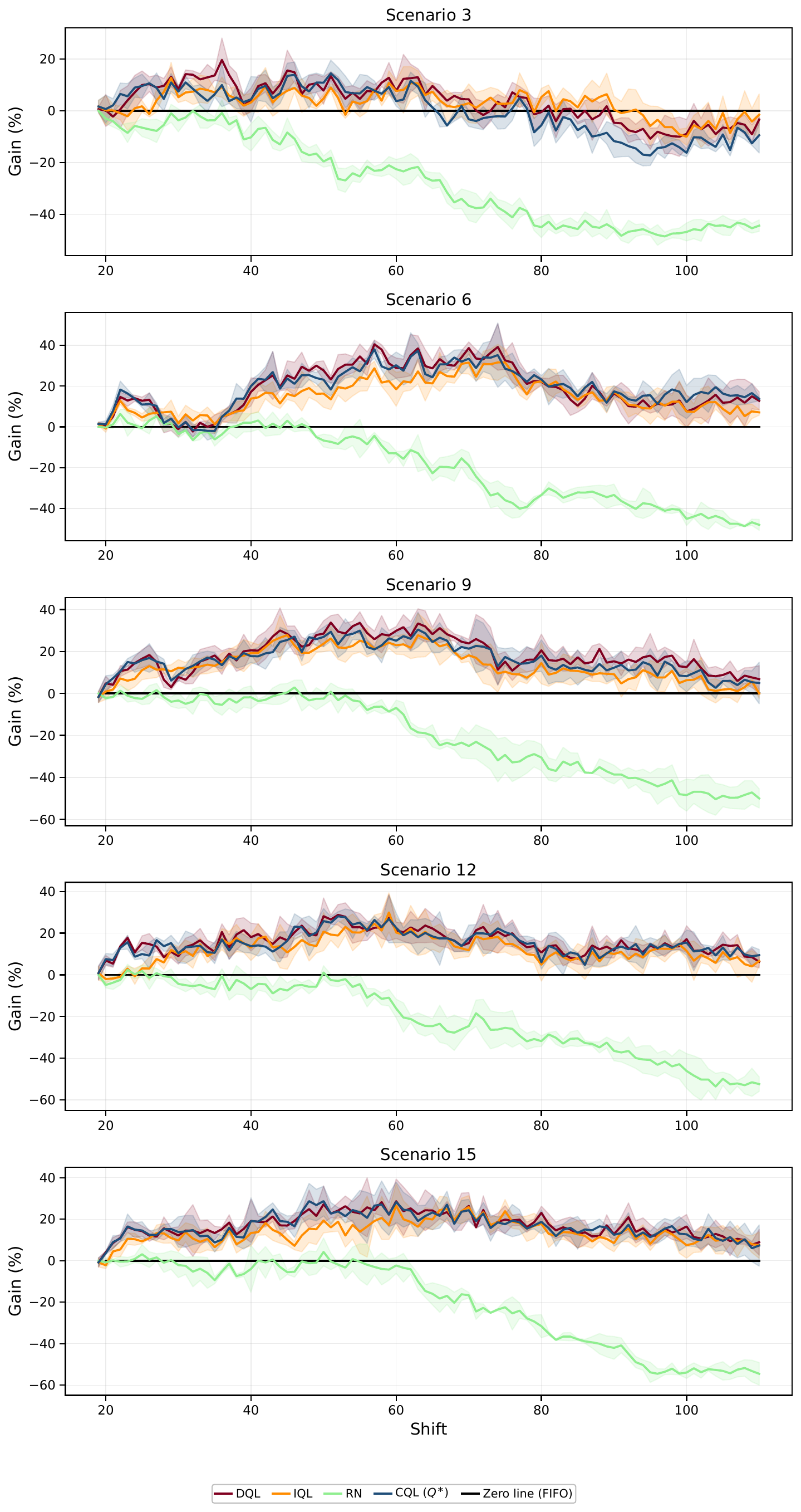}
        \caption{Against Random: Scenarios 3, 6, 9, 12, and 15}
    \end{subfigure}
    \hfill
    \begin{subfigure}[t]{0.49\textwidth}
        \centering
        \includegraphics[width=\textwidth]{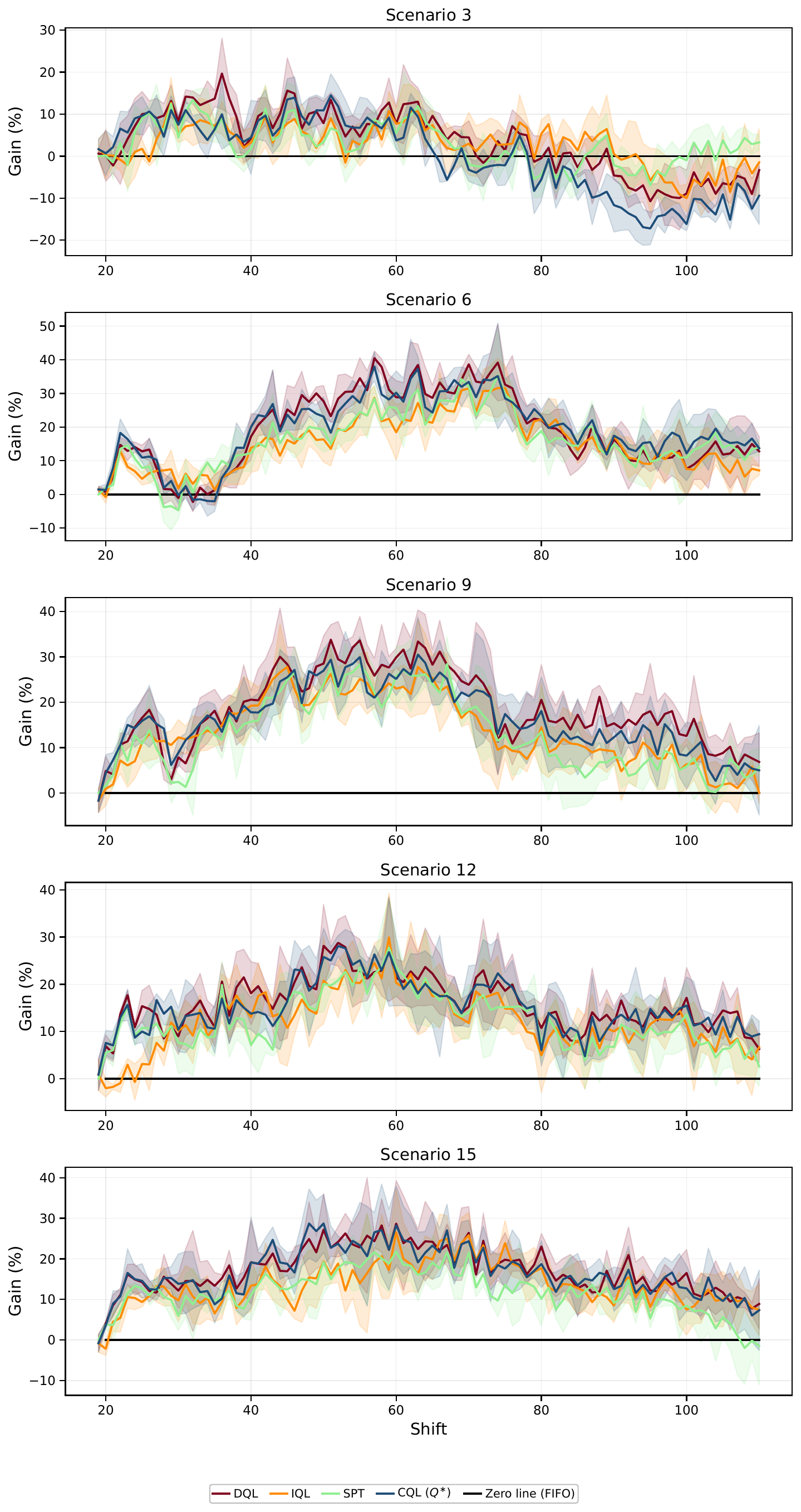}
        \caption{Against SPT: Scenarios 3, 6, 9, 12, and 15}
    \end{subfigure}
    \caption{Scenario-by-scenario Throughput gain (\%) of the offline agents relative to the FIFO baseline for the first group of evaluation scenarios. The left panel shows the Random policy and the right panel shows the SPT policy, following the comparative layout used in Figure \ref{fig:offline_spt-rn-side-by-side}.}
    \label{fig:offline-scen-throughput-g1}
\end{figure}

\begin{figure}[H]
    \centering
    \begin{subfigure}[t]{0.49\textwidth}
        \centering
        \includegraphics[width=\textwidth]{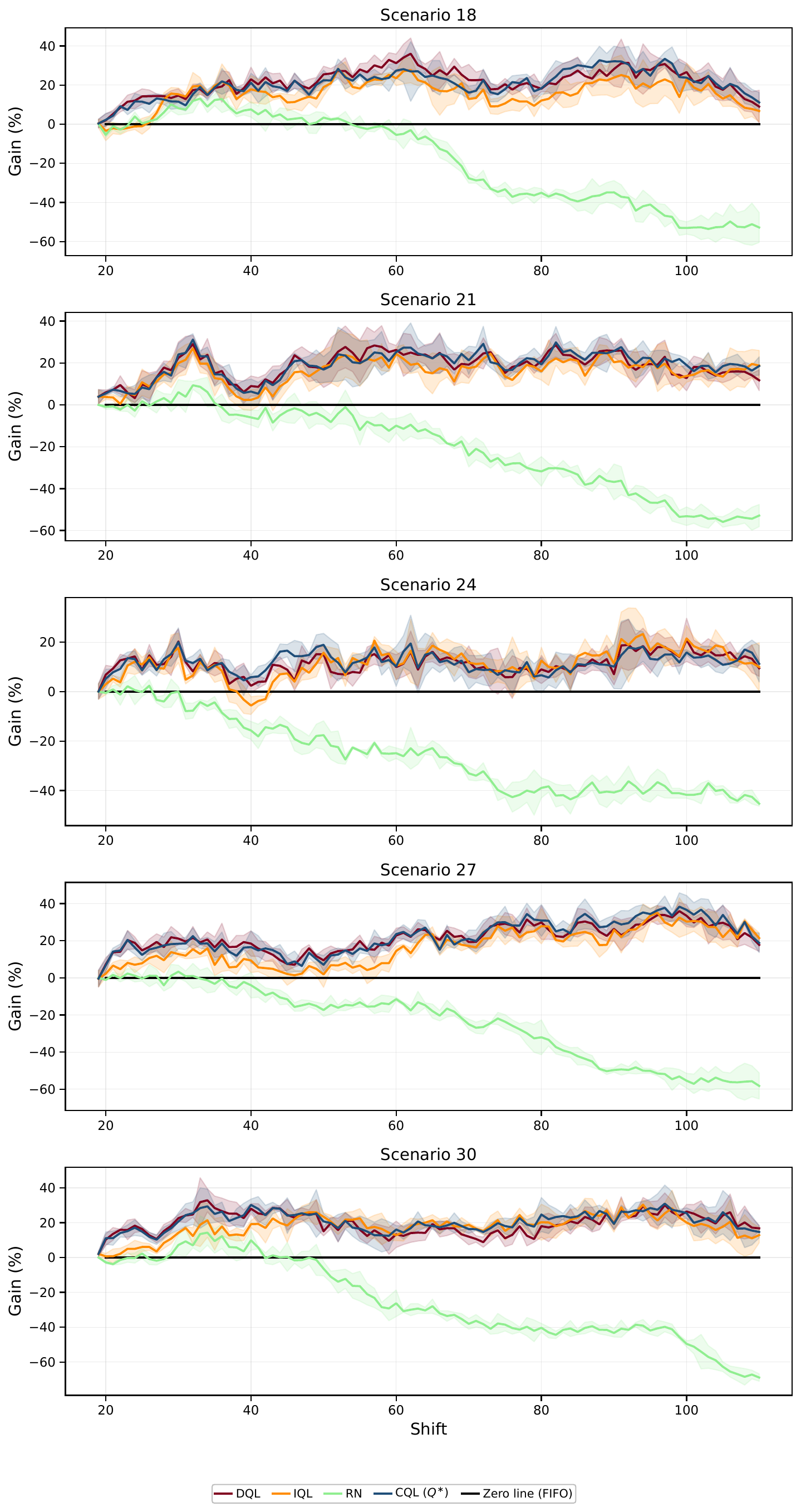}
        \caption{Against Random: Scenarios 18, 21, 24, 27, and 30}
    \end{subfigure}
    \hfill
    \begin{subfigure}[t]{0.49\textwidth}
        \centering
        \includegraphics[width=\textwidth]{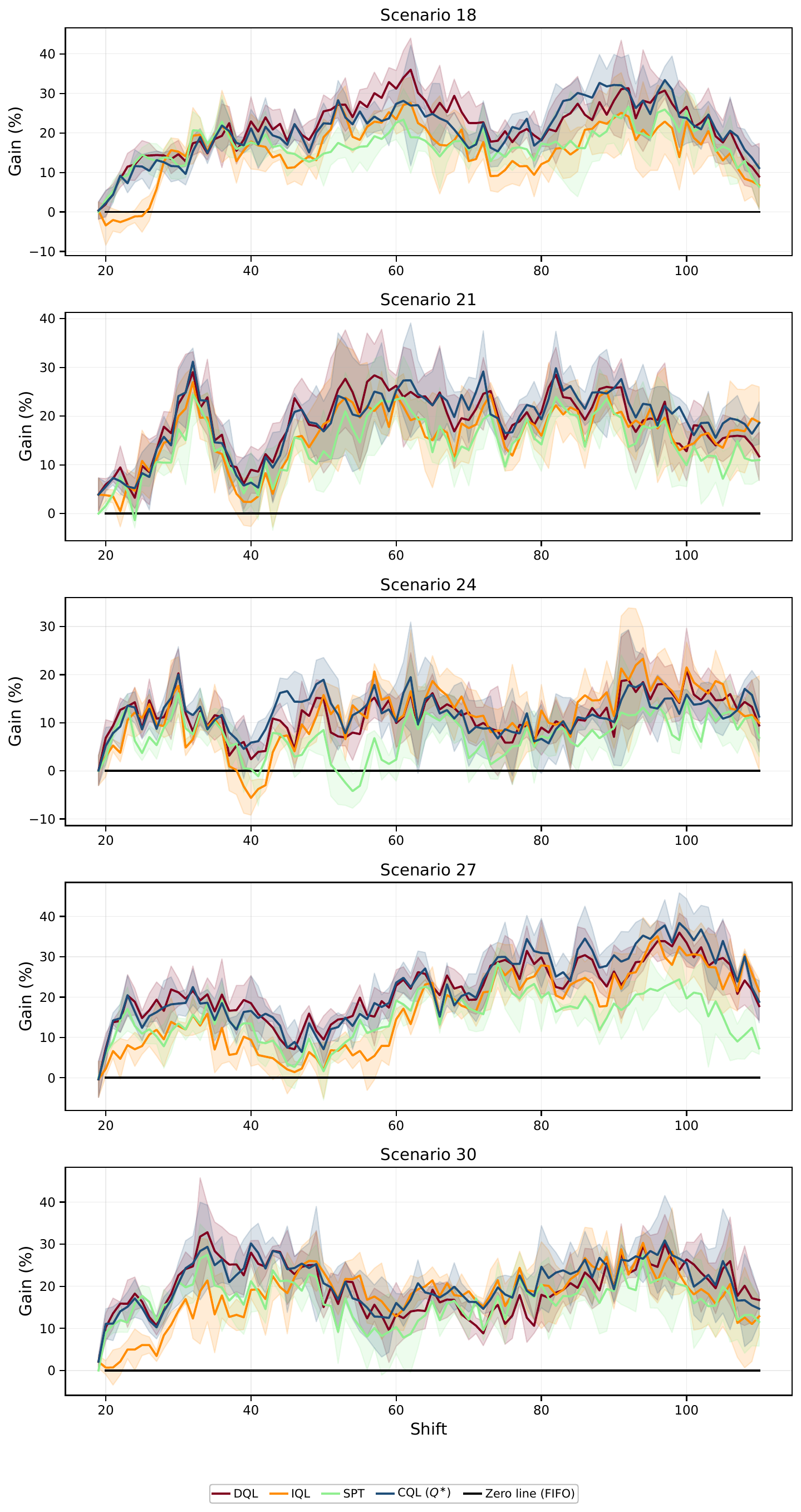}
        \caption{Against SPT: Scenarios 18, 21, 24, 27, and 30}
    \end{subfigure}
    \caption{Scenario-by-scenario Throughput gain (\%) of the offline agents relative to the FIFO baseline for the second group of evaluation scenarios, extending the comparison across all scenarios.}
    \label{fig:offline-scen-throughput-g2}
\end{figure}

\begin{figure}[H]
    \centering
    \begin{subfigure}[t]{0.49\textwidth}
        \centering
        \includegraphics[width=\textwidth]{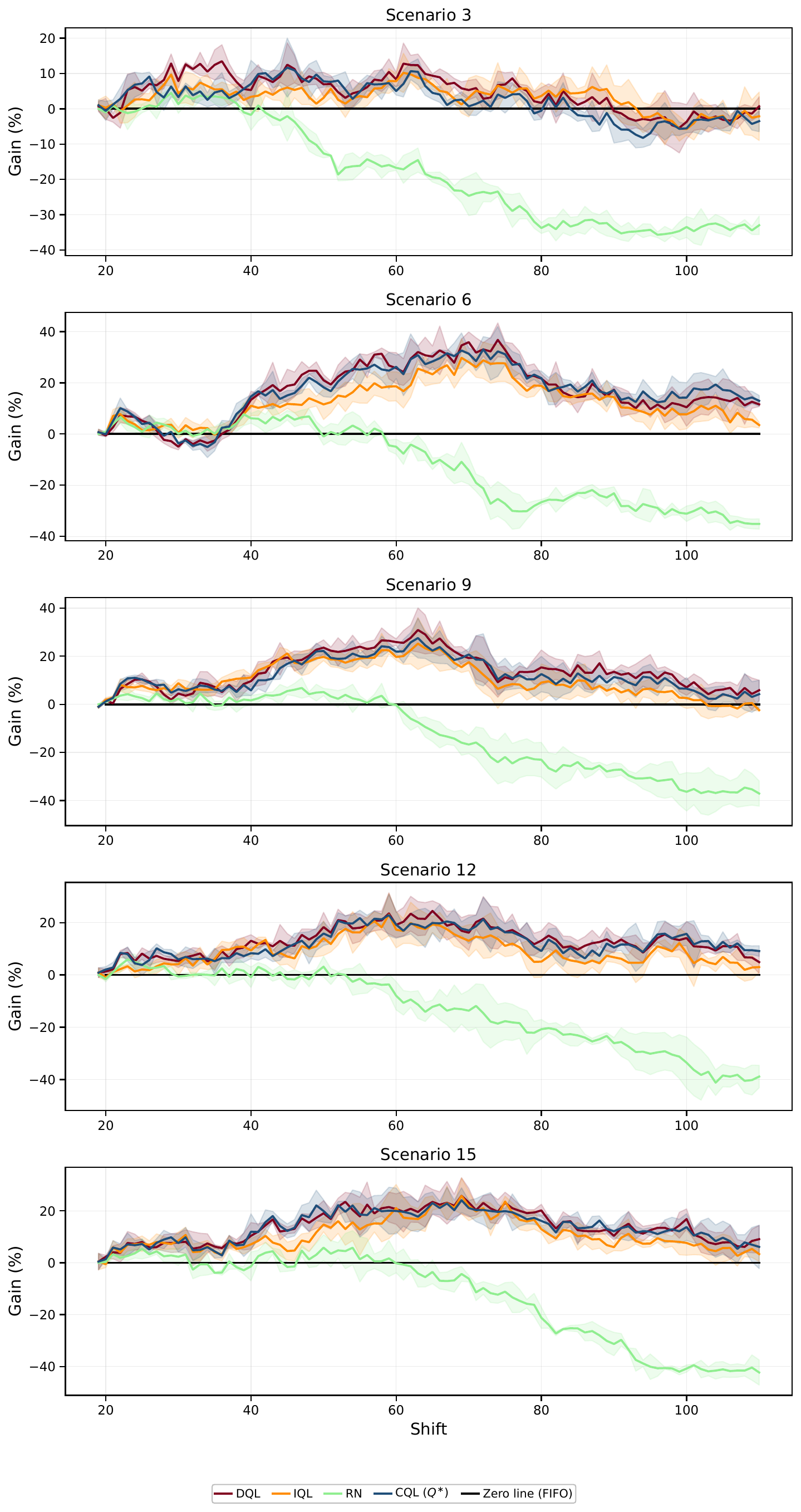}
        \caption{Against Random: Scenarios 3, 6, 9, 12, and 15}
    \end{subfigure}
    \hfill
    \begin{subfigure}[t]{0.49\textwidth}
        \centering
        \includegraphics[width=\textwidth]{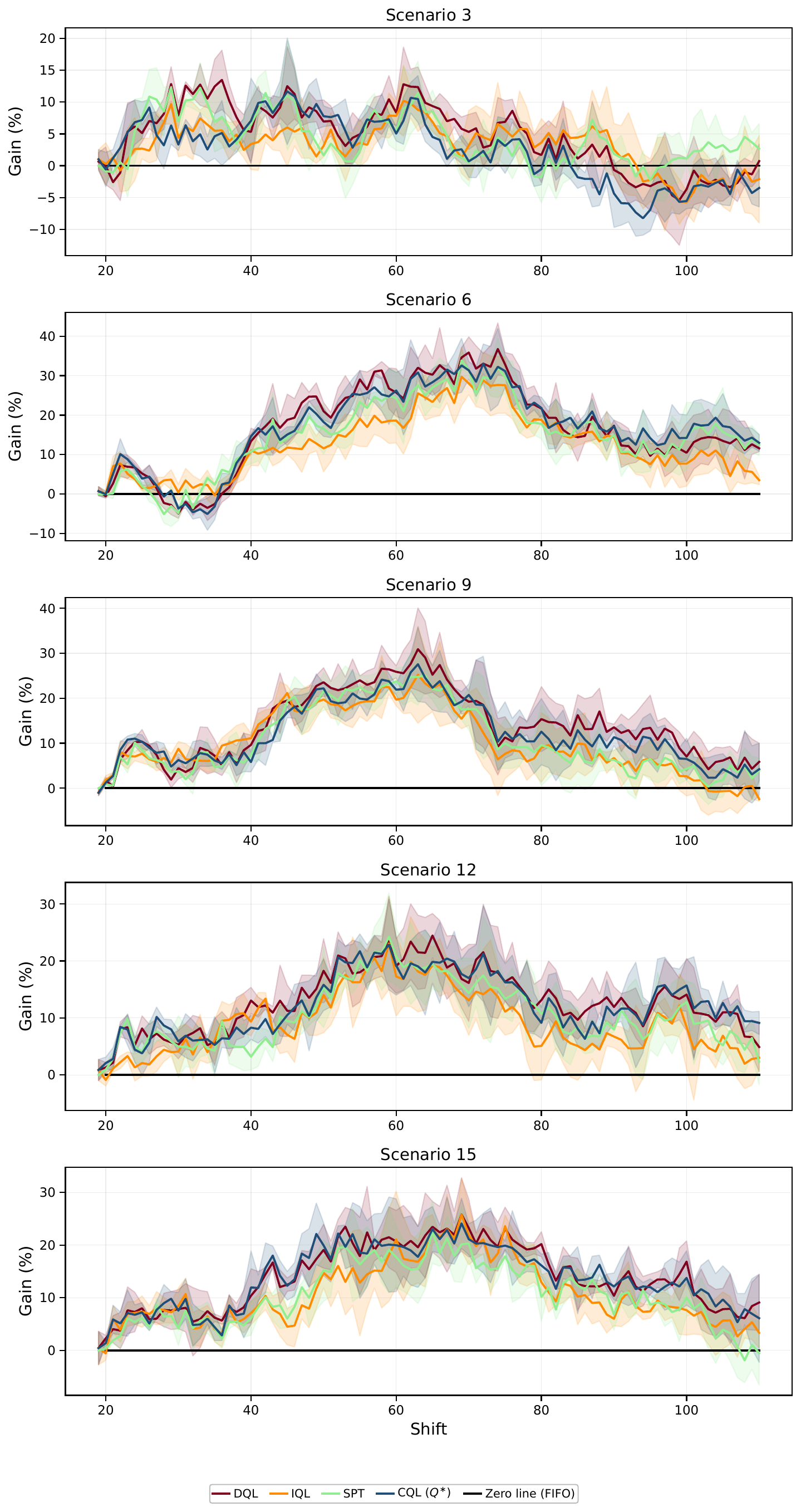}
        \caption{Against SPT: Scenarios 3, 6, 9, 12, and 15}
    \end{subfigure}
    \caption{Scenario-by-scenario Saturation gain (\%) of the offline agents relative to the FIFO baseline for the second group of evaluation scenarios, extending the comparison across all scenarios.}
    \label{fig:offline-scen-saturation-g1}
\end{figure}

\begin{figure}[H]
    \centering
    \begin{subfigure}[t]{0.49\textwidth}
        \centering
        \includegraphics[width=\textwidth]{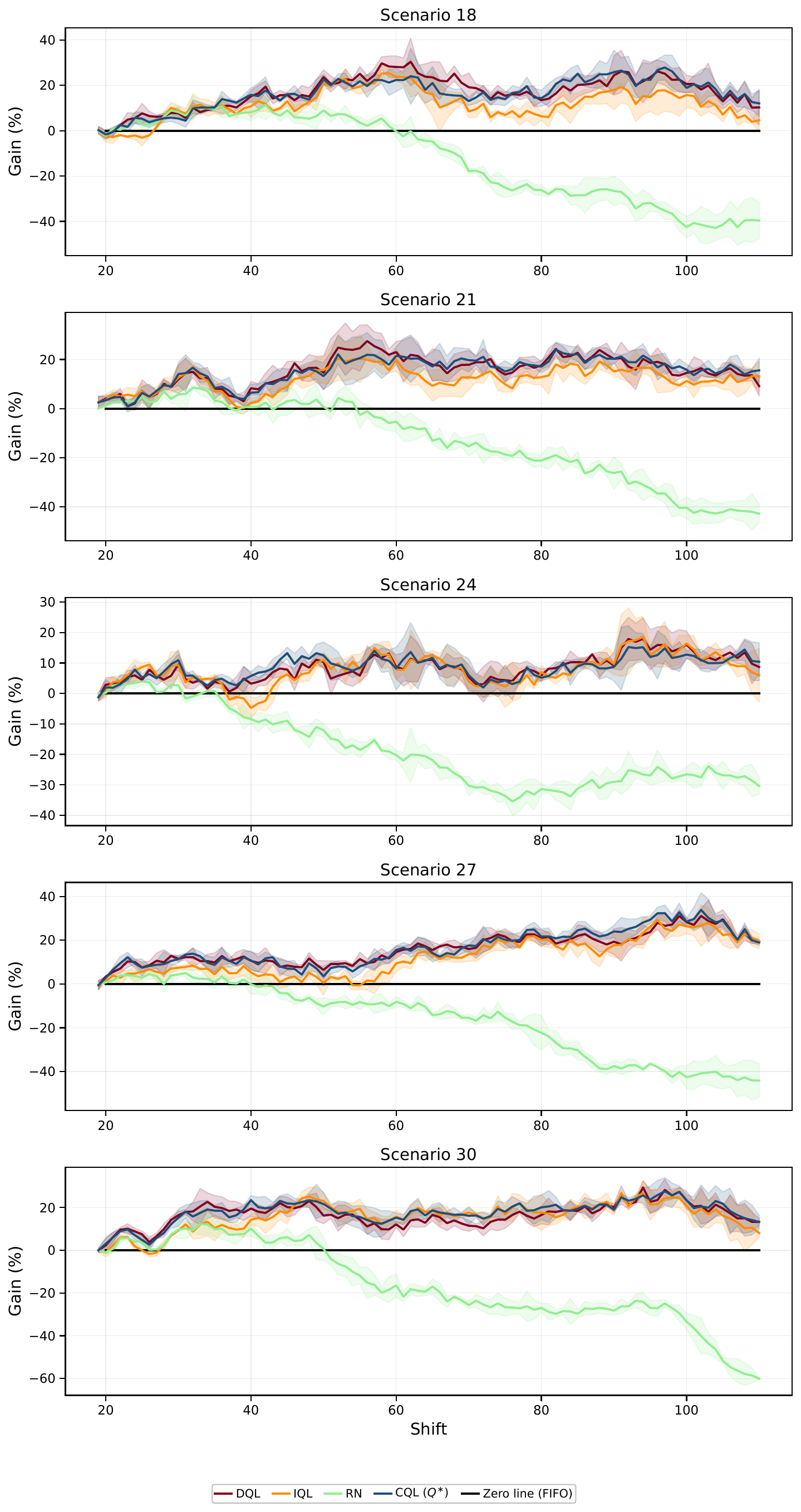}
        \caption{Against Random: Scenarios 18, 21, 24, 27, and 30}
    \end{subfigure}
    \hfill
    \begin{subfigure}[t]{0.49\textwidth}
        \centering
        \includegraphics[width=\textwidth]{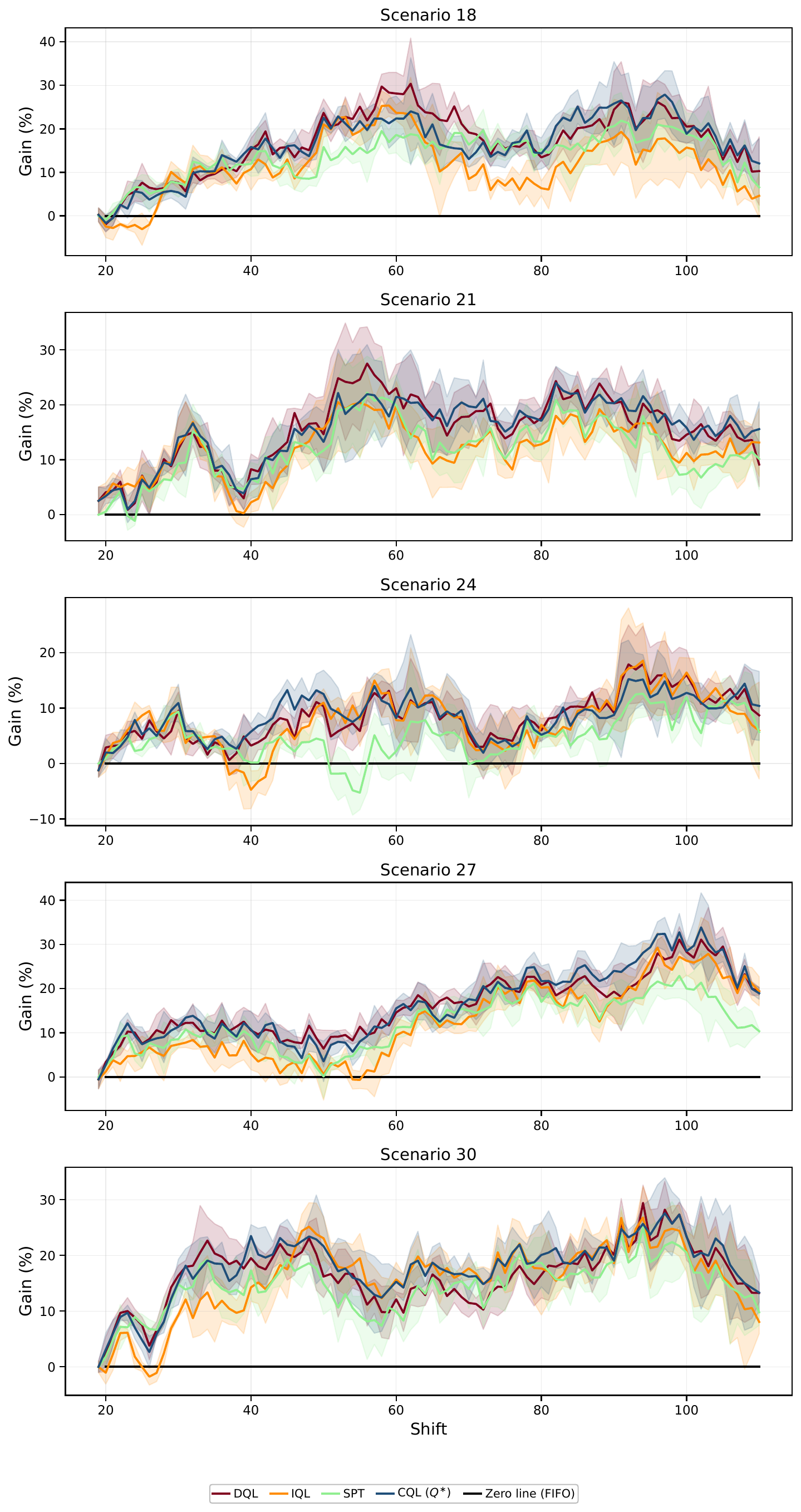}
        \caption{Against SPT: Scenarios 18, 21, 24, 27, and 30}
    \end{subfigure}
    \caption{Scenario-by-scenario Saturation gain (\%) of the offline agents relative to the FIFO baseline for the second group of evaluation scenarios, extending the comparison across all scenarios.}
    \label{fig:offline-scen-saturation-g2}
\end{figure}

\begin{figure}[H]
    \centering
    \begin{subfigure}[t]{0.49\textwidth}
        \centering
        \includegraphics[width=\textwidth]{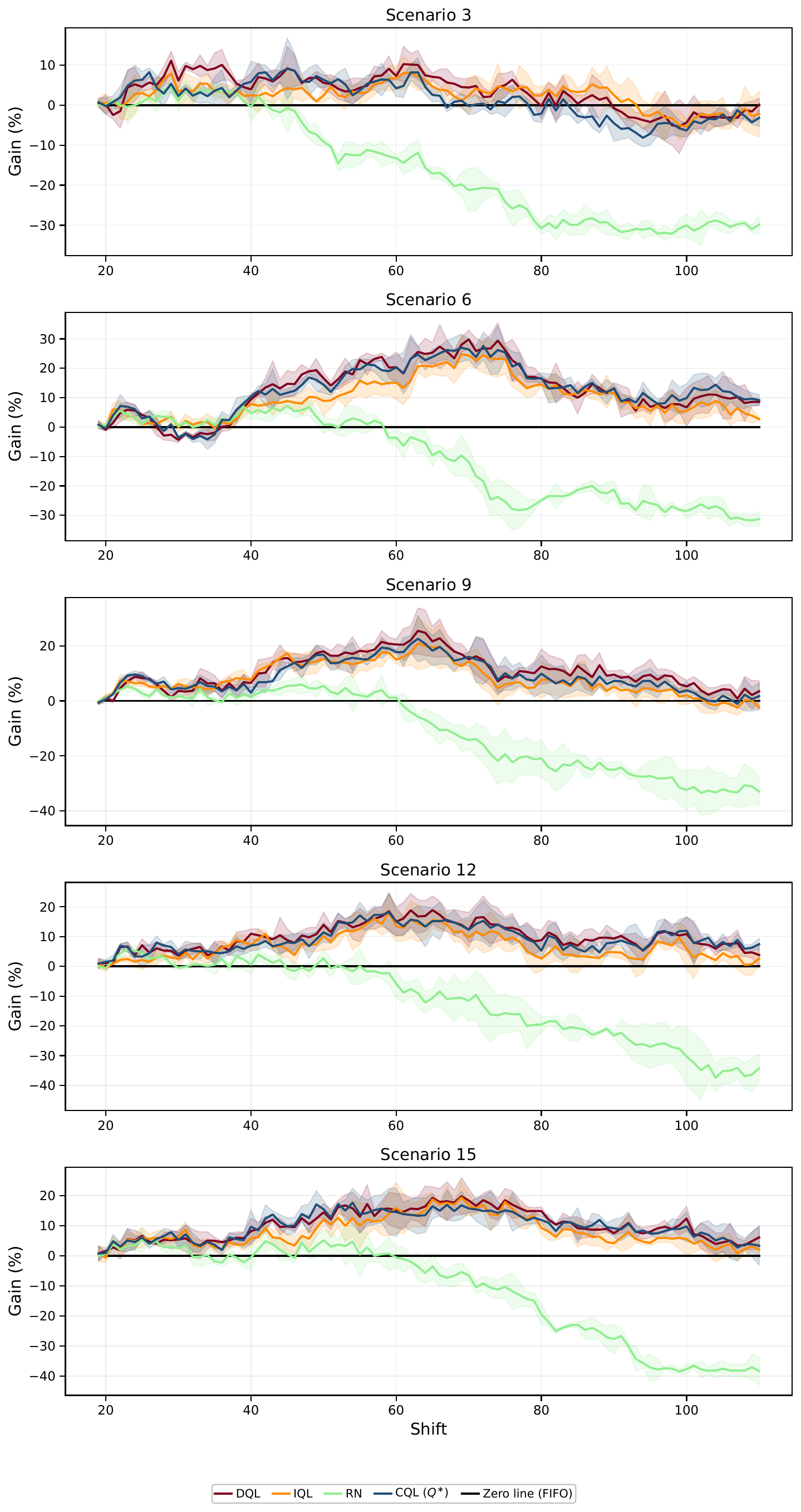}
        \caption{Against Random: Scenarios 3, 6, 9, 12, and 15}
    \end{subfigure}
    \hfill
    \begin{subfigure}[t]{0.49\textwidth}
        \centering
        \includegraphics[width=\textwidth]{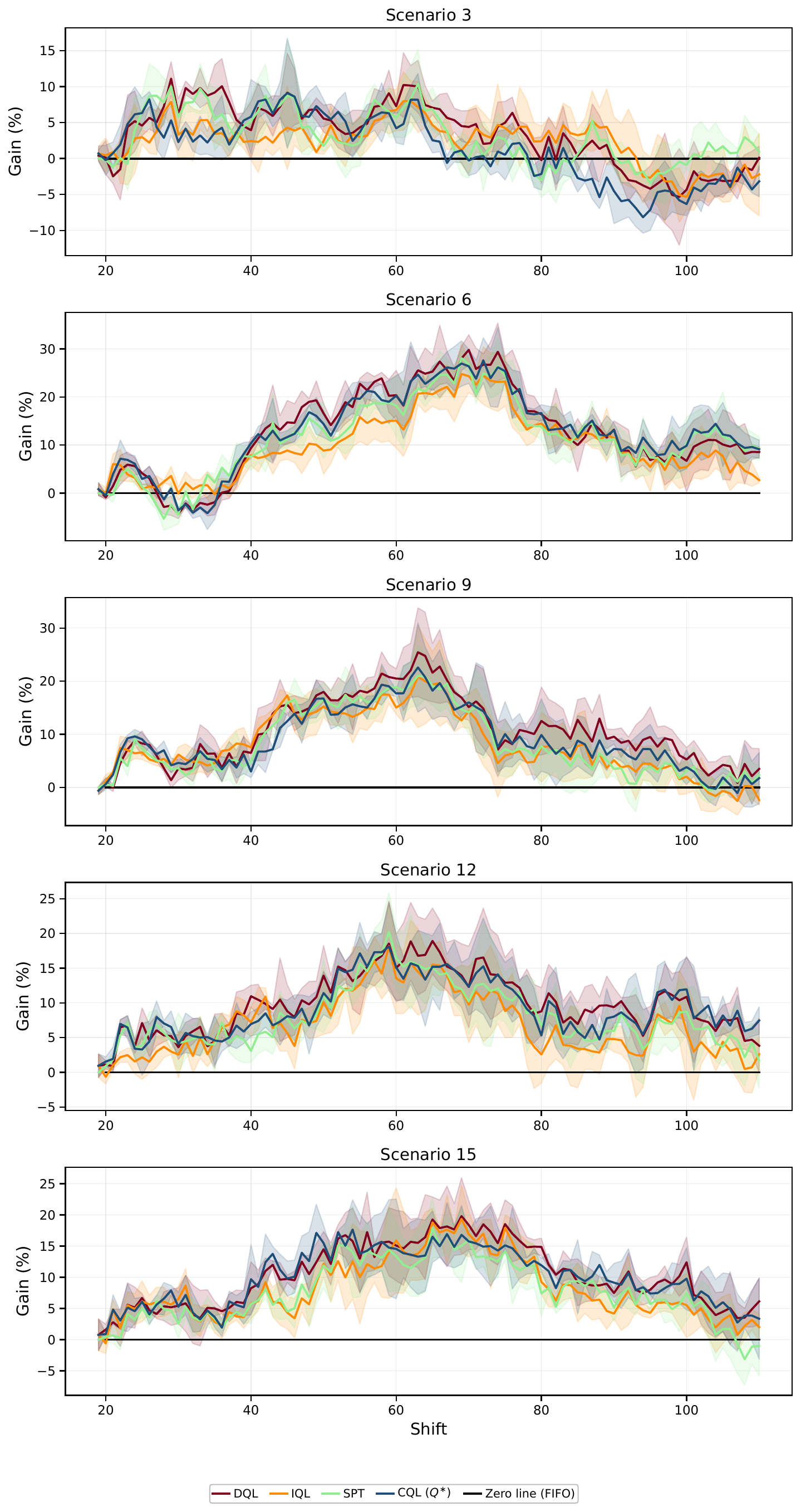}
        \caption{Against SPT: Scenarios 3, 6, 9, 12, and 15}
    \end{subfigure}
    \caption{Scenario-by-scenario Load gain (\%) of the offline agents relative to the FIFO baseline for the second group of evaluation scenarios, extending the comparison across all scenarios..}
    \label{fig:offline-scen-windowload-g1}
\end{figure}

\begin{figure}[H]
    \centering
    \begin{subfigure}[t]{0.49\textwidth}
        \centering
        \includegraphics[width=\textwidth]{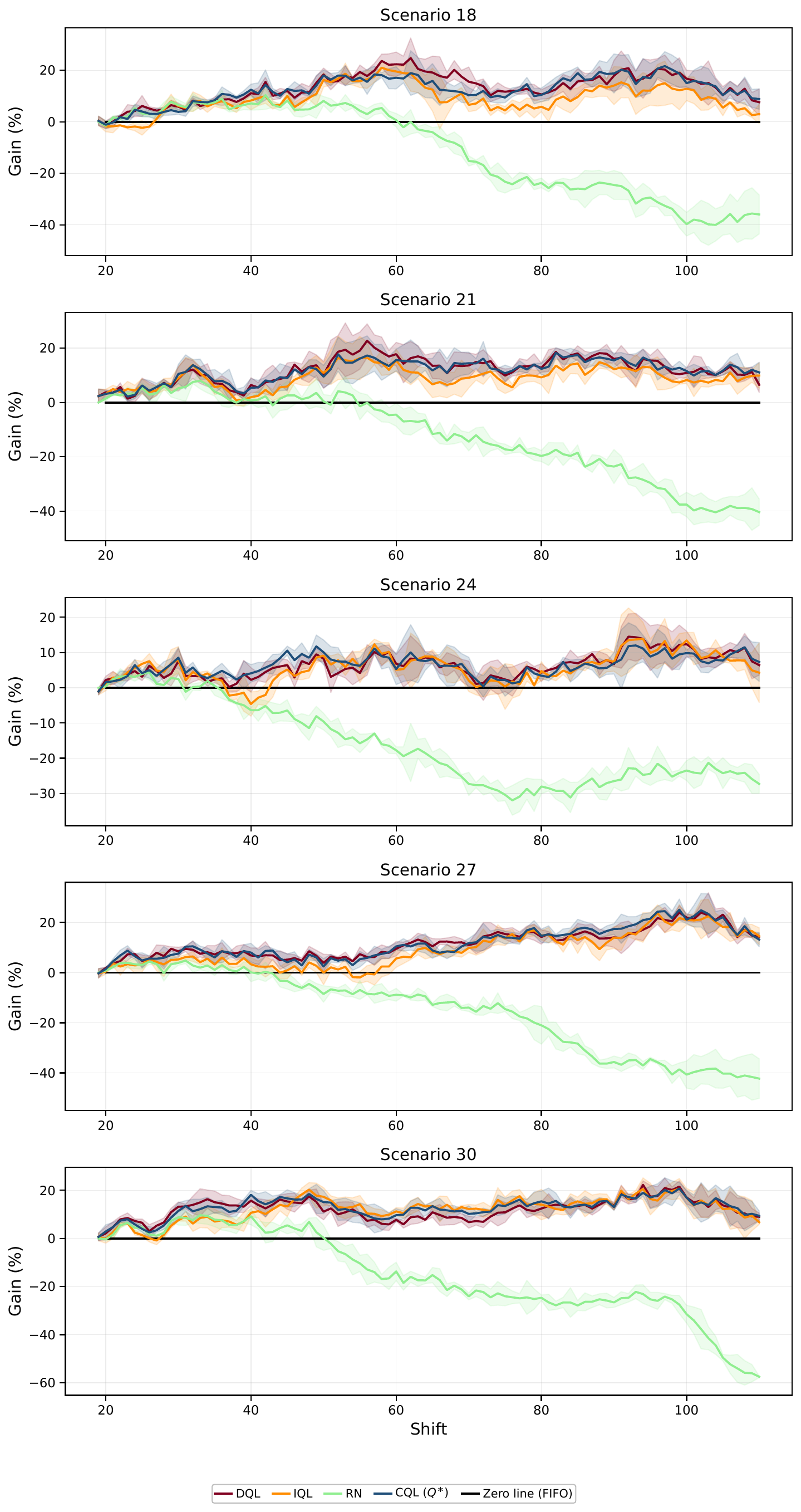}
        \caption{Against Random: Scenarios 18, 21, 24, 27, and 30}
    \end{subfigure}
    \hfill
    \begin{subfigure}[t]{0.49\textwidth}
        \centering
        \includegraphics[width=\textwidth]{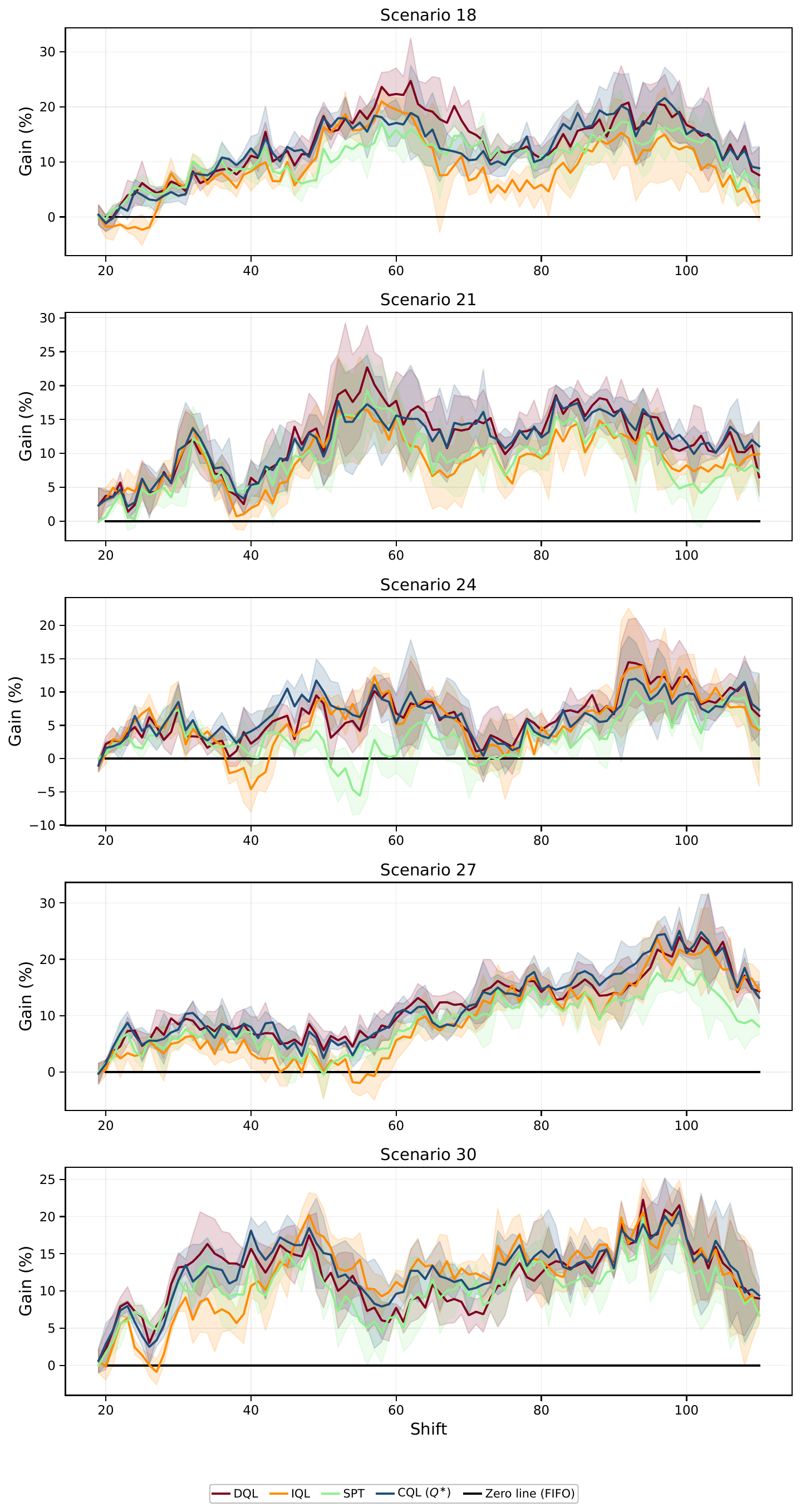}
        \caption{Against SPT: Scenarios 18, 21, 24, 27, and 30}
    \end{subfigure}
    \caption{Scenario-by-scenario Load gain (\%) of the offline agents relative to the FIFO baseline for the second group of evaluation scenarios, extending the comparison across all scenarios.}
    \label{fig:offline-scen-windowload-g2}
\end{figure}

\paragraph{Offline Agents}
\label{sec:offline_performance_scenarios}
To complement the aggregated offline results reported in section \ref{subsubsec:offline}, Figures~\ref{fig:offline-scen-throughput-g1}--\ref{fig:offline-scen-windowload-g2} present a more detailed scenario-by-scenario analysis of the KPI gains relative to the FIFO baseline, organized in the same side-by-side layout as Figure \ref{fig:offline_spt-rn-side-by-side}. For each KPI, the left panel shows the Random setting, and the right panel shows the SPT setting. To improve readability, the ten evaluation scenarios are divided into two figure groups. The first group includes Scenarios~3, 6, 9, 12, and 15, while the second group includes Scenarios~18, 21, 24, 27, and 30. This organization facilitates comparison of the offline agents across baselines, scenarios, and shift bins for each KPI. While Table~\ref{tab:20_offline_agents} reports the mean performance across all evaluated scenarios, these grouped figures provide a finer-grained view of how consistently the observed gains are achieved under different operating conditions. 

\newpage

\begin{figure}[H]
    \centering
    \begin{subfigure}[t]{0.49\textwidth}
        \centering
        \includegraphics[width=\textwidth]{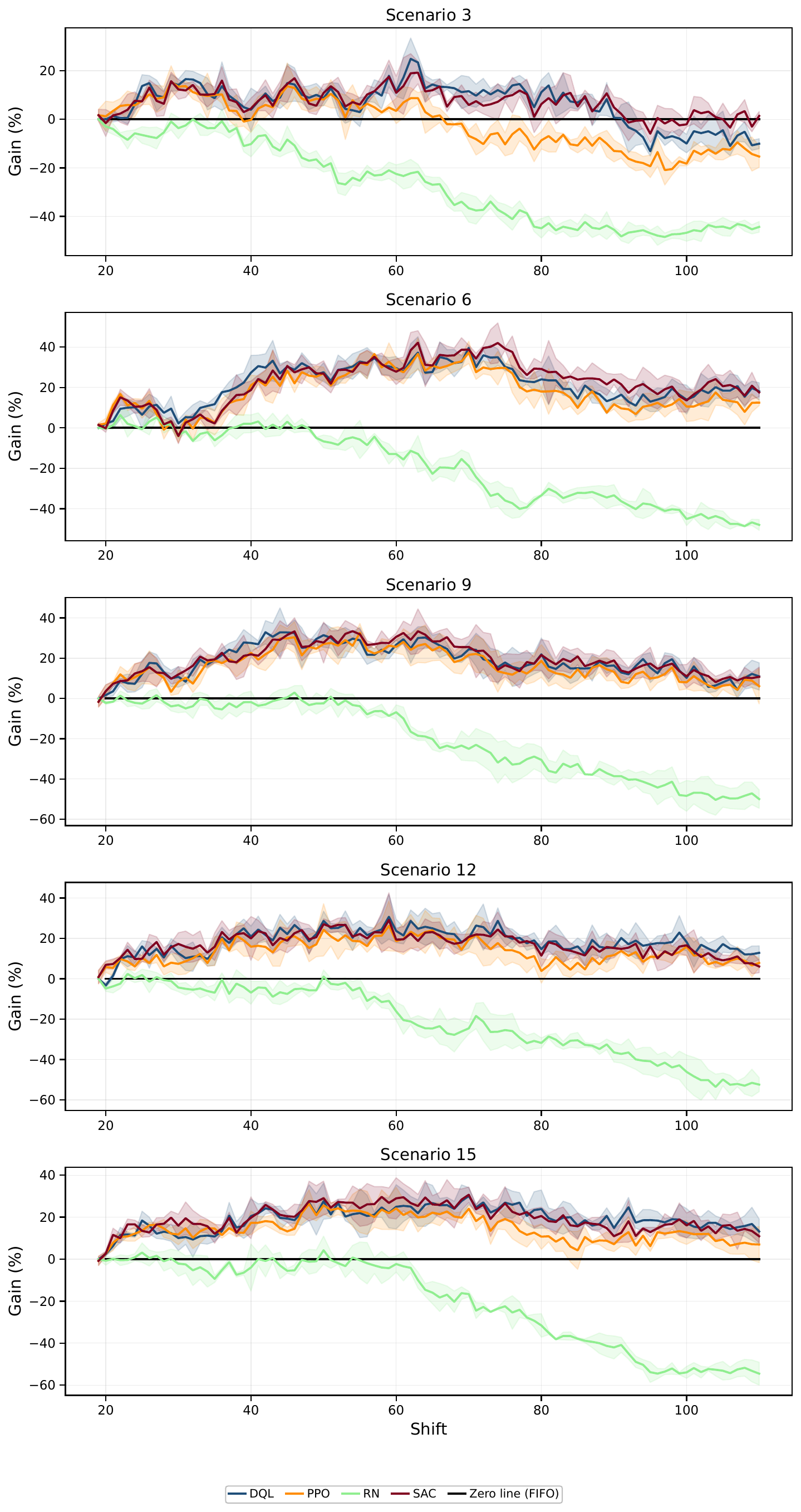}
        \caption{Against Random: Scenarios 3, 6, 9, 12, and 15}
    \end{subfigure}
    \hfill
    \begin{subfigure}[t]{0.49\textwidth}
        \centering
        \includegraphics[width=\textwidth]{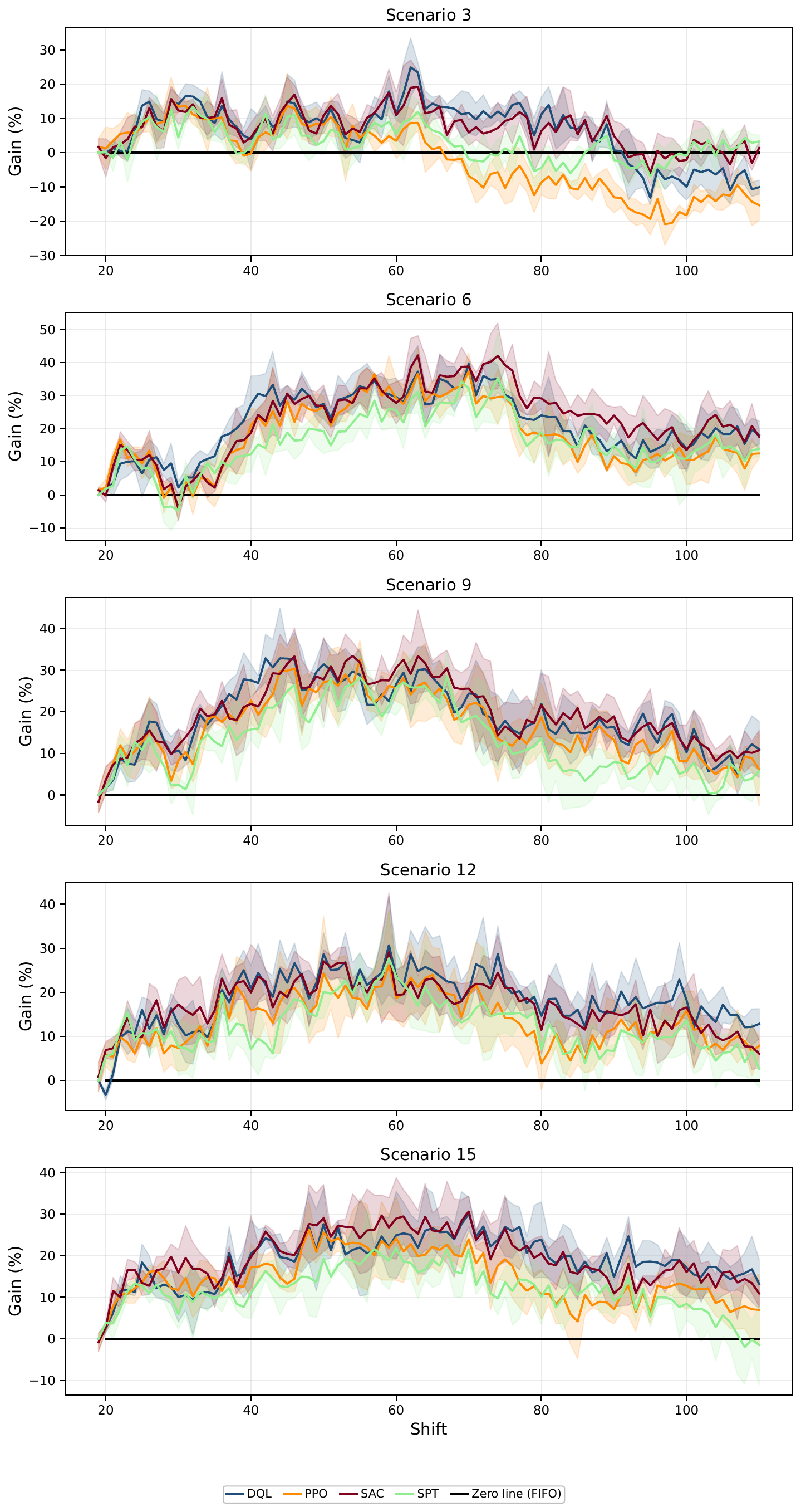}
        \caption{Against SPT: Scenarios 3, 6, 9, 12, and 15}
    \end{subfigure}
    \caption{Scenario-by-scenario Throughput gain (\%) of the online agents relative to the FIFO baseline for the first group of evaluation scenarios. The left panel shows results under the Random policy, while the right panel corresponds to the SPT policy.}
    \label{fig:online-scen-throughput-g1}
\end{figure}

\begin{figure}[H]
    \centering
    \begin{subfigure}[t]{0.49\textwidth}
        \centering
        \includegraphics[width=\textwidth]{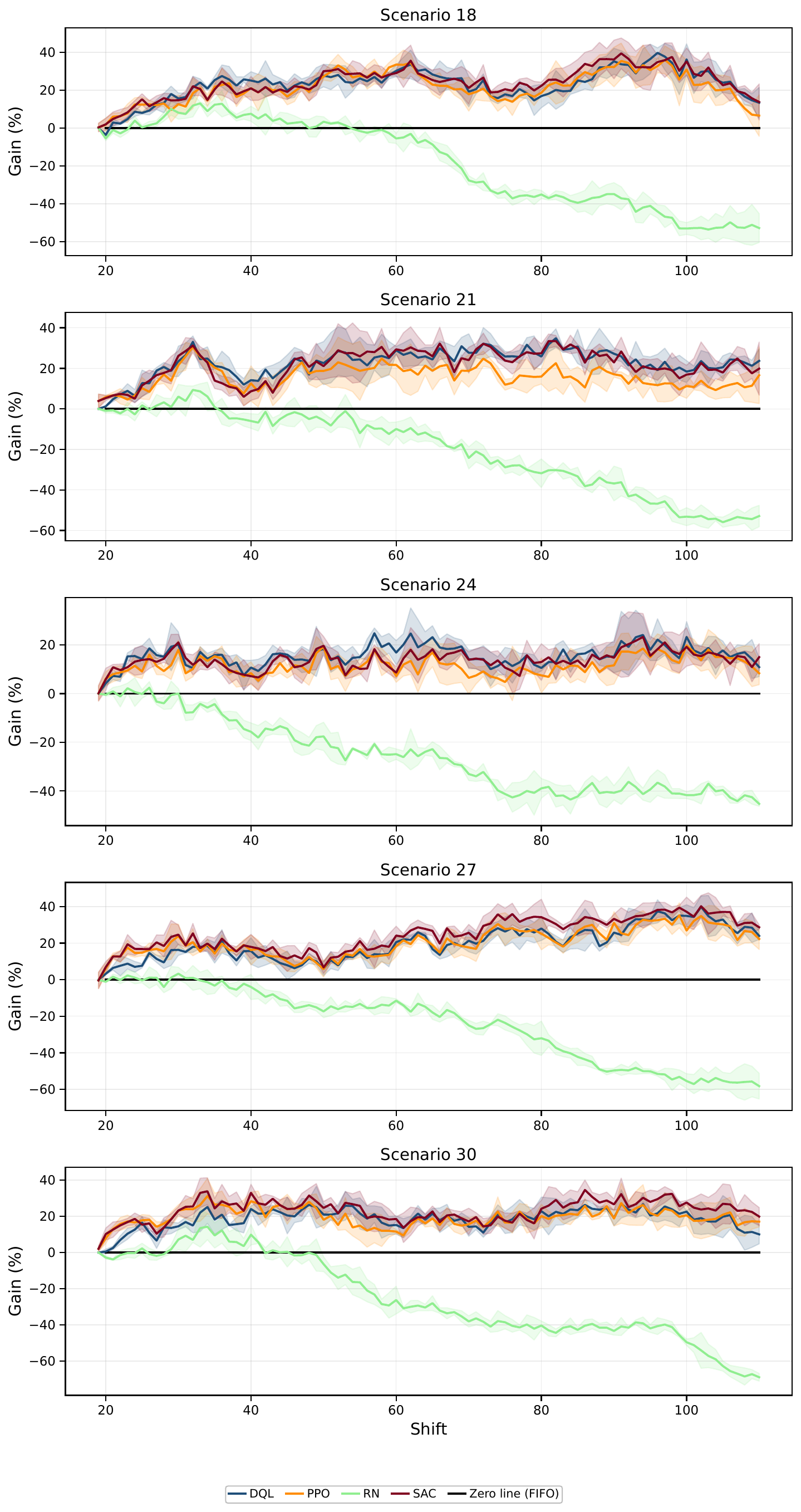}
        \caption{Against Random: Scenarios 18, 21, 24, 27, and 30}
    \end{subfigure}
    \hfill
    \begin{subfigure}[t]{0.49\textwidth}
        \centering
        \includegraphics[width=\textwidth]{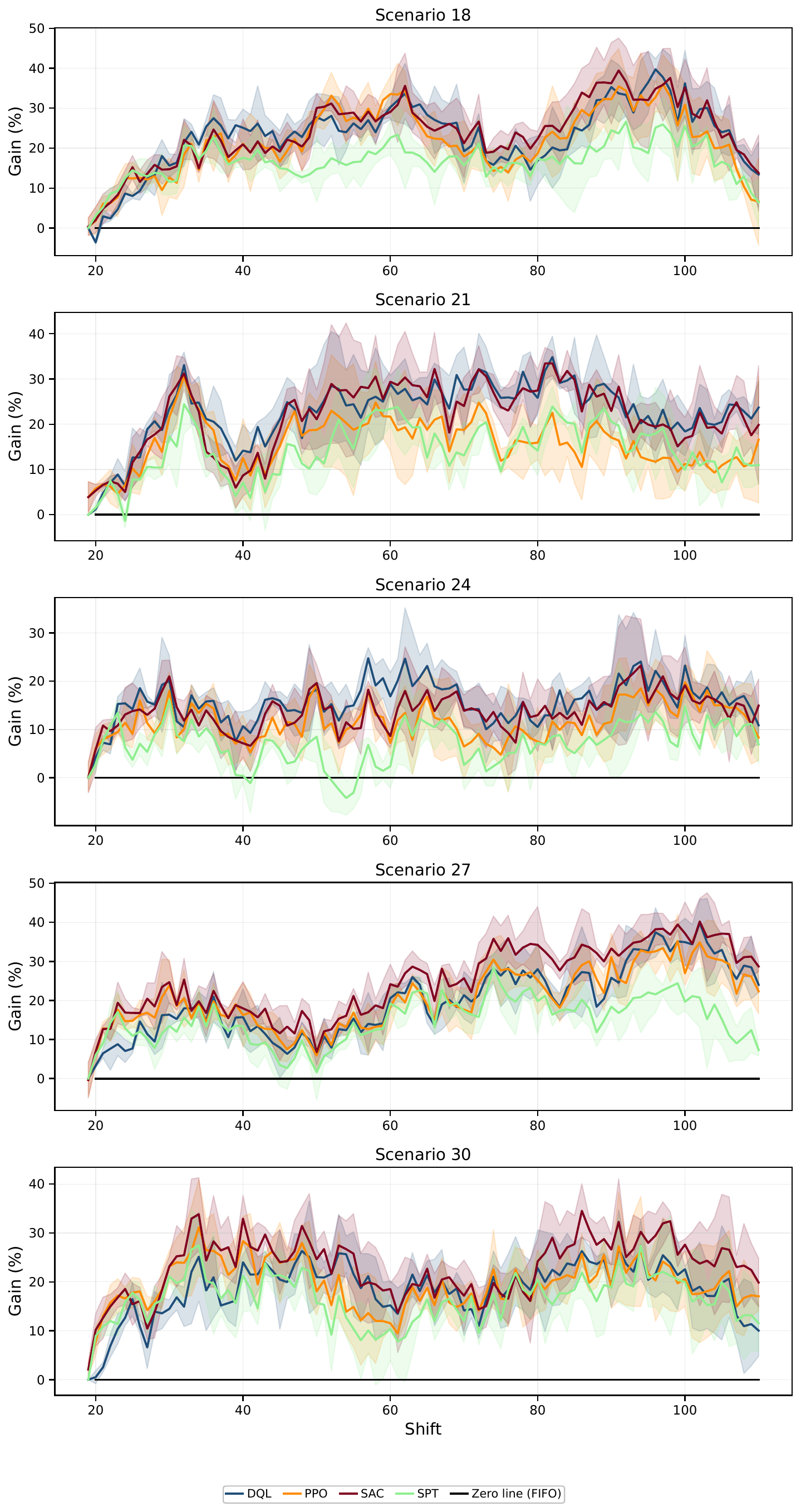}
        \caption{Against SPT: Scenarios 18, 21, 24, 27, and 30}
    \end{subfigure}
    \caption{Scenario-by-scenario Throughput gain (\%) of the online agents relative to the FIFO baseline for the second group of evaluation scenarios, extending the comparison across all scenarios.}
    \label{fig:online-scen-throughput-g2}
\end{figure}

\begin{figure}[H]
    \centering
    \begin{subfigure}[t]{0.49\textwidth}
        \centering
        \includegraphics[width=\textwidth]{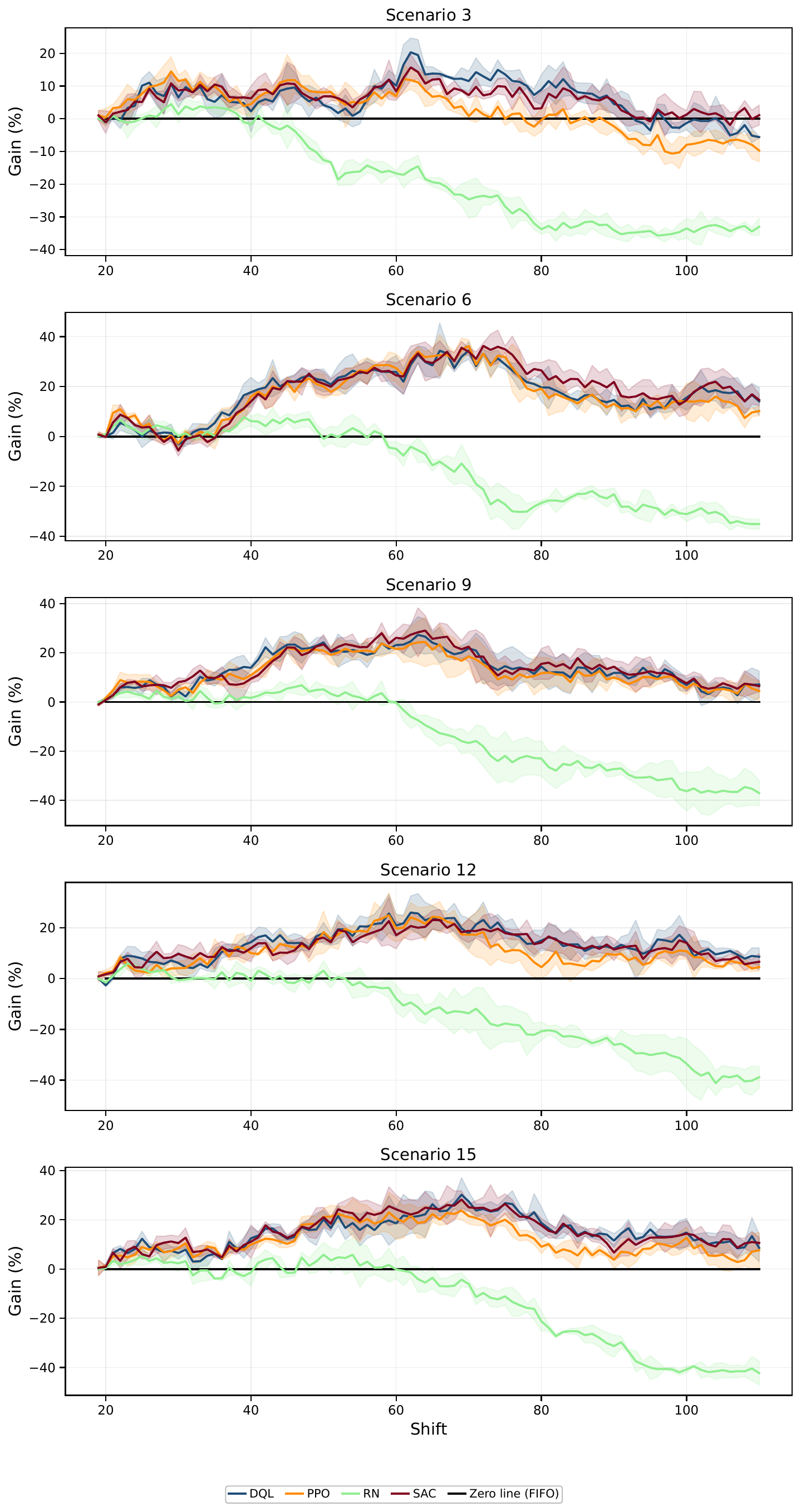}
        \caption{Against Random: Scenarios 3, 6, 9, 12, and 15}
    \end{subfigure}
    \hfill
    \begin{subfigure}[t]{0.49\textwidth}
        \centering
        \includegraphics[width=\textwidth]{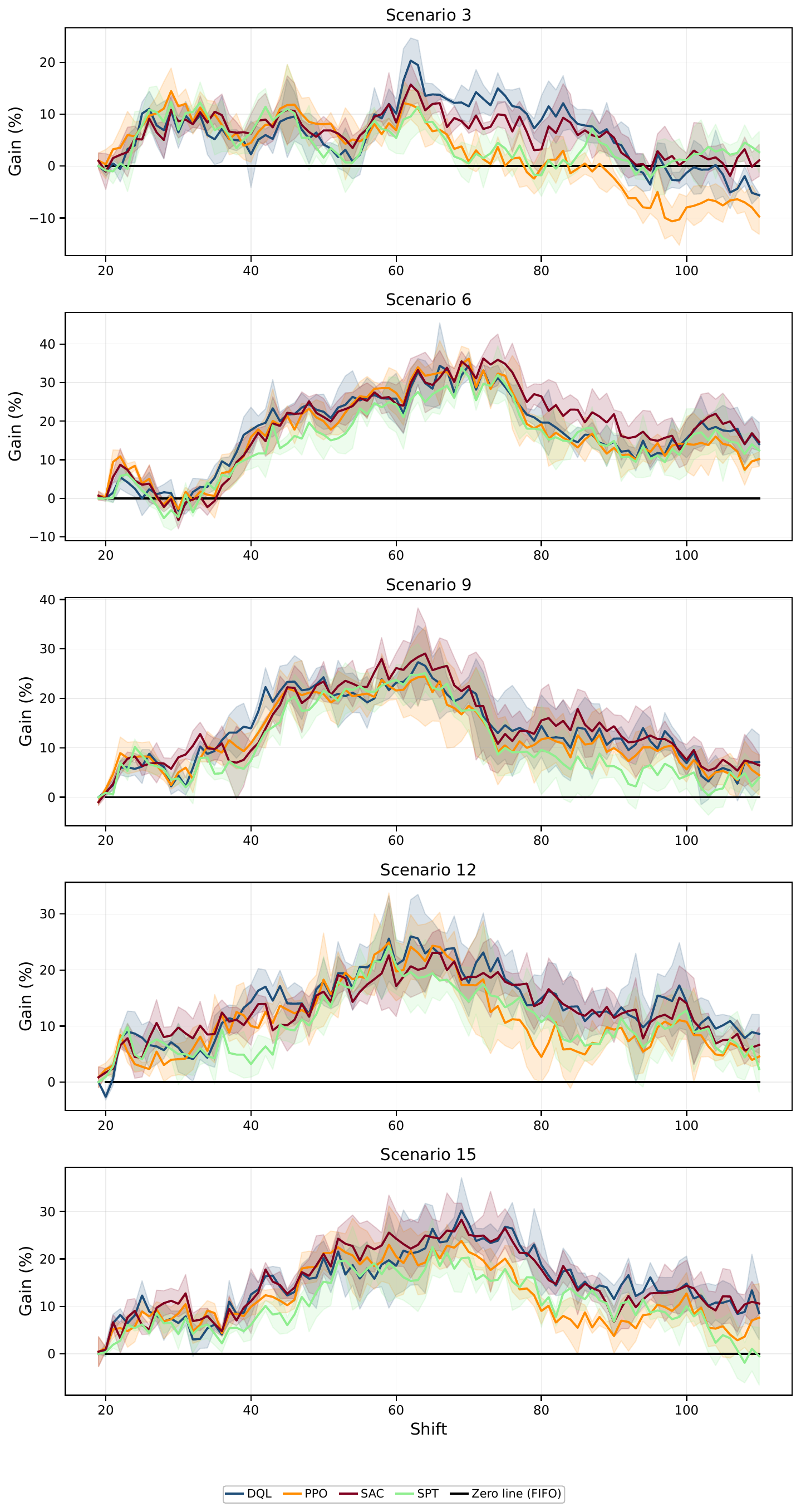}
        \caption{Against SPT: Scenarios 3, 6, 9, 12, and 15}
    \end{subfigure}
    \caption{Scenario-by-scenario Saturation gain (\%) of the online agents relative to the FIFO baseline for the first group of evaluation scenarios.}
    \label{fig:online-scen-saturation-g1}
\end{figure}

\begin{figure}[H]
    \centering
    \begin{subfigure}[t]{0.49\textwidth}
        \centering
        \includegraphics[width=\textwidth]{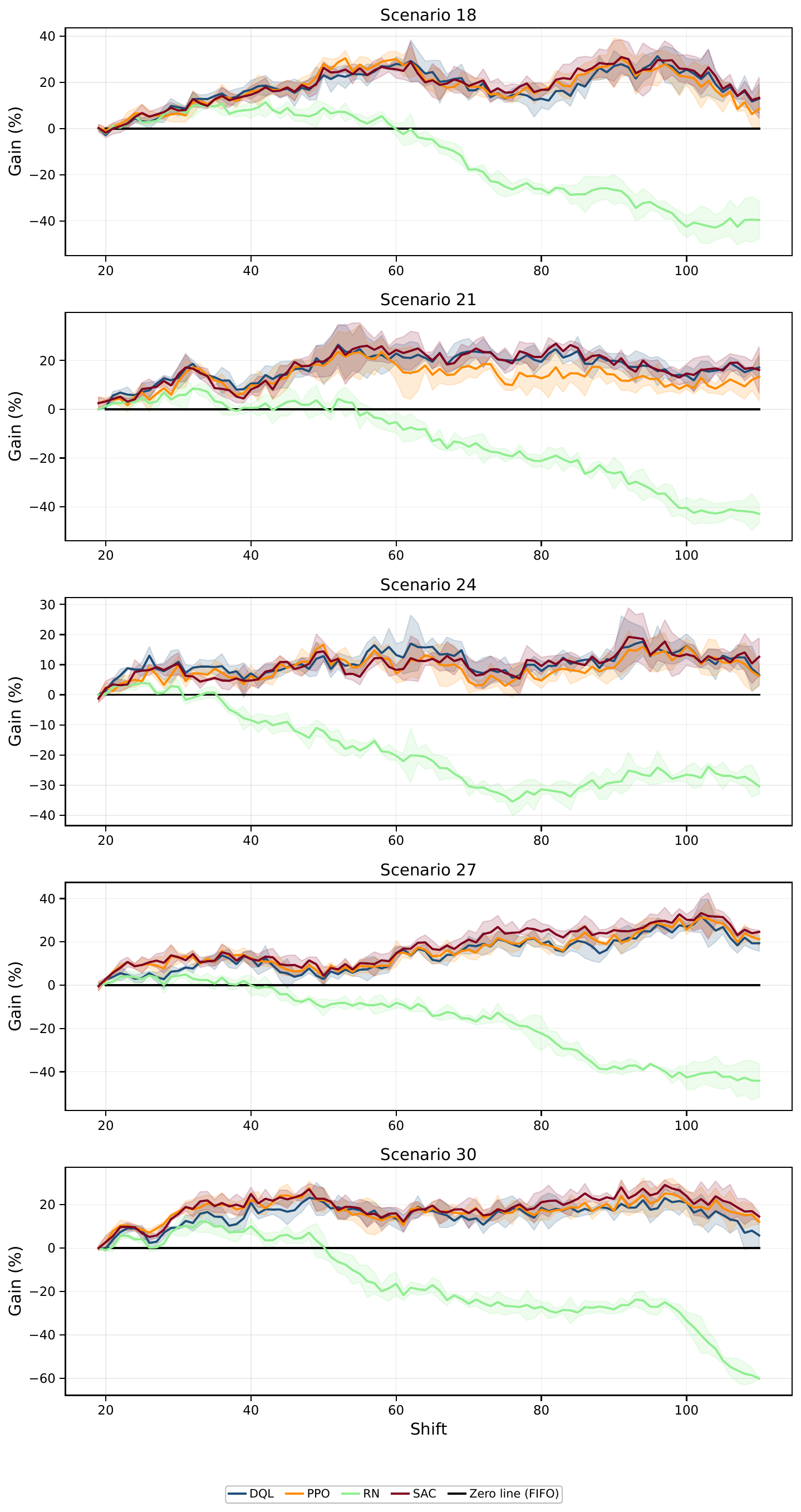}
        \caption{Against Random: Scenarios 18, 21, 24, 27, and 30}
    \end{subfigure}
    \hfill
    \begin{subfigure}[t]{0.49\textwidth}
        \centering
        \includegraphics[width=\textwidth]{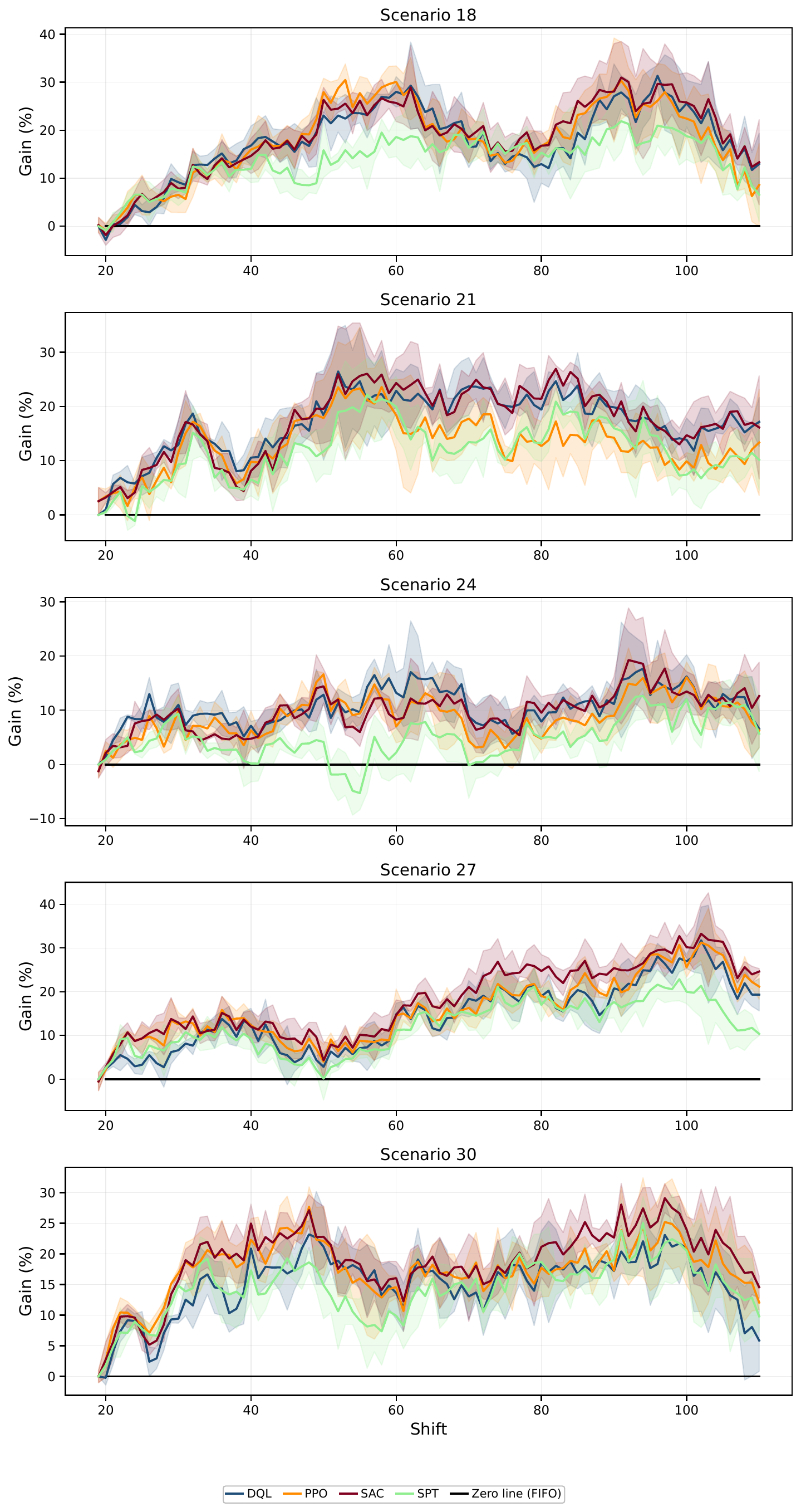}
        \caption{Against SPT: Scenarios 18, 21, 24, 27, and 30}
    \end{subfigure}
    \caption{Scenario-by-scenario Saturation gain (\%) of the online agents relative to the FIFO baseline for the second group of evaluation scenarios.}
    \label{fig:online-scen-saturation-g2}
\end{figure}

\begin{figure}[H]
    \centering
    \begin{subfigure}[t]{0.49\textwidth}
        \centering
        \includegraphics[width=\textwidth]{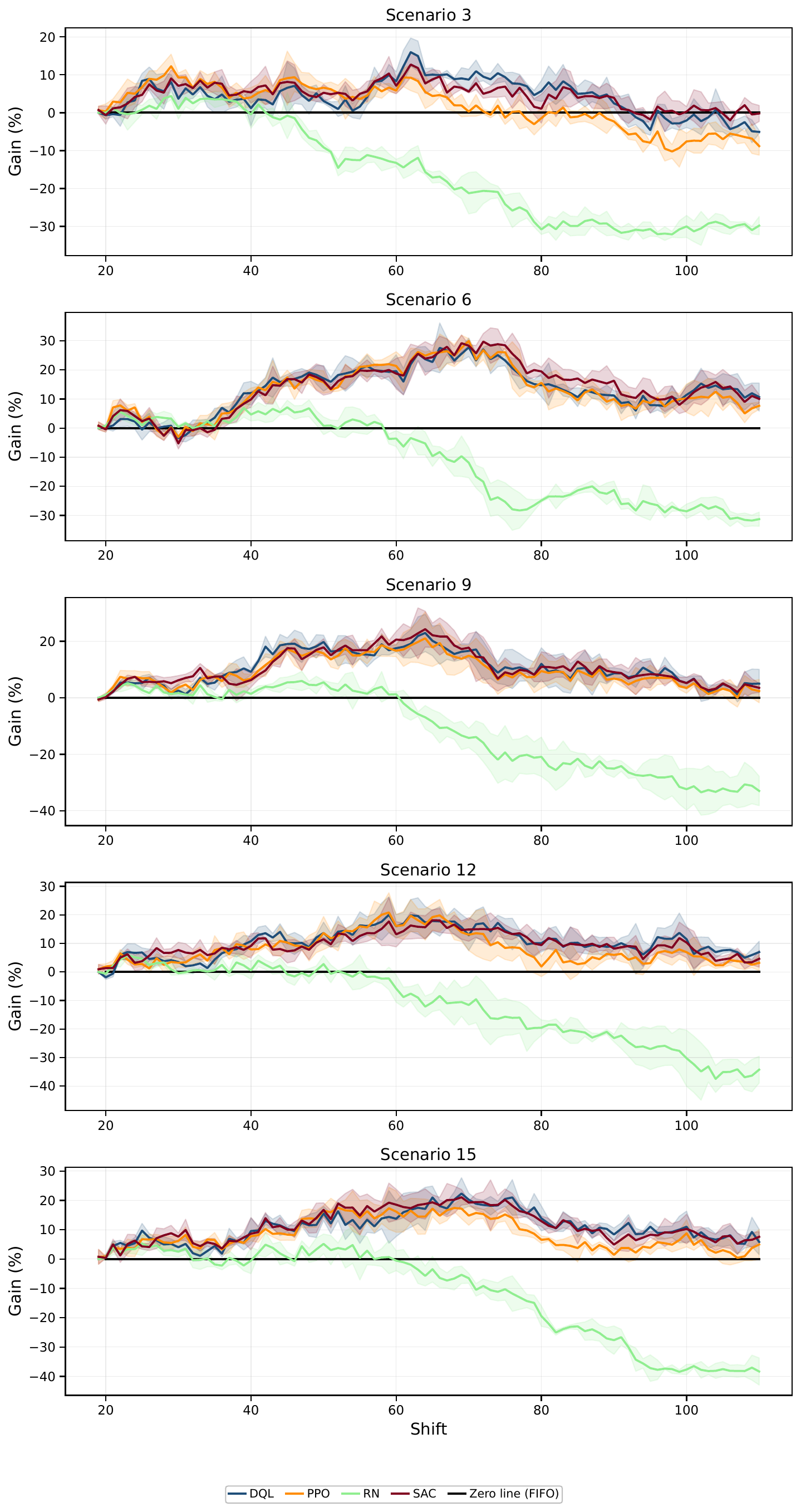}
        \caption{Against Random: Scenarios 3, 6, 9, 12, and 15}
    \end{subfigure}
    \hfill
    \begin{subfigure}[t]{0.49\textwidth}
        \centering
        \includegraphics[width=\textwidth]{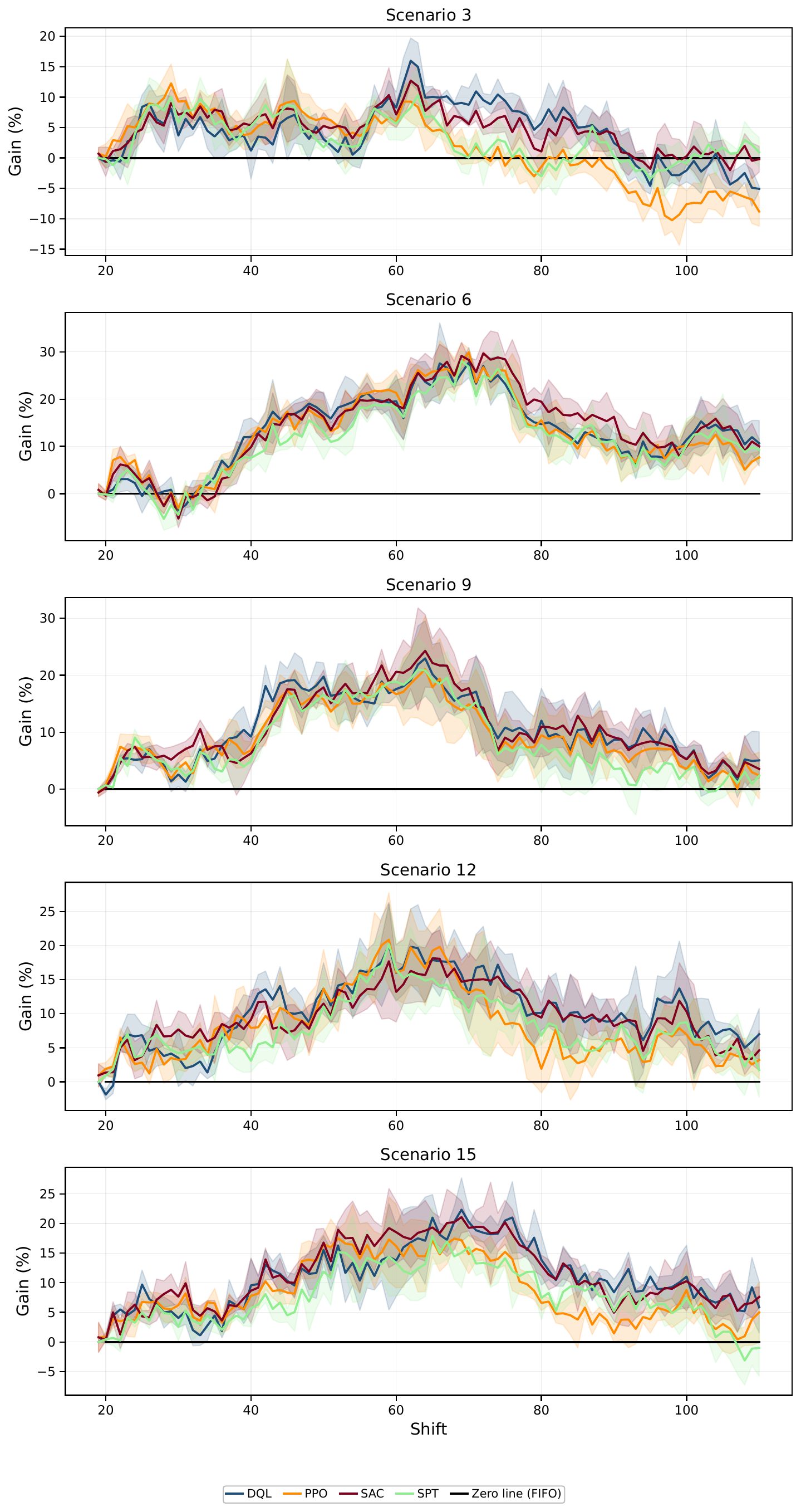}
        \caption{Against SPT: Scenarios 3, 6, 9, 12, and 15}
    \end{subfigure}
    \caption{Scenario-by-scenario Load gain (\%) of the online agents relative to the FIFO baseline for the first group of evaluation scenarios.}
    \label{fig:online-scen-windowload-g1}
\end{figure}

\begin{figure}[H]
    \centering
    \begin{subfigure}[t]{0.49\textwidth}
        \centering
        \includegraphics[width=\textwidth]{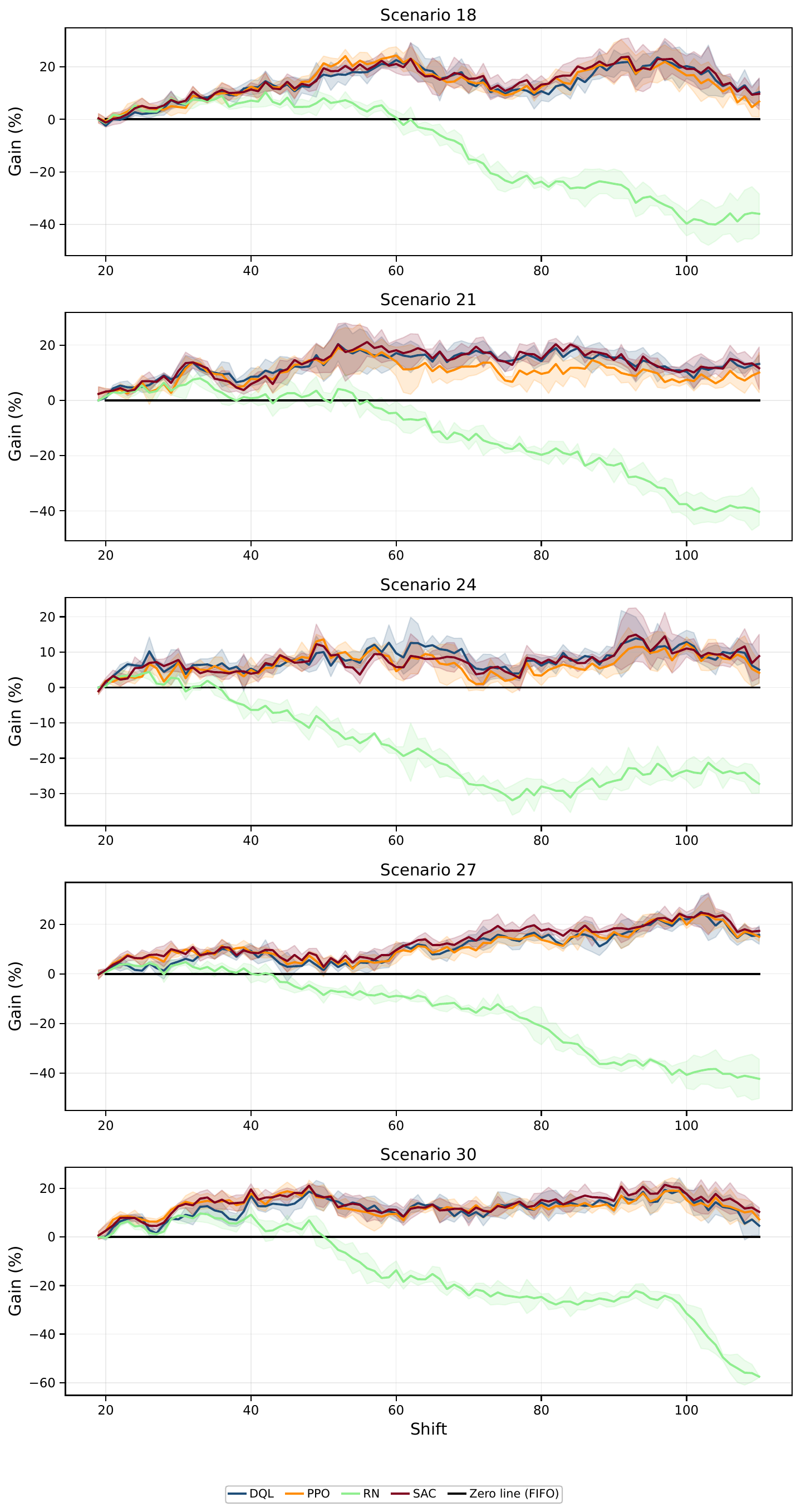}
        \caption{Against Random: Scenarios 18, 21, 24, 27, and 30}
    \end{subfigure}
    \hfill
    \begin{subfigure}[t]{0.49\textwidth}
        \centering
        \includegraphics[width=\textwidth]{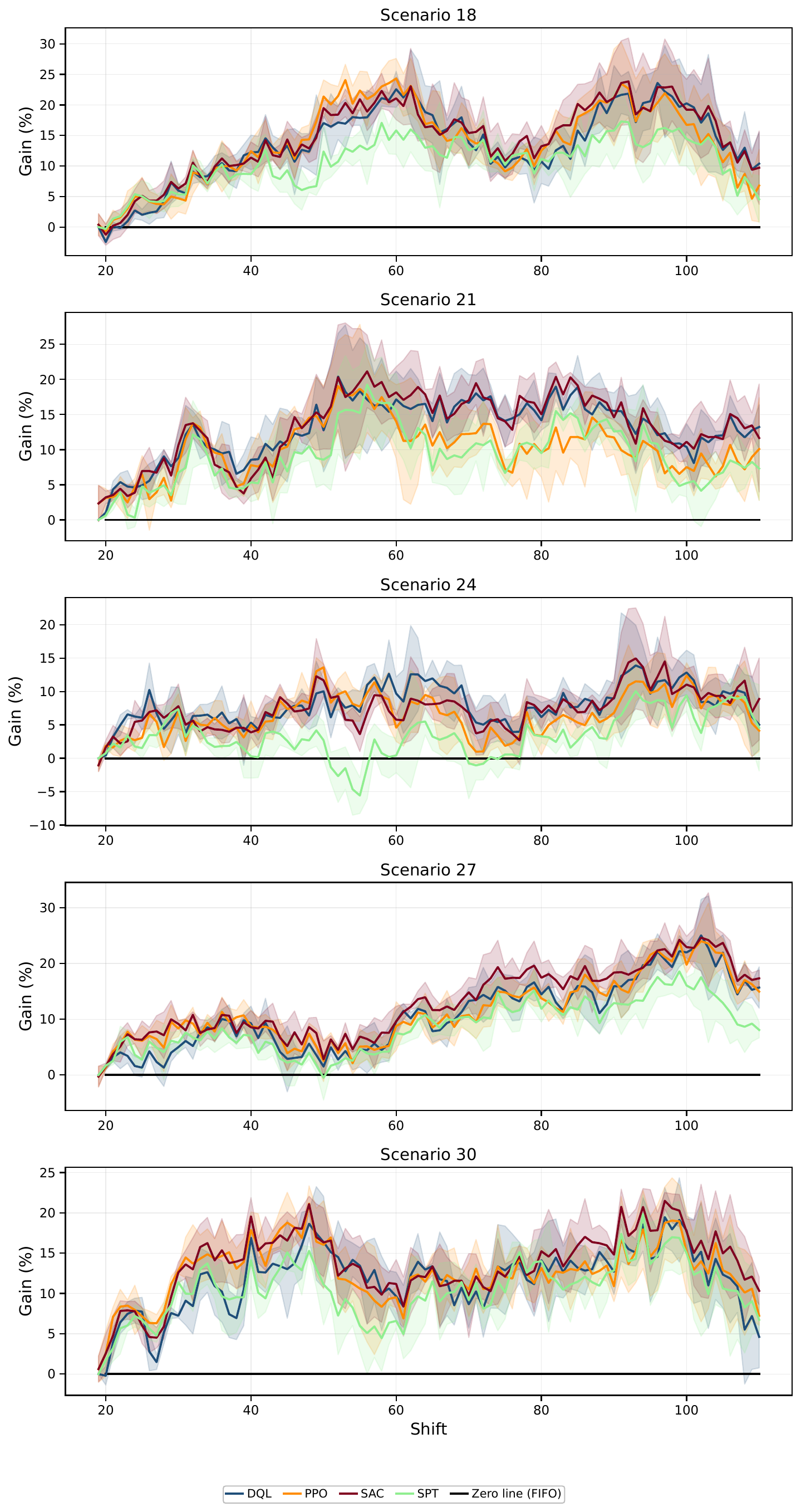}
        \caption{Against SPT: Scenarios 18, 21, 24, 27, and 30}
    \end{subfigure}
    \caption{Scenario-by-scenario Load gain (\%) of the online agents relative to the FIFO baseline for the second group of evaluation scenarios.}
    \label{fig:online-scen-windowload-g2}
\end{figure}

\paragraph{Online Agents}
\label{sec:online_performance_scenarios}
To complement the aggregated online results reported in Section~\ref{subsec:online}, Figures~\ref{fig:online-scen-throughput-g1}--\ref{fig:online-scen-windowload-g2} present a more detailed scenario-by-scenario analysis of the KPI gains relative to the FIFO baseline, organized in the same side-by-side layout as the corresponding figures in the main text. For each KPI, the left panel shows the Random setting and the right panel shows the SPT setting. To improve readability, the ten evaluation scenarios are divided into two figure groups. The first group includes Scenarios~3, 6, 9, 12, and 15, while the second group includes Scenarios~18, 21, 24, 27, and 30. This organization facilitates comparison of the online agents across baselines, scenarios, and shift bins for each KPI. While Table~\ref{tab:online} reports the mean performance across all evaluated scenarios, these grouped figures provide a finer-grained view of how consistently the observed gains are achieved under different operating conditions. Across all panels, the black horizontal line denotes the FIFO (Zero Line) baseline, while burgundy , dark blue, and orange represent SAC, DQL, and PPO, respectively. In addition, the light green curve denotes Random in the left panels, whereas represents SPT in the right panels.


\newpage
\paragraph{PPO with Segment-Only Objective}
\label{app:ppo_segment_only_plots}

To complement Table~\ref{tab:ppo_reward}, this appendix reports additional scenario-wise plots for the plain PPO variant trained with \emph{segment-level reward only}. The reported values are KPI gains (\%) relative to the FIFO baseline, while the Random and SPT policies are included as reference curves. In these plots, the black horizontal line denotes the FIFO reference, the burgundy curve denotes PPO, and the light green curve denotes the reference policy, namely Random in the left panels and SPT in the right panels. The figures report throughput, saturation, and load, where saturation captures instantaneous equipment utilization and load represents utilization measured over a time window.


\begin{figure}[H]
    \centering
    \begin{subfigure}[t]{0.49\textwidth}
        \centering
        \includegraphics[width=\textwidth]{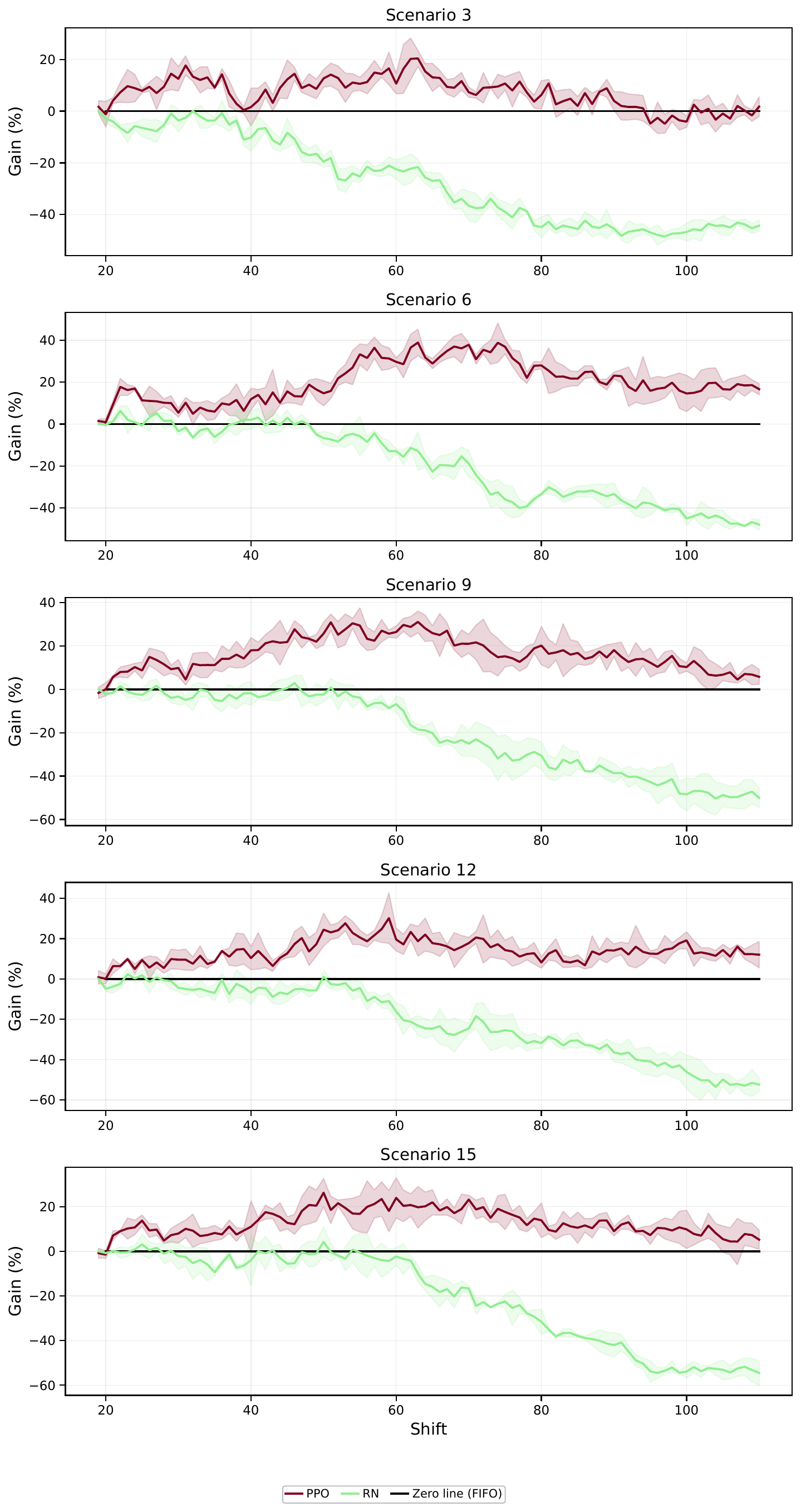}
        \caption{Against Random: Scenarios 3, 6, 9, 12, and 15.}
        \label{fig:ppo-seg-throughput-g1-random}
    \end{subfigure}
    \hfill
    \begin{subfigure}[t]{0.49\textwidth}
        \centering
        \includegraphics[width=\textwidth]{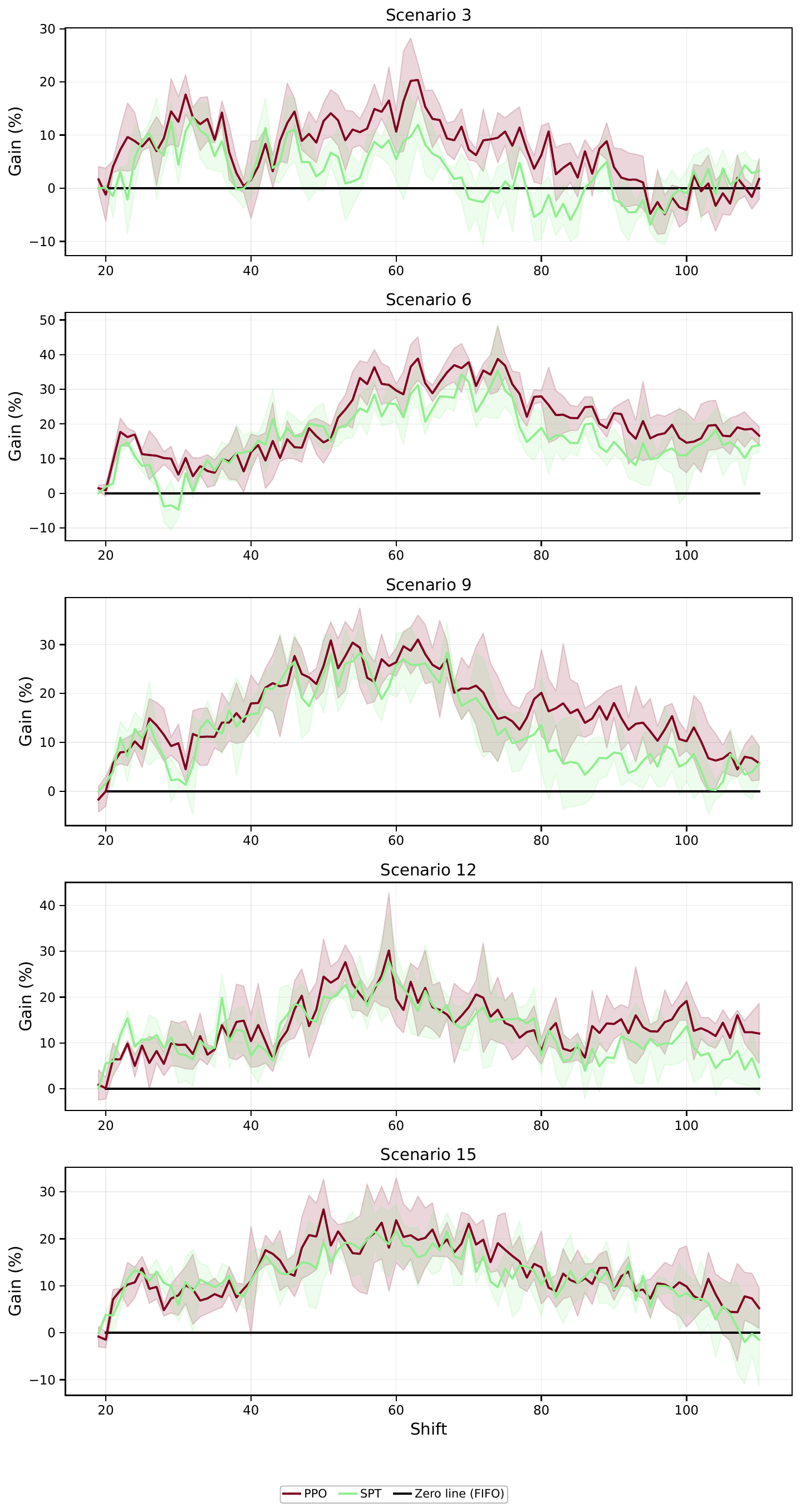}
        \caption{Against SPT: Scenarios 3, 6, 9, 12, and 15.}
        \label{fig:ppo-seg-throughput-g1-spt}
    \end{subfigure}
    \caption{Scenario-by-scenario Throughput gain (\%) of the plain PPO trained agent with segment level reward only relative to the FIFO baseline for the first group of evaluation scenarios. The left panel shows results under the Random policy, while the right panel corresponds to the SPT policy.}
    \label{fig:ppo-seg-throughput-g1}
\end{figure}

\begin{figure}[H]
    \centering
    \begin{subfigure}[t]{0.49\textwidth}
        \centering
        \includegraphics[width=\textwidth]{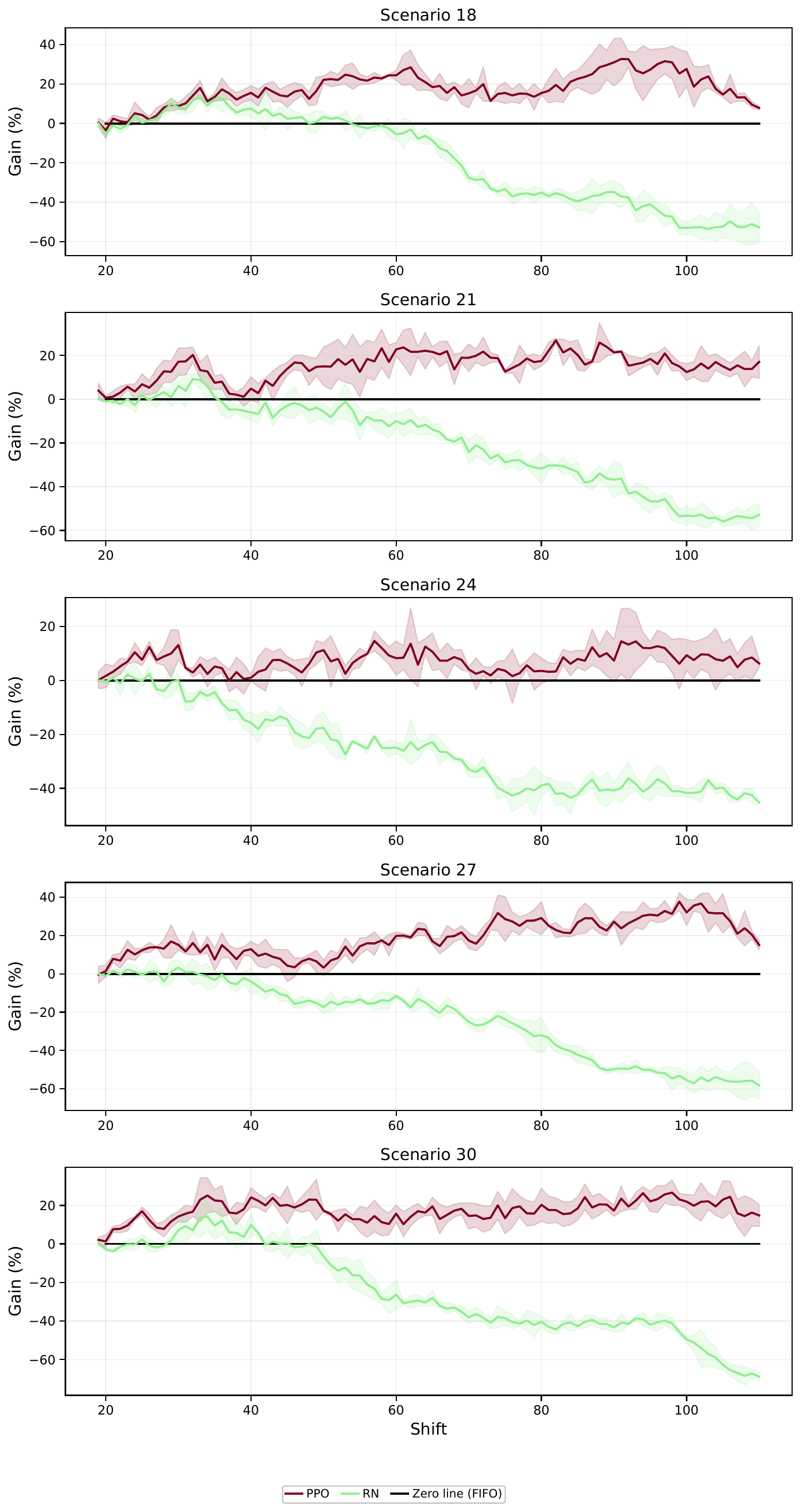}
        \caption{Against Random: Scenarios 18, 21, 24, 27, and 30.}
        \label{fig:ppo-seg-throughput-g2-random}
    \end{subfigure}
    \hfill
    \begin{subfigure}[t]{0.49\textwidth}
        \centering
        \includegraphics[width=\textwidth]{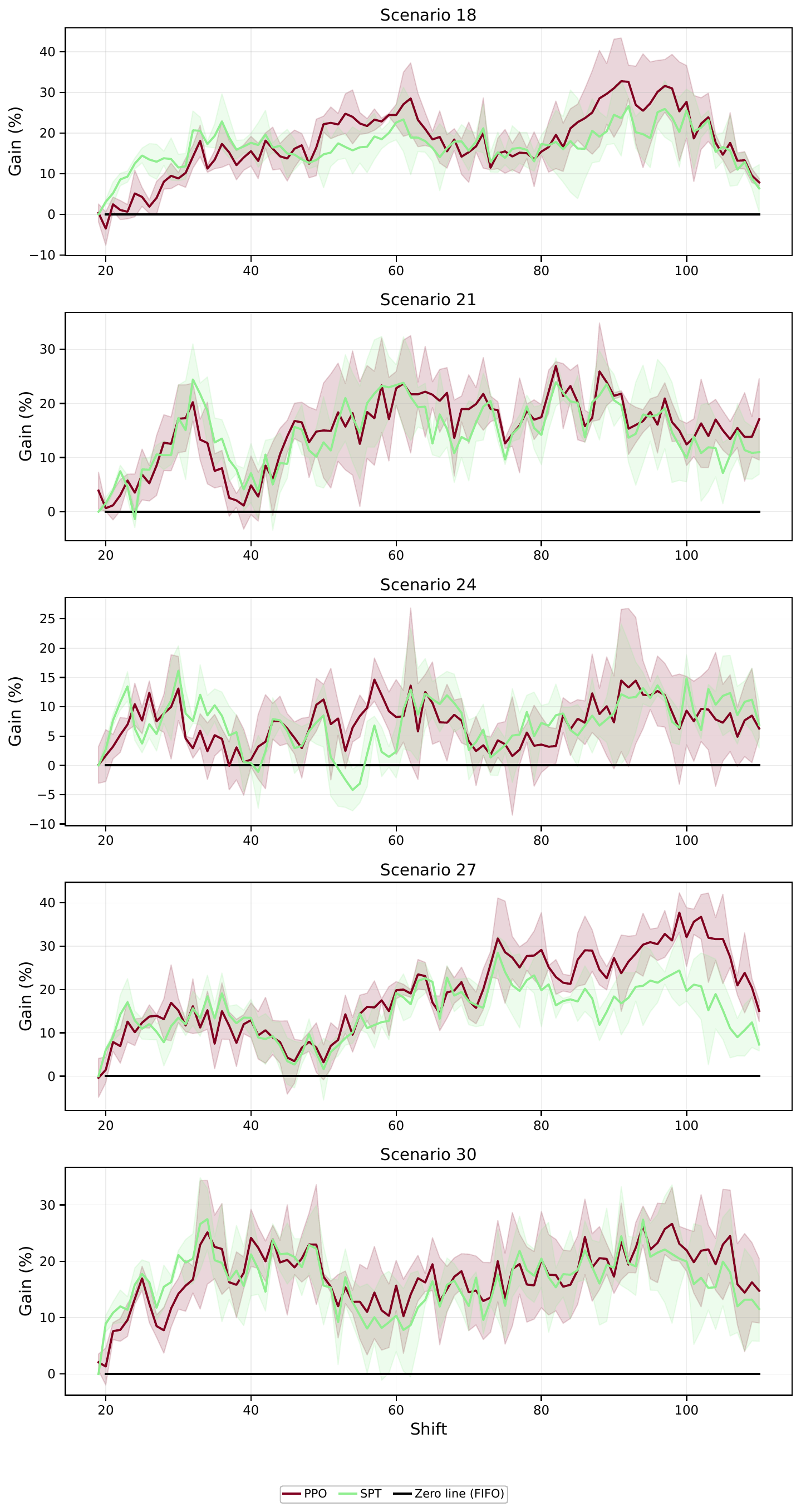}
        \caption{Against SPT: Scenarios 18, 21, 24, 27, and 30.}
        \label{fig:ppo-seg-throughput-g2-spt}
    \end{subfigure}
    \caption{Scenario-by-scenario throughput gains (\%) relative to the FIFO baseline for plain PPO trained with segment-level reward only, for the second group of evaluation scenarios.}
    \label{fig:ppo-seg-throughput-g2}
\end{figure}


\begin{figure}[H]
    \centering
    \begin{subfigure}[t]{0.49\textwidth}
        \centering
        \includegraphics[width=\textwidth]{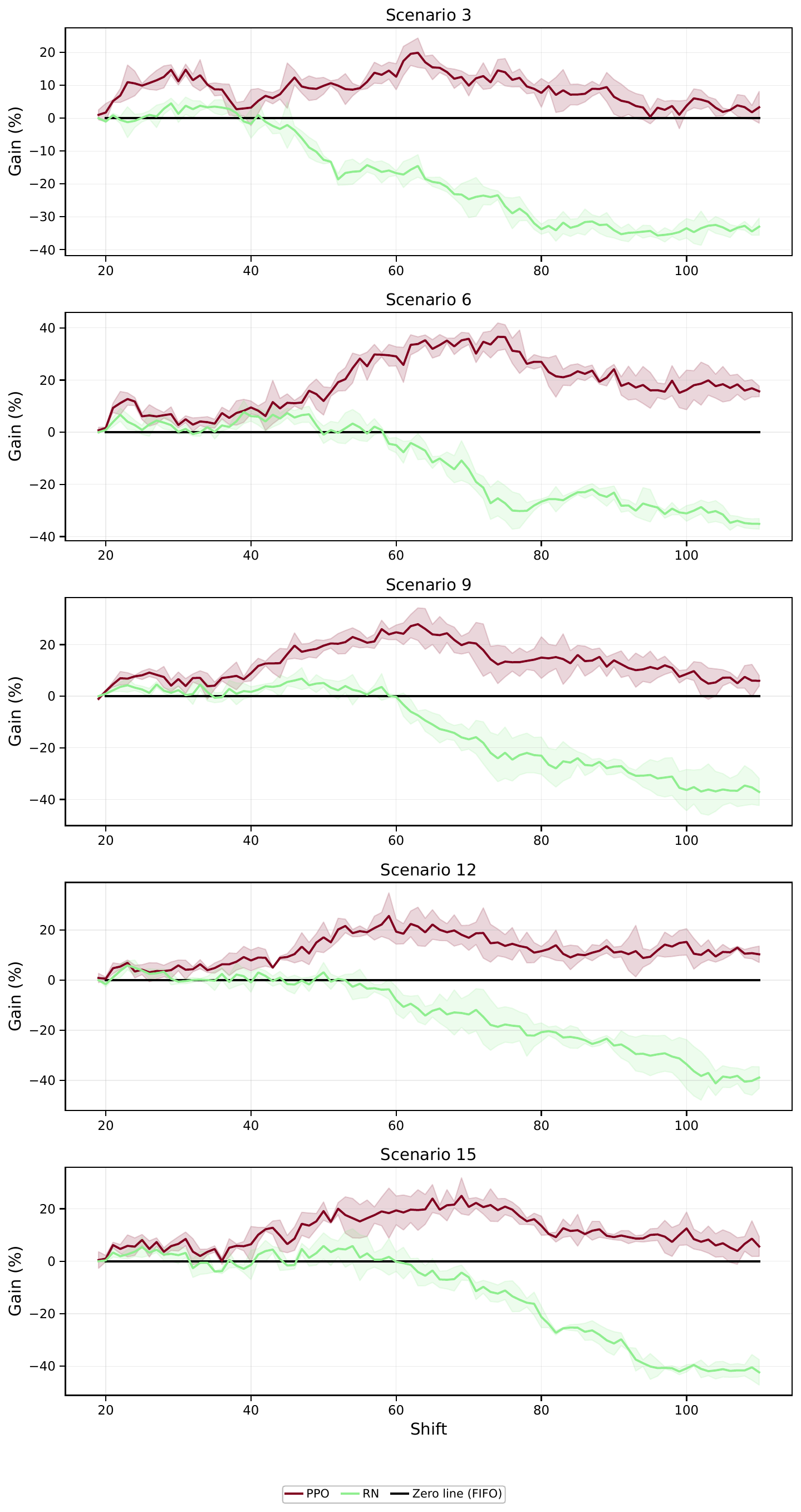}
        \caption{Against Random: Scenarios 3, 6, 9, 12, and 15.}
        \label{fig:ppo-seg-saturation-g1-random}
    \end{subfigure}
    \hfill
    \begin{subfigure}[t]{0.49\textwidth}
        \centering
        \includegraphics[width=\textwidth]{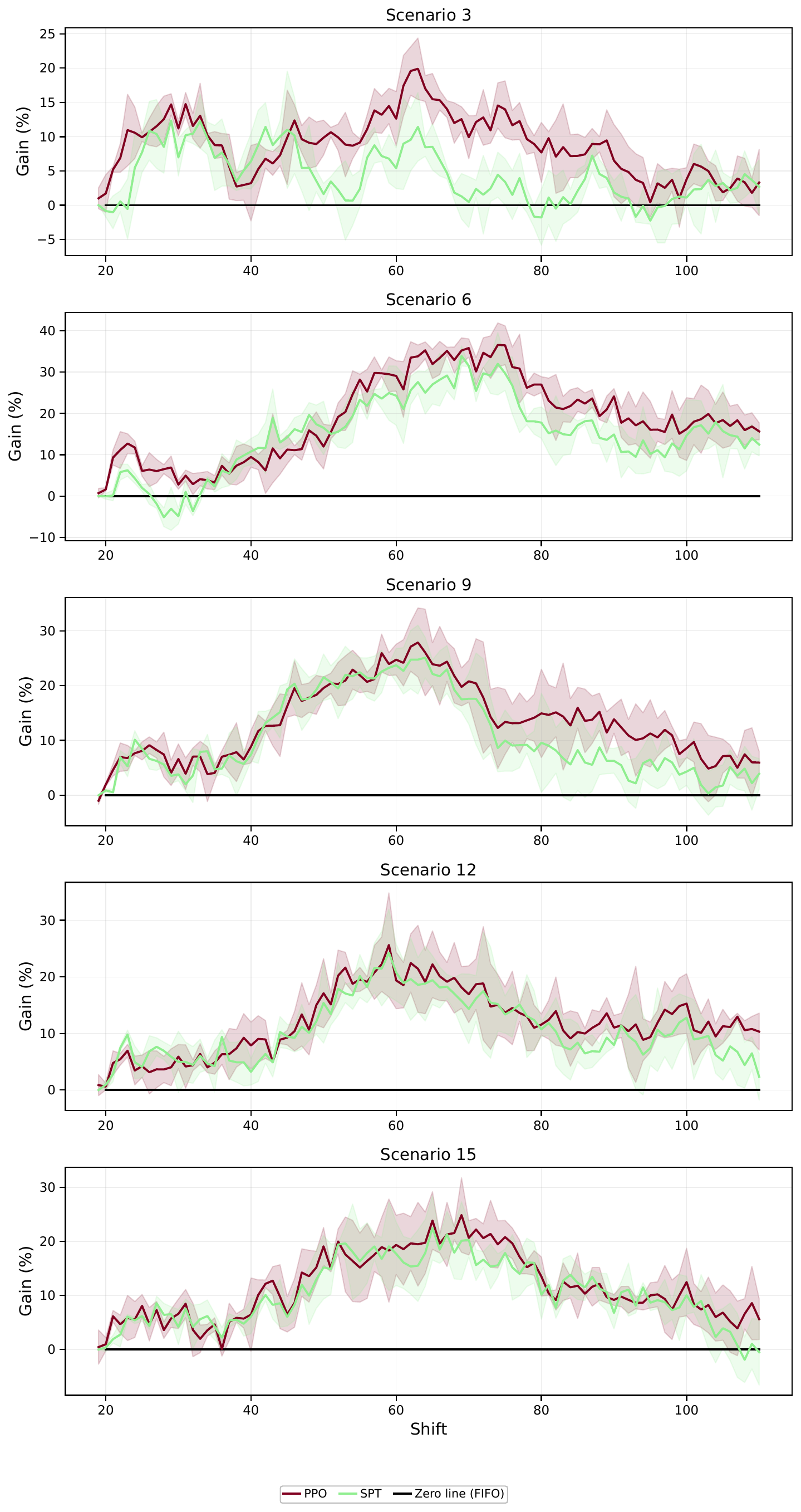}
        \caption{Against SPT: Scenarios 3, 6, 9, 12, and 15.}
        \label{fig:ppo-seg-saturation-g1-spt}
    \end{subfigure}
    \caption{Scenario-by-scenario saturation gains (\%) relative to the FIFO baseline for plain PPO trained with segment-level reward only, for the first group of evaluation scenarios.}
    \label{fig:ppo-seg-saturation-g1}
\end{figure}

\begin{figure}[H]
    \centering
    \begin{subfigure}[t]{0.49\textwidth}
        \centering
        \includegraphics[width=\textwidth]{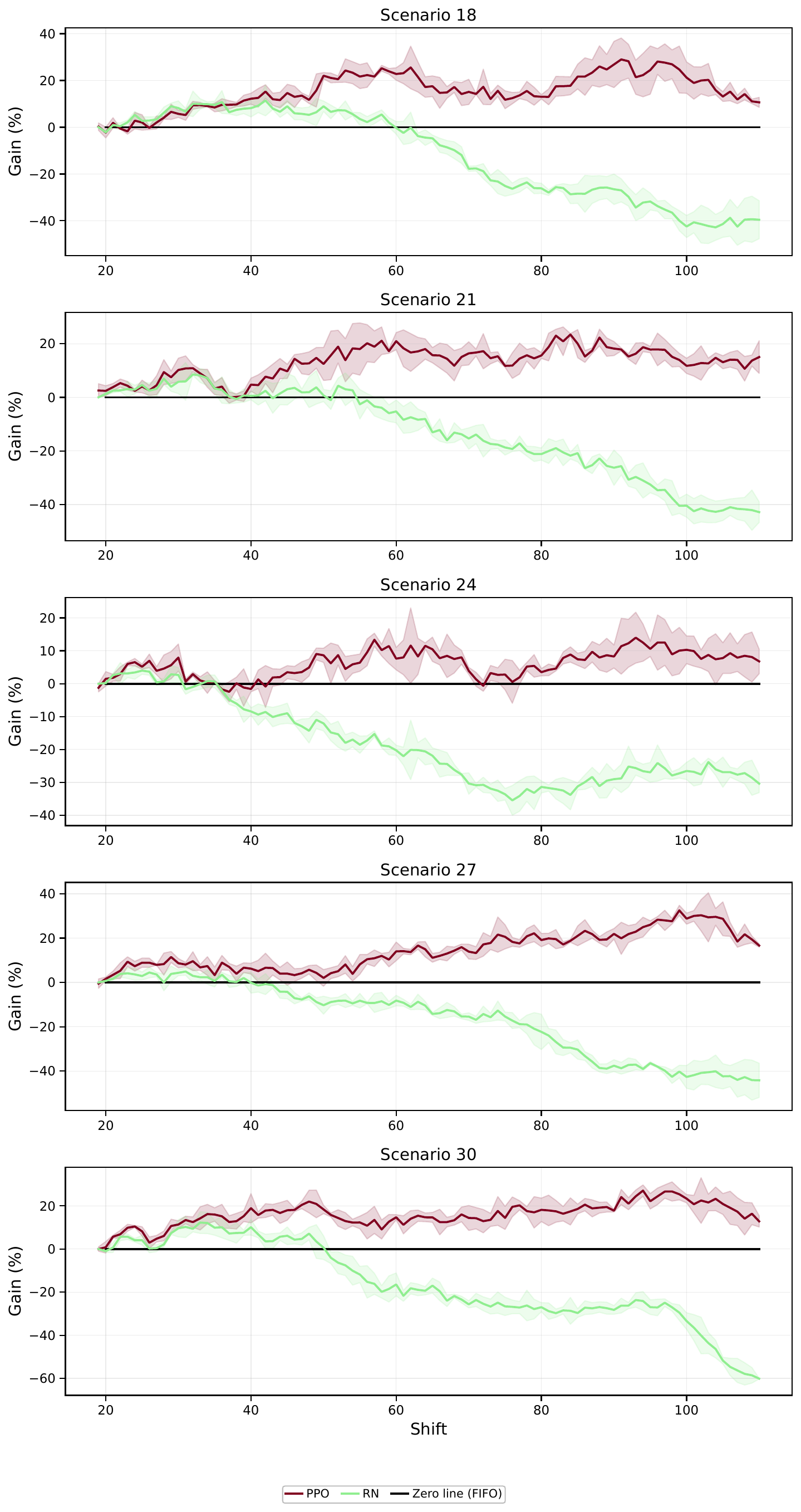}
        \caption{Against Random: Scenarios 18, 21, 24, 27, and 30.}
        \label{fig:ppo-seg-saturation-g2-random}
    \end{subfigure}
    \hfill
    \begin{subfigure}[t]{0.49\textwidth}
        \centering
        \includegraphics[width=\textwidth]{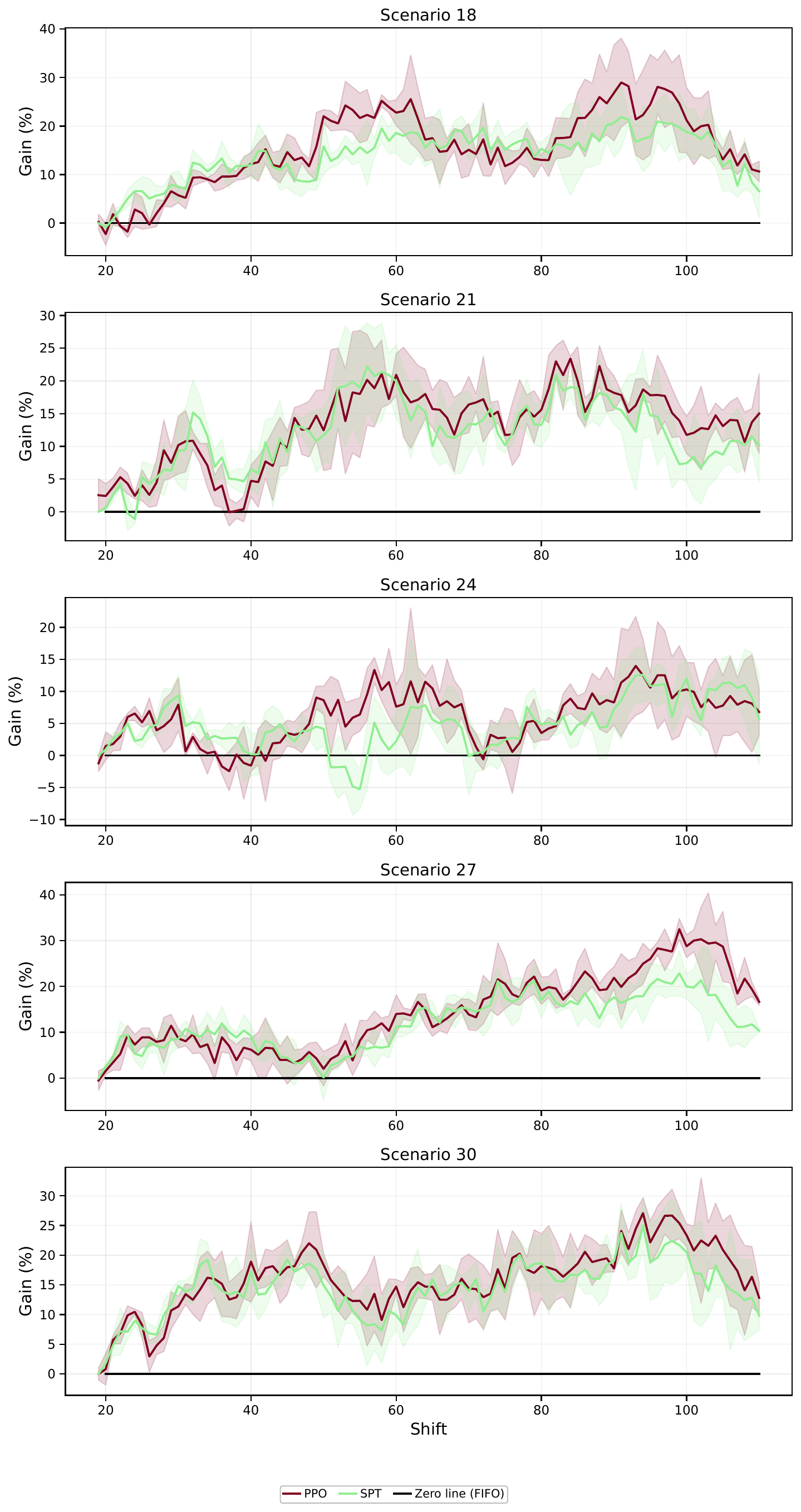}
        \caption{Against SPT: Scenarios 18, 21, 24, 27, and 30.}
        \label{fig:ppo-seg-saturation-g2-spt}
    \end{subfigure}
    \caption{Scenario-by-scenario saturation gains (\%) relative to the FIFO baseline for plain PPO trained with segment-level reward only, for the second group of evaluation scenarios.}
    \label{fig:ppo-seg-saturation-g2}
\end{figure}


\begin{figure}[H]
    \centering
    \begin{subfigure}[t]{0.49\textwidth}
        \centering
        \includegraphics[width=\textwidth]{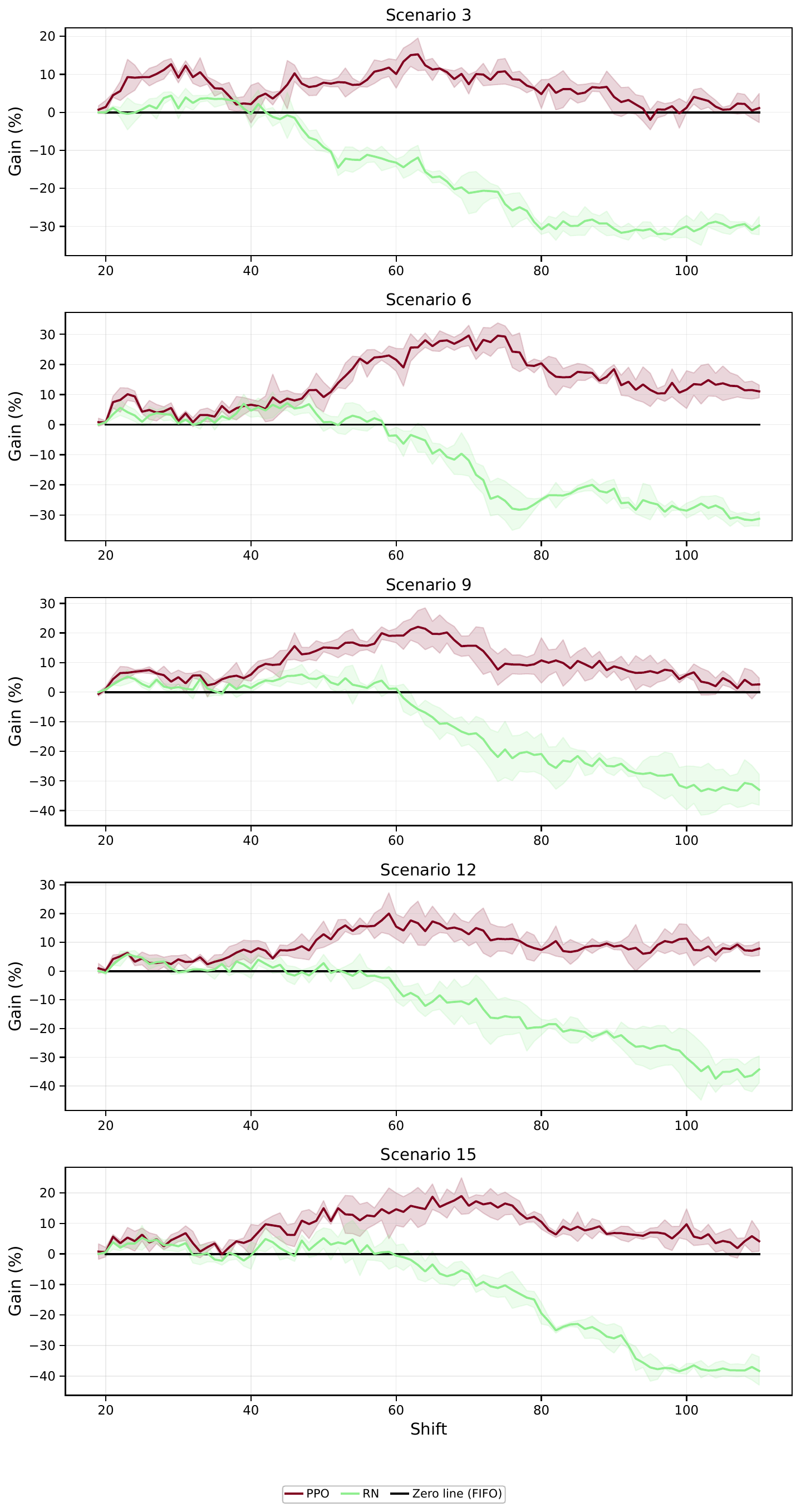}
        \caption{Against Random: Scenarios 3, 6, 9, 12, and 15.}
        \label{fig:ppo-seg-load-g1-random}
    \end{subfigure}
    \hfill
    \begin{subfigure}[t]{0.49\textwidth}
        \centering
        \includegraphics[width=\textwidth]{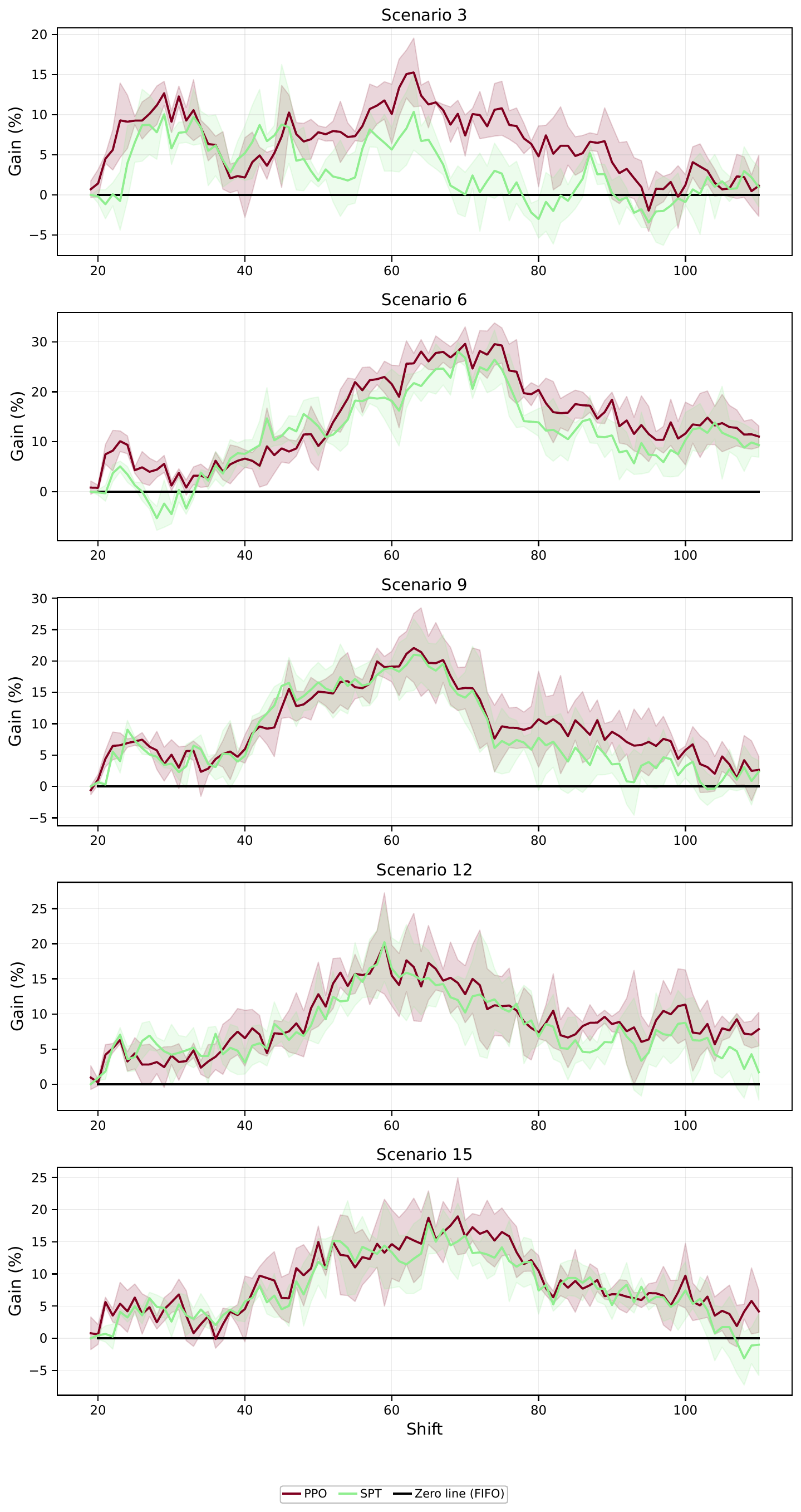}
        \caption{Against SPT: Scenarios 3, 6, 9, 12, and 15.}
        \label{fig:ppo-seg-load-g1-spt}
    \end{subfigure}
    \caption{Scenario-by-scenario load gains (\%) relative to the FIFO baseline for plain PPO trained with segment-level reward only, for the first group of evaluation scenarios. Here, load represents utilization measured over a time window.}
    \label{fig:ppo-seg-load-g1}
\end{figure}

\begin{figure}[H]
    \centering
    \begin{subfigure}[t]{0.49\textwidth}
        \centering
        \includegraphics[width=\textwidth]{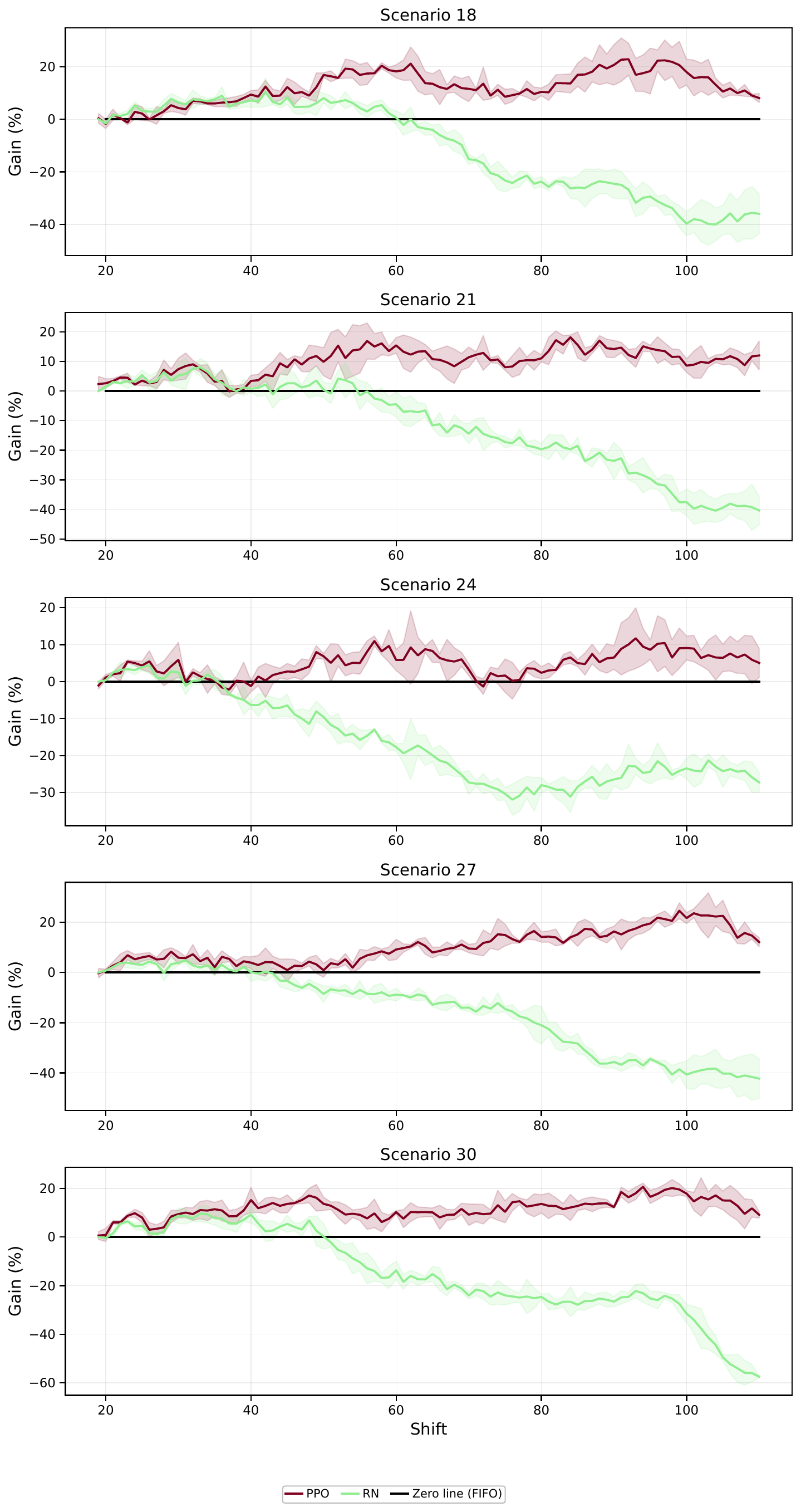}
        \caption{Against Random: Scenarios 18, 21, 24, 27, and 30.}
        \label{fig:ppo-seg-load-g2-random}
    \end{subfigure}
    \hfill
    \begin{subfigure}[t]{0.49\textwidth}
        \centering
        \includegraphics[width=\textwidth]{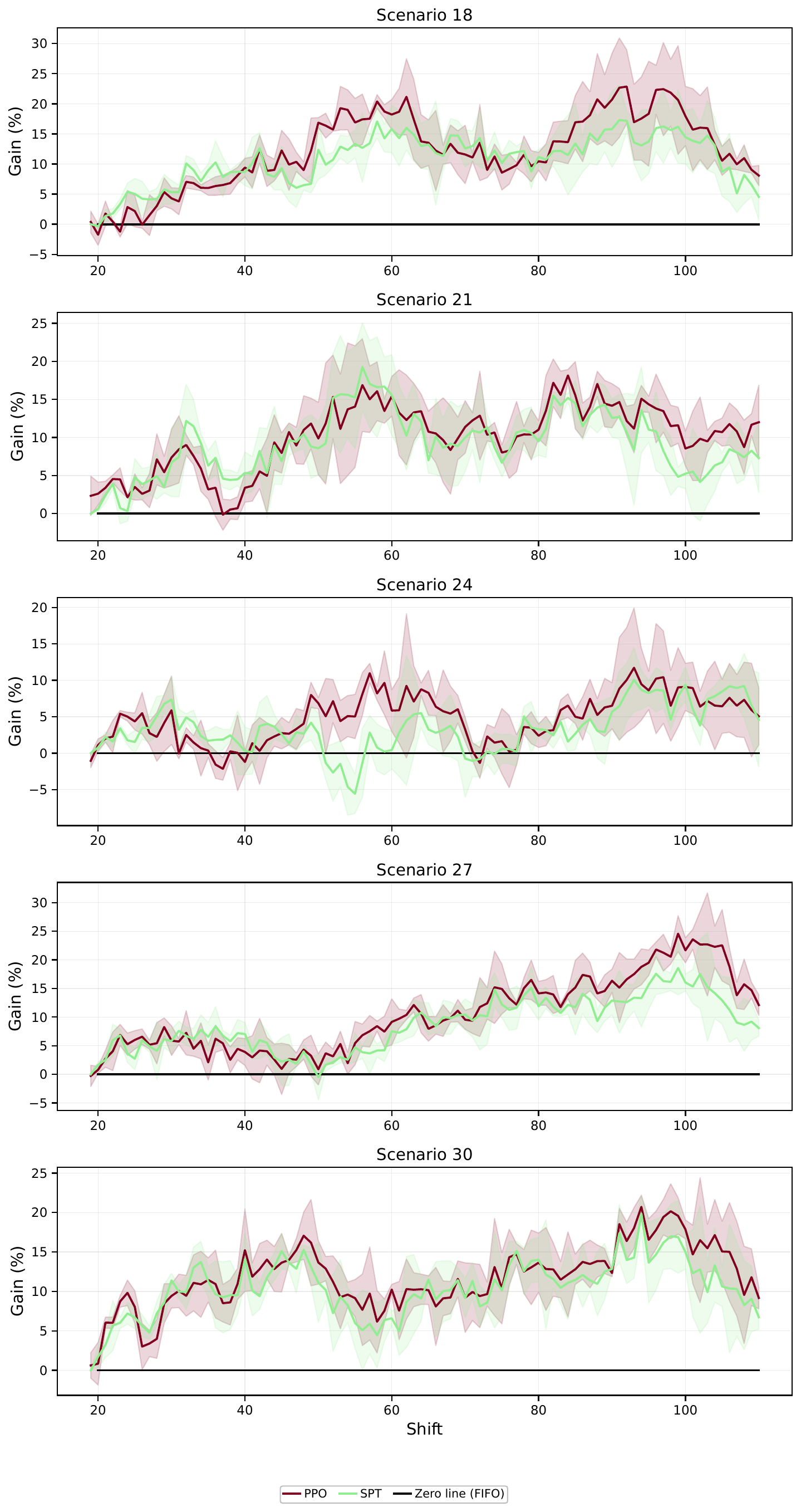}
        \caption{Against SPT: Scenarios 18, 21, 24, 27, and 30.}
        \label{fig:ppo-seg-load-g2-spt}
    \end{subfigure}
    \caption{Scenario-by-scenario load gains (\%) relative to the FIFO baseline for plain PPO trained with segment-level reward only, for the second group of evaluation scenarios. Here, load represents utilization measured over a time window.}
    \label{fig:ppo-seg-load-g2}
\end{figure}

\end{document}